\documentclass[accepted]{uai2025} 
                        
\usepackage[american]{babel}
\usepackage{csquotes}

\usepackage[
  style=authoryear-comp,
  natbib=true,
  backend=biber
]{biblatex}
\usepackage{mathtools} 
\usepackage{booktabs} 
\usepackage{tikz} 

\usepackage{array}
\usepackage{graphicx}
\usepackage{longtable}
\usepackage{wrapfig}
\usepackage{rotating}
\usepackage{amsmath}
\usepackage{amssymb}
\usepackage{bm}
\usepackage{capt-of}
\usepackage{subcaption}
\usepackage{booktabs}
\usepackage{multirow}
\usepackage{pdflscape}
\usepackage{cleveref}
\newcommand{\footnotehyperlink}[1]{\footnote{\hyperlink{#1}{#1}}}

\newcommand{\pdfplotcolumns}[3][0.2\columnwidth]{
  \parbox[c]{#1}{\includegraphics[width=#1, trim={5mm 5mm 5mm 5mm}, clip]{#2_unconditional_#3.pdf}} &
  \parbox[c]{#1}{\includegraphics[width=#1, trim={5mm 5mm 5mm 5mm}, clip]{#2_conditional_#3.pdf}}
}

\usepackage[acronym]{glossaries}
\glsdisablehyper

\newacronym{BNF}{BNF}{elementwise Bernstein normalizing fLow}
\newacronym{CDF}{CDF}{cumulative distribution function}
\newacronym{ECDF}{ECDF}{empirical \gls{CDF}}
\newacronym{PDF}{PDF}{probability density function}
\newacronym{CTM}{CTM}{conditional transformation model}
\newacronym{GAM}{GAM}{generalized additive model}
\newacronym{IAF}{IAF}{inverse autoregressive flow}
\newacronym{MADE}{MADE}{masked autoencoder for distribution estimation}
\newacronym{MAF}{MAF}{masked autoregressive flow}
\newacronym{CF}{CF}{coupling flow}
\newacronym{HMAF}{HMAF}{hybrid \gls{MAF}}
\newacronym{HCF}{HCF}{hybrid \gls{CF}}
\newacronym{MCTM}{MCTM}{multivariate conditional transformation model}
\newacronym{MVN}{MVN}{multivariate normal}
\newacronym{NF}{NF}{normalizing flow}
\newacronym{NVP}{real NVP}{real-valued non-volume preserving flows}
\newacronym{PIT}{PIT}{probability integral transform}
\newacronym{SAP}{SAP}{structured additive predictor}
\newacronym{xAI}{xAI}{explainable AI}
\newacronym{NLL}{NLL}{negative logarithmic likelihood}
\newacronym{QQ}{Q-Q}{quantile-quantile}

\usepackage{tikz}
\usetikzlibrary{tikzmark}
\definecolor{mpl_blue}{HTML}{1f77b4}
\definecolor{mpl_orange}{HTML}{ff7f0e}

\DeclareMathOperator*{\nll}{NLL}

\DeclareMathOperator{\softmax}{softmax}
\DeclareMathOperator{\softplus}{softplus}

\graphicspath{{./gfx}}

\title{Hybrid Bernstein Normalizing Flows for Flexible Multivariate Density Regression with Interpretable Marginals}

\author[1]{\href{mailto:<marcel.arpogaus@htwg-konstanz.de>?Subject=Your UAI 2025 paper}{Marcel Arpogaus}{}}
\author[2]{Thomas Kneib}
\author[3,4]{Thomas Nagler}
\author[3,4]{David R\"ugamer}

\affil[1]{%
    Institute for Applied Research \\
    HTWG - University of Applied Sciences \\
    Konstanz, GERMANY
}
\affil[2]{%
    Chair of Statistics and Campus Institute Data Science (CIDAS) \\
    University of Göttingen \\
    Göttingen, GERMANY
}
\affil[3]{%
    Department of Statistics \\
    LMU Munich\\
    Munich, GERMANY
}
\affil[4]{%
    Munich Center for Machine Learning (MCML) \\
    Munich, GERMANY
}
\begin{document}
\maketitle

\begin{abstract}
  Density regression models allow a comprehensive understanding of data by modeling the complete conditional probability distribution.
  While flexible estimation approaches such as~\glspl{NF} work particularly well in multiple dimensions, interpreting the input-output relationship of such models is often difficult, due to the black-box character of deep learning models.
  In contrast, existing statistical methods for multivariate outcomes such as~\glspl{MCTM} are restricted in flexibility and are often not expressive enough to represent complex multivariate probability distributions.
  In this paper, we combine~\glspl{MCTM} with state-of-the-art and autoregressive~\glspl{NF} to leverage the transparency of~\glspl{MCTM} for modeling interpretable feature effects on the marginal distributions in the first step and the flexibility of neural-network-based~\gls{NF} techniques to account for complex and non-linear relationships in the joint data distribution. We demonstrate our method's versatility in various numerical experiments and compare it with~\glspl{MCTM} and other \gls{NF} models on both simulated and real-world data.
\end{abstract}

\section{Introduction}
\label{sec:introduction}

Many real-world regression problems involve modeling of complex conditional distributions of an outcome $Y$ given an input vector $\mathbf{x} = [x_1,\ldots,x_D]^\top$.
In such instances, a mere prediction of the mean would be insufficient to capture the aleatoric uncertainty inherent in the data-generating process and a comprehensive understanding requires estimating the conditional distribution $Y|\mathbf{x}$~\citep{Kneib2021}.
This becomes more challenging for multivariate outcomes $\mathbf{Y} \in \mathbb{R}^D$, both in terms of estimation and model interpretation.
While \acrfull{NF}~\citep{Papamakarios2018} models can represent such complex conditional multivariate distributions, their deep neural network architectures often hinder interpretability.
Statistical methods for flexible density estimation in high dimensions have largely focused on graphical models, often involving copulas \citep[e.g.,][]{Liu2011,Bauer2016,Nagler2016}.
A notable exception is the \gls{MCTM}, a transformation model approach~\citep{Hothorn2018} that shares similarities with \glspl{NF} and models the joint relationship similar to copulas.
While less flexible than \glspl{NF}, \glspl{MCTM} preserve the interpretability of the feature-outcome relationship.

\begin{figure}[t!]
  \centering
  \includegraphics[width=0.7\columnwidth]{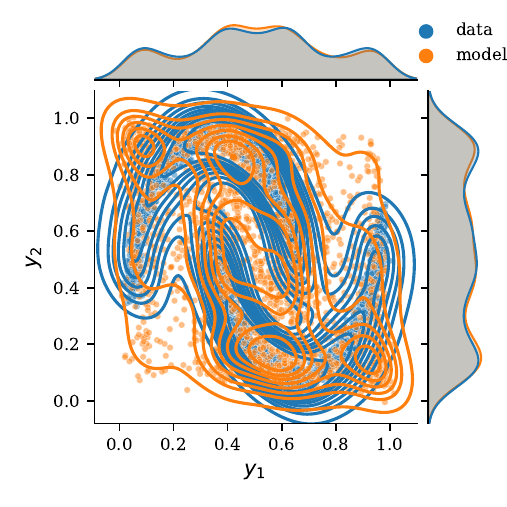}
  \caption{An \gls{MCTM} (orange) applied to a complex data distribution (blue).
    The model captures marginals well but fails to capture the dependence structure.}
  \label{fig:mctmfail}
\end{figure}

\paragraph{Our Contribution}
To bridge the gap between flexible black-box and rigid but interpretable statistical models, we propose a hybrid approach combining the transparency of \glspl{MCTM} with the flexibility of \glspl{NF}.
We thereby provide a method that can be applied if the understanding of feature effects on each response variable is indispensable. At the same time, our method allows complex modeling of the dependency structure of the outcome dimensions using the idea of masking from autoregressive flows.
We evaluate its effectiveness against \glspl{MCTM} and other density estimation models on simulated and real-world datasets.

\paragraph{Notation} We refer to random variables with a capital Latin letter, e.g., $Y$, and their realizations (observed or sampled) with a lowercase Latin letter, e.g.\ $y$.
The corresponding \gls{CDF} is referred to as $F_Y$ and the \gls{PDF} as $f_Y$.
For sets and vectors, we use bold-faced variables, e.g.\ $\mathbf{Y}=\{Y_1,\dots,Y_J\}$.
For individual parameters, optimized during the training procedure we use lowercase Greek letters, e.g.\ $\vartheta$, and bold-faced Greek letters for sets or vectors of parameters, e.g.\ $\bm{\vartheta}=\left(\vartheta_1,\ldots,\vartheta_M\right)^\top$.
Bold capital Greek letters indicate tuples, e.g.\ $\bm{\Theta}=\left(\bm{\theta}_1,\bm{\theta}_2,\alpha,\beta\right)$.

\section{Background and Related Work}
\label{sec:background}

Several methods have emerged for modeling complex conditional distributions.
Among these methods, transformation-based models have proven to be highly flexible and have been extensively used since their introduction by~\citet{Box1964}.

\subsection{Probability Transformation}

Transformation models are based on the \emph{probability-transformation theorem}, using a bijective and continuously differentiable \emph{transformation function} $h$ to map a complex distribution $F_Y$ to a simpler one $F_Z$ (often standard normal)~\citep{Kneib2021} via $F_Y = F_Z(h(y)).$ This allows density estimation without restrictive shape assumptions~\citep{Hothorn2018}.
The density $f_Y$ is calculated from the base density $f_Z$ by
\begin{equation}\label{eq:cov}
  f_Y(y) = f_Z\left(h(y)\right) \left|\det\nabla{h}(y)\right|,
\end{equation}
where the absolute Jacobian determinant $\left|\det\nabla{h}(y)\right|$ reflects the change in density induced by the transformation.
To sample from $F_Y$, we can draw samples $z$ from $F_Z$ and apply the inverse transformation $h^{-1}$ to obtain $y$.

\subsection{Normalizing Flows and Transformation Models}

\glspl{NF} apply the probability-transformation theorem using a series of $K$ simple bijective transformation functions $h_k$ to form more expressive transformations $h(z)= h_K \circ h_{K-1} \circ \dots \circ h_1(z)$ \cite[see, e.g.][for a comprehensive review]{Papamakarios2021}.
Other methods use a single flexible transformation like sum-of-squares polynomials~\citep{Jaini2019} or splines~\citep{Durkan2019}.
\Glspl{CTM} also focus on single transformation function, which, however, can be easily interpreted~\citep[see, e.g.,][]{Hothorn2018}.
This idea has also recently been combined with neural-networks~\citep{Baumann2021,Sick2021,Arpogaus2023,Kook2024}.

A fundamental difference between \glspl{NF} and \glspl{CTM} is the definition of the transformation direction~\citep{Kook2024}:
\glspl{NF} transform a simple base density $f_Z(z)$ into a more complex target density $f_Y(y)$ through a series of transformations.
\glspl{CTM} do it vice-versa, using only a single transformation function, often utilizing flexible Bernstein polynomials (see~\Cref{sec:bernstein_poly} for details).

\glspl{CTM} allow interpretability through the use of \glspl{SAP}~\citep{Klein2022,Rugamer2023} (see \Cref{sec:sap,sec:interpretation_details}).
The black-box character of neural networks in \glspl{NF} hinders interpretability, and variants using decision trees~\citep{Papastefanopoulos2025} and graphical models~\citep{Wehenkel2021a} have been explored in recent years.

The previous concepts can be further extended to multivariate objects $\mathbf{y} = (y_1,\ldots,y_J)^\top$ and can also be applied conditional on $U$ features $\mathbf{x} = (x_1,\ldots,x_U)^\top$ as described in the following.

\subsection{Multivariate and Conditional Models}

To model a potentially multivariate conditional distribution $F_{\mathbf{Y}|\mathbf{X}=\mathbf{x}}$,
we define
$z_j=h(y_j, \bm{\theta}_j) \text{ with } \bm{\theta}_j=c_j(\mathbf{y},\mathbf{x})$
where $h$ is a bijective \emph{transformation function} for the element $y_j$ with parameters $\bm{\theta}_j$ obtained from the \emph{conditioner} $c_j$~\citep{Papamakarios2021}.
The latter can be an arbitrarily complex function that depends on the features $\mathbf{x}$ and, in the multivariate case, on some variables of the response $\mathbf{y}$.
The limiting factor here is the computational tractability of the Jacobian in \Cref{eq:cov}.

\paragraph{Multivariate Conditional Transformation Models} To make computations tractable and provide a model that is better to interpret, \glspl{MCTM} define a triangular transformation $H(\bm{y},\bm\Theta(\bm{x}))=\left(h_1(y_1|\bm{\theta}_1),\ldots,h_J(y_J|y_1,\ldots,y_{J-1},\bm{\theta}_1,\dots,\bm{\theta}_J)\right)^\top$.
This Tranformation is characterized by a set of linearly combined marginal basis transformations $\tilde{h}_j(y_j|\bm{\theta}_j)$, scaled with a $(J \times J)$ triangular coefficient matrix $\bm{\Lambda}$ to encode structural information: $h_j(y_j|y_1,\ldots,y_{j-1},\bm{\theta}_1,\dots,\bm{\theta}_j) = \lambda_{j,1} \tilde{h}_{1}(y_1|\bm{\theta}_1)+\ldots+\lambda_{j,j-1} \tilde{h}_{j-1}(y_{j-1}|\bm{\theta}_{j-1}) + \tilde{h}_j(y_j|\bm{\theta}_j)$.
The function $\bm\Theta: \bm{x}\mapsto(\bm\Lambda,\bm\theta_1,\ldots,\bm\theta_J)$ accounts for feature effects on the parameters~\citep{Klein2022}.
Hence,~\glspl{MCTM} can only model linear dependencies between the response variables $y_j$, making them an inadequate choice for complex joint distributions as shown in \Cref{fig:mctmfail}.
This limitation motivates the development of more flexible approaches, such as the hybrid model proposed in this paper.

\paragraph{Normalizing Flows} In the~\gls{NF} literature, various approaches are used to deal with computations involving the aforementioned Jacobian. Among these, autoregressive models are one of the most prevalent ones. Due to the autoregressive structure, these \glspl{NF} yield a triangular Jacobian and thus the determinant simplifies to the product of the Jacobian's diagonal entries~\citep{Papamakarios2021,Kobyzev2021}.
These models factorize a multivariate distribution based on the chain rule of probability. They are typically implemented by conditioning the transformation $h_j$ of the $j$th element of $\mathbf{y}$ on all previous elements $\mathbf{y}_{<j} = (y_1, \dots, y_{j - 1})$ or a subset of those.
This can be seen as a non-linear generalization of the triangular coefficient matrix $\Lambda$ used in~\glspl{MCTM}~\citep{Kobyzev2021}.
There are two widely adopted neural-network architectures for implementing an autoregressive conditioner $c_j$:~\glspl{MAF}~\citep{Papamakarios2018} and \glspl{CF}~\citep{Dinh2017}.

\glspl{MAF} generalize this idea by implementing the conditioner with neural networks that incorporate autoregressive constraints through parameter masking, inspired by the \gls{MADE} architecture \citep{Germain2015}.
The main advantage is that the parameters can be obtained in one neural network pass $\left(\bm{\theta}_1,\ldots,\bm{\theta}_J\right)^\top=\left(c_1(),c_2(y_1),\ldots,c_J(\mathbf{y}_{<j})\right)^\top$. Further, if $h$ and the conditioning network are expressive enough, they remain universal approximators to transform between any two distributions~\citep{Papamakarios2021}.
In practice, however, the results highly depend on the ordering of the input variables.
Because some orderings can be extremely difficult to learn, multiple~\glspl{MAF} layers are often stacked with permutations between.


\glspl{CF} can provide fast sampling and density evaluation by splitting the response vector $\mathbf{y}$ into two subsets $$\mathbf{y}=(\underbrace{y_1,\ldots,y_j}_{\mathbf{y}_{\leq j}},\underbrace{y_{j+1},\ldots,y_{J}}_{\mathbf{y}_{> j}})^\top$$ and then applying the transformation $h$ only to one subset conditioned on the other.
Typically, these subsets are chosen to contain half of the variables and the result of the transformation on the first subset is then permuted and another transformation is applied to the remaining subset: $\mathbf{z}=\left(h(\mathbf{y}_{\leq d},c_1(\mathbf{y}_{> d},\mathbf{x}), h(\mathbf{y}_{> d},c_2(\mathbf{y}_{\leq d}, \mathbf{x})\right)^\top$.

\paragraph{Graphical Transformation Models} are an extension of \glspl{MCTM}, closely related to the methodology proposed in this paper.
\citet{Herp2025} replaced the Gaussian copula by parameterizing the $\lambda_{i,j}$ entries of $\bm{\Lambda}$ using B-splines, dependent on the previous variables, resulting in an additive \gls{CF}: $h_j(y_j|y_1,\ldots,y_{j-1},\bm{\theta}_1,\dots,\bm{\theta}_j) = \lambda_{j,1}\left(\tilde{h}_{1}(y_1|\bm{\theta}_1)\right)+\ldots+\lambda_{j,j-1}\left(\tilde{h}_{j-1}(y_{j-1}|\bm{\theta}_{j-1})\right) + \tilde{h}_j(y_j|\bm{\theta}_j)$.
To enhance expressiveness of the dependencies while preserving marginals, multiple \glspl{CF} are chained via permutations.
P-spline penalties mediate between the flexible \glspl{CF} and the baseline MCTM.
This framework improves interpretability of nonlinear conditional dependencies through local conditional pseudo-correlations and enables sparse undirected graphical models via a LASSO penalty promoting conditional independence.
However, it offers less flexibility as a \gls{NF} for decorrelation than the approach proposed here and has not yet been extended to the conditional setting.

Further background is given in \Cref{sec:mctm} and \ref{sec:ar_models}.

\section{Hybrid Conditional Masked Autoregressive Bernstein Flows}
\label{sec:method}

To bridge the gap between the limited expressiveness of~\glspl{MCTM} and the black-box nature of general~\glspl{NF}, we propose a combined approach that leverages the strengths of both methodologies.
This hybrid approach allows us to capture complex dependencies in the joint distribution while retaining interpretability for the marginal distributions.

\subsection{Model Specification}

Let $\mathbf{y} = (y_1, \dots, y_J)^\top \in \mathbb{R}^J$ be a $J$-dimensional response vector and $\mathbf{x} = (x_1, \dots, x_U)^\top \in \mathbb{R}^U$ be a vector of $U$ features.
Our goal is to learn the conditional joint density $f_{\mathbf{Y} | \mathbf{X}}(\mathbf{y} | \mathbf{x})$.

\paragraph{Step 1: Modeling Marginal Distributions}

\begin{figure}[t]
  \centering%
  \begin{subfigure}[b]{.45\linewidth}%
    \includegraphics[width=\linewidth]{moons_y.png}%
  \end{subfigure}
  \tikzmark{moons-1}%
  \hfill %
  \tikzmark{moons-h1}%
  \begin{subfigure}[b]{.45\linewidth}%
    \includegraphics[width=\linewidth]{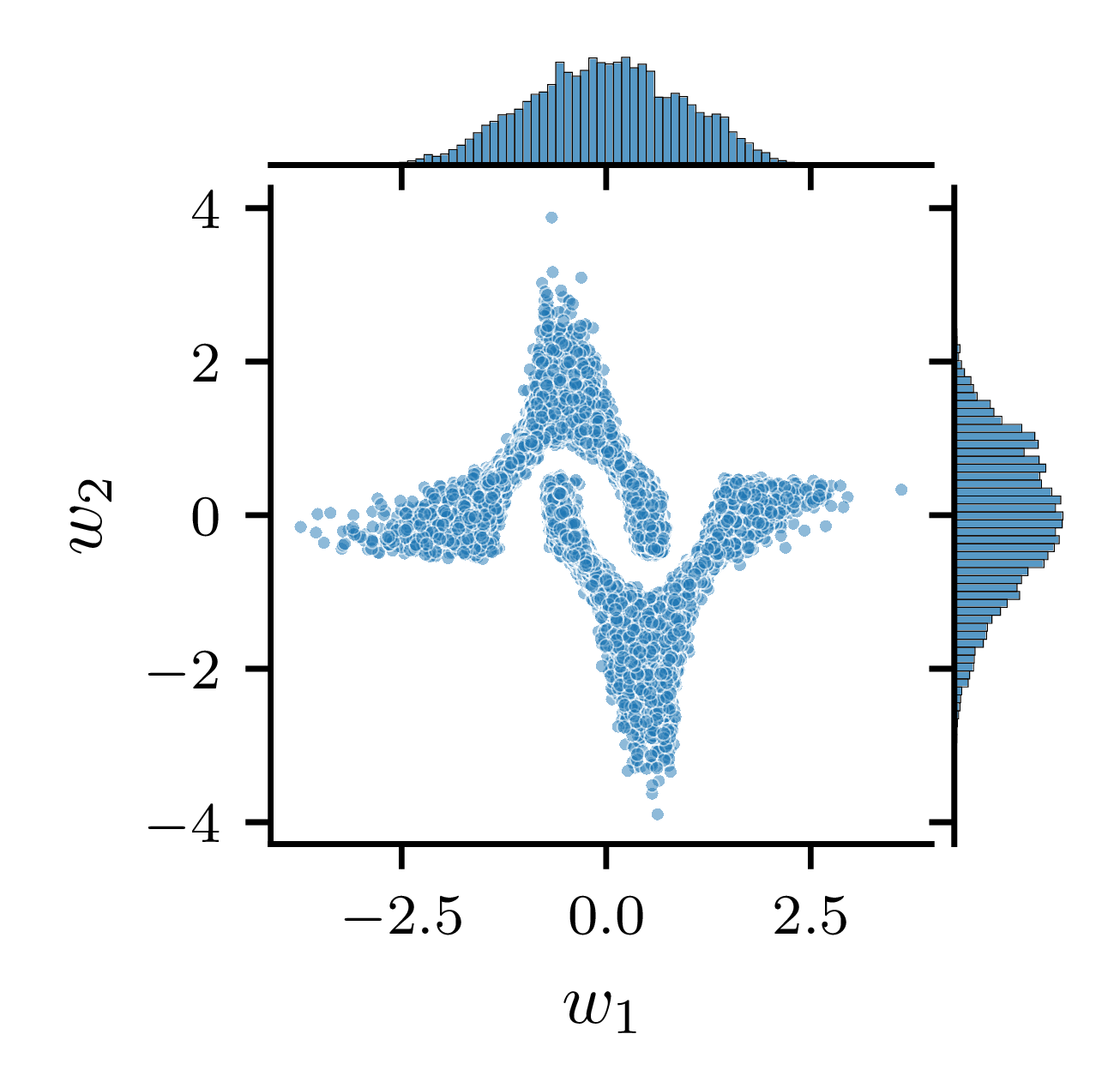}%
  \end{subfigure}
  \caption{The first transformation $H_1$ maps the marginals to the base distribution $F_Z$.}\label{fig:h1}%
  \tikz[remember picture, overlay, color=mpl_blue]{
    \draw[-latex, line width=1pt]
    ([yshift=.1\linewidth, xshift=-.03\linewidth]pic cs:moons-1)
    --([yshift=.1\linewidth, xshift=.05\linewidth]pic cs:moons-h1)
    node[midway,below] () {$H_1(\mathbf{y}|\mathbf{x})$};
  }%
  \tikz[remember picture, overlay, color=mpl_blue]{
    \draw[-latex, line width=1pt]
    ([yshift=.35\linewidth, xshift=.05\linewidth]pic cs:moons-h1)
    --([yshift=.35\linewidth, xshift=-.03\linewidth]pic cs:moons-1)
    node[midway,above] () {$H^{-1}_1(\mathbf{w}|\mathbf{x})$};
  }%
  \vspace{-12pt}
\end{figure}

First, we model the marginal distributions of each response variable $Y_j$ given $\mathbf{X}$ using a transformation model:
\begin{equation*}
  \begin{aligned}
    H_1(\mathbf{y},\mathbf{\Theta}\left(\mathbf{x}\right))
     & = \left(h_1(y_1,\bm{\theta}_{1,\mathbf{x}}),\ldots,h_1(y_J,\bm{\theta}_{J,\mathbf{x}})\right)^\top = \\
     & =\mathbf{w} = (w_1,\ldots,w_J)^\top,
  \end{aligned}
\end{equation*}
where each element of $\mathbf{W} = (W_1, \dots, W_J)^\top$ follows the base distribution $F_Z$ (\Cref{fig:h1}).
%

Specifically, for each $j = 1, \dots, J$, we assume:
\begin{equation*}
  \mathbb{P}(Y_j \leq y_j | \mathbf{X} = \mathbf{x}) = F_{Y_j | \mathbf{X}}(y_j | \mathbf{x}) = F_Z(h_1(y_j, \bm{\theta}_{j,\mathbf{x}})),
\end{equation*}
where $F_Z$ is a known base distribution
and $h_1(\cdot)\colon \mathbb{R} \rightarrow \mathbb{R}$ is a strictly monotonic marginal transformation function, with potentially feature-dependent parameters $\bm{\theta}_{j,\mathbf{x}}$ obtained form the conditioning function $\mathbf{\Theta}\left(\mathbf{x}\right)$.

For interpretability, a (shifted) Bernstein polynomial can be used:
\begin{equation*} 
  \begin{split}
    h_1(y_j, \bm{\theta}_{j,\mathbf{x}}) & =  \bm\alpha_j(y_j)^\top \bm{\vartheta}_j + \beta_{j,\mathbf{x}},           \\
    \bm\alpha_j(y_j)^\top \bm{\vartheta}_j  & =  \frac{1}{M+1} \sum_{i = 0}^M \text{Be}_i(\tilde{y}_j) \vartheta_{ji},
  \end{split}
\end{equation*}
where $\text{Be}_m(\tilde{y}_j)$ is the density of a Beta distribution with parameters $i+1$ and $M-i+1$ evaluated at the normalized response $\tilde{y}_j = (y_j - l_j)/(u_j - l_j) \in [0, 1]$, with $u_j > l_j$ defining the support of $Y_j$\footnote{The transformation is generally unbounded, as we apply linear extrapolation outside the bound of the polynomial, as all our experiments show. However, other transformations may be considered if expressiveness in the outer tails is a requirement}.
The vector $\bm{\vartheta}_j = (\vartheta_{j0}, \dots, \vartheta_{jM})^\top$ contains the Bernstein coefficients, which are constrained to be increasing for monotonicity (see~\Cref{sec:bernstein_poly} for details).
The shift term $\beta_{j,\mathbf{x}}$ allows the marginal distribution of $Y_j$ to vary with the features.
Instead of a feature-dependent shift term, we can also let $\bm{\vartheta}_j$ depend on $\mathbf{x}$ to change the shape of the transformation $h_1$ (or combine feature-dependent shift with feature-dependent $\bm{\vartheta}_j$s).

\paragraph{Step 2: Modeling Dependency Structures}

We model the dependencies between elements of $\mathbf{W}$ using an autoregressive flow $H_2(\mathbf{w},\mathbf{\Psi}(\mathbf{w}, \mathbf{x})): \mathbf{W}\to\mathbf{Z}, \mathbf{w}\mapsto\mathbf{z}$:
\begin{equation*}
  \begin{split}
    z_{1} & = h_2(w_1 | \bm{\psi}_{1, \mathbf{x}}),                                         \\
    z_{j} & = h_2(w_j | \bm{\psi}_{j, \mathbf{w}_{<j}, \mathbf{x}}), \quad j = 2, \dots, J,
  \end{split}
\end{equation*}
where $h_2(\cdot)$ is an increasing transformation function, with parameters $\bm{\psi}_{j, \mathbf{w}_{<j}, \mathbf{x}}$ depending on previous elements of $\mathbf{w}$ and features $\mathbf{x}$.
\begin{figure}[t]
  \centering%
  \begin{subfigure}[b]{.45\linewidth}%
    \includegraphics[width=\linewidth]{moons_w.png}%
  \end{subfigure}
  \tikzmark{moons-2}%
  \hfill %
  \tikzmark{moons-h2}%
  \begin{subfigure}[b]{.45\linewidth}%
    \includegraphics[width=\linewidth]{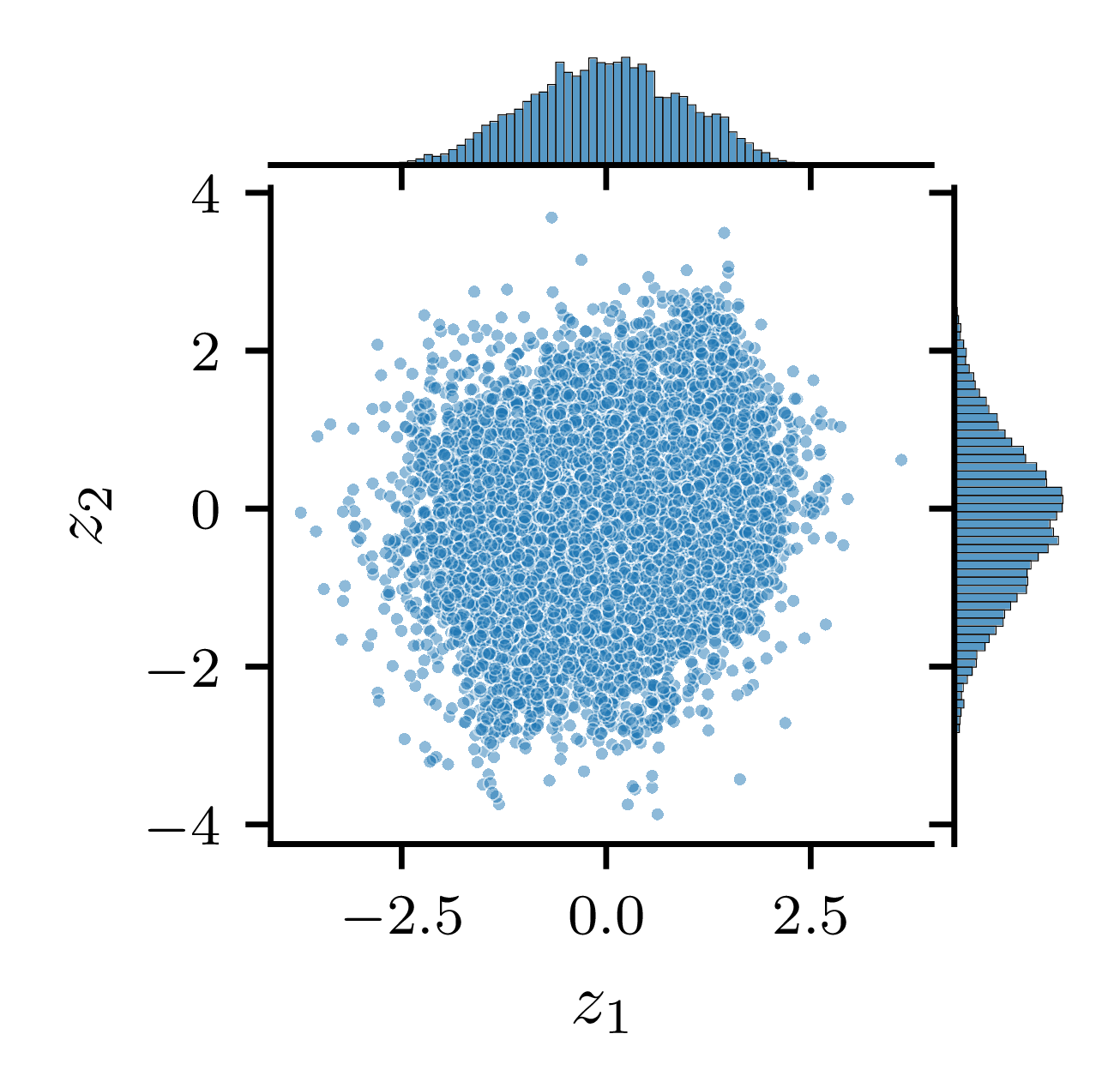}%
  \end{subfigure}
  \caption{The second transformation $H_2$ removes the dependency structure.}\label{fig:h2}%
  \tikz[remember picture, overlay, color=mpl_blue]{
    \draw[-latex, line width=1pt]
    ([yshift=.1\linewidth, xshift=-.03\linewidth]pic cs:moons-2)
    --([yshift=.1\linewidth, xshift=.05\linewidth]pic cs:moons-h2)
    node[midway,below] () {$H_2(\mathbf{w}|\mathbf{x})$};
  }%
  \tikz[remember picture, overlay, color=mpl_blue]{
    \draw[-latex, line width=1pt]
    ([yshift=.35\linewidth, xshift=.05\linewidth]pic cs:moons-h2)
    --([yshift=.35\linewidth, xshift=-.03\linewidth]pic cs:moons-2)
    node[midway,above] () {$H^{-1}_2(\mathbf{z}|\mathbf{x})$};
  }%
\end{figure}
For $h_2$, flexible transformation functions like Bernstein polynomials or splines can be used.

We combine ideas from \glspl{CF} and \glspl{MAF} to construct the conditioner.
Similar to \glspl{CF}, we do not apply any transformation to $w_1$ but pass it as a conditional input to the masked neural network that estimates the parameters of the subsequent transformations $\bm{\Psi}(\mathbf{w},\mathbf{x})=\left(\bm{\psi}_{\mathbf{x}},\bm{\psi}_{\mathbf{w}_{<1}, \mathbf{x}},\ldots,\bm{\psi}_{\mathbf{w}_{<J}, \mathbf{x}} \right)^\top$.
Binary masking, similar to \gls{MADE}, ensures that the $j$th parameters do not depend on $\mathbf{z}_{1,\ge j}$.
Autoregressive transformations can be stacked if a single transformation is not expressive enough.
Through this autoregressive transformation, we aim to ``de-correlate'' the elements of $\mathbf{W}$ such that $\mathbf{Z}$ follows a multivariate standard normal distribution $\mathcal{N}_{J}(\mathbf{0},\mathbf{I})$.

The joint density of the original response vector $\mathbf{Y}$ is then:\footnote{We use the short notation $H_1\left(\mathbf{y},\mathbf{\Theta}\left(\mathbf{x}\right)\right) = H_1(\mathbf{y}|\mathbf{x})$ and $H_2\left(\mathbf{w}, \mathbf{\Psi}\left(\mathbf{w}, \mathbf{x} \right) \right) = H_2(\mathbf{w}|\mathbf{x})$.}
\begin{equation*}
  \begin{aligned}
    f_{\mathbf{Y} | \mathbf{X}}\left(\mathbf{y} | \mathbf{x}\right)
     & = f_{Z}\left(H\left(\mathbf{y} | \mathbf{x} \right) \right)
    \left| \nabla H\left(\mathbf{y} | \mathbf{x} \right) \right|             \\
     & = f_{Z}\left(
    H_2\left(
      H_1\left(\mathbf{y}|\mathbf{x}\right) | \mathbf{x}
      \right)
    \right)                                                                  \\
     & \quad  \cdot \left|
    \nabla H_2\left(
    H_1\left(\mathbf{y}|\mathbf{x}\right)
    |\mathbf{x}
    \right) \nabla H_1\left(\mathbf{y}|\mathbf{x}\right)
    \right|.
  \end{aligned}
\end{equation*}

\paragraph{Model Training and Inference}

We train the hybrid model by minimizing the conditional negative log-likelihood of the data $\mathcal{D}$,
\begin{equation*}
  \begin{aligned}
   \nll(\bm \omega) &= -\sum_{(\mathbf{y}, \mathbf{x})\in\mathcal{D}}
    \log f_{\mathbf{Y} | \mathbf{X}}\left(\mathbf{y} | \mathbf{x}\right)\\
                       &= 
    -\sum_{(\mathbf{y}, \mathbf{x})\in\mathcal{D}}
    \log f_Z\left(H\left(\mathbf{y} | \mathbf{x}, \bm{\omega} \right) \right)
    \left| \nabla H\left(\mathbf{y} | \mathbf{x}, \bm{\omega} \right) \right|,
  \end{aligned}
\end{equation*}
to optimize the weights $\bm{\omega}$ of the conditioning functions $\bm{\Theta}_x$ and $\bm{\Psi}_x$ parameterizing the normalizing flow $H\left(\mathbf{y} | \mathbf{x}, \bm{\omega} \right) = H_2 \circ H_1(\mathbf{y} | \mathbf{x}, \bm{\omega})$.

This involves optimizing the parameters of the marginal transformation functions $H_1$ and the parameters of the autoregressive flow $H_2$, which are controlled by masked neural networks. Optimization can be done using any suitable gradient-based optimization algorithm, such as Adam~\citep{Kingma2017a}.
Once the model is trained, we can sample from the joint distribution of $\mathbf{Y} | \mathbf{X}$ by:
\begin{enumerate}
  \item Sample $\mathbf{z}$ from the base distribution $F_Z$.
  \item Apply the inverse autoregressive flow to obtain $\mathbf{w} = H_2^{-1}(\mathbf{z}|\mathbf{x})$.
  \item Apply the inverse marginal transformation to obtain $\mathbf{y} = H_1^{-1}(\mathbf{w}| \mathbf{x})$.
\end{enumerate}
In short: $\mathbf{y}=H_1^{-1}(H_2^{-1}(\mathbf{z}|\mathbf{x})|\mathbf{x})$ with $\mathbf{z} \sim F_Z$.

\subsection{Interpretability}
Our hybrid approach combines \gls{CTM}'s interpretability with the flexibility of autoregressive \glspl{NF}:
The marginal transformation step ($H_1$) may use shifted Bernstein polynomials, which allows quantifying feature effects on the marginal distribution, similar to coefficients in linear models or \glspl{GAM} \citep{Wood2017}.
For instance, with a logistic base distribution, linear effect coefficients in the \gls{SAP} can be interpreted as linear changes on the level of log-odds ratios (see \Cref{sec:sap,sec:interpretation_details} for more details).
%
Our model prioritizes marginal interpretability while accepting a trade-off for the dependence structure to gain flexibility via the masked autoregressive flow in step two ($H_2$).
It is suitable for scenarios requiring understanding individual feature effects while modeling complex relationships between response variables.
Understanding how features affect the dependence structure in this step is challenging and presents another open research question.

\subsection{Relation to Copula Methods}
\label{sec:copula}


\begin{figure*}[t]%
\centering 
  \begin{subfigure}[t]{0.3\linewidth}
    \centering
    \includegraphics[width=\linewidth]{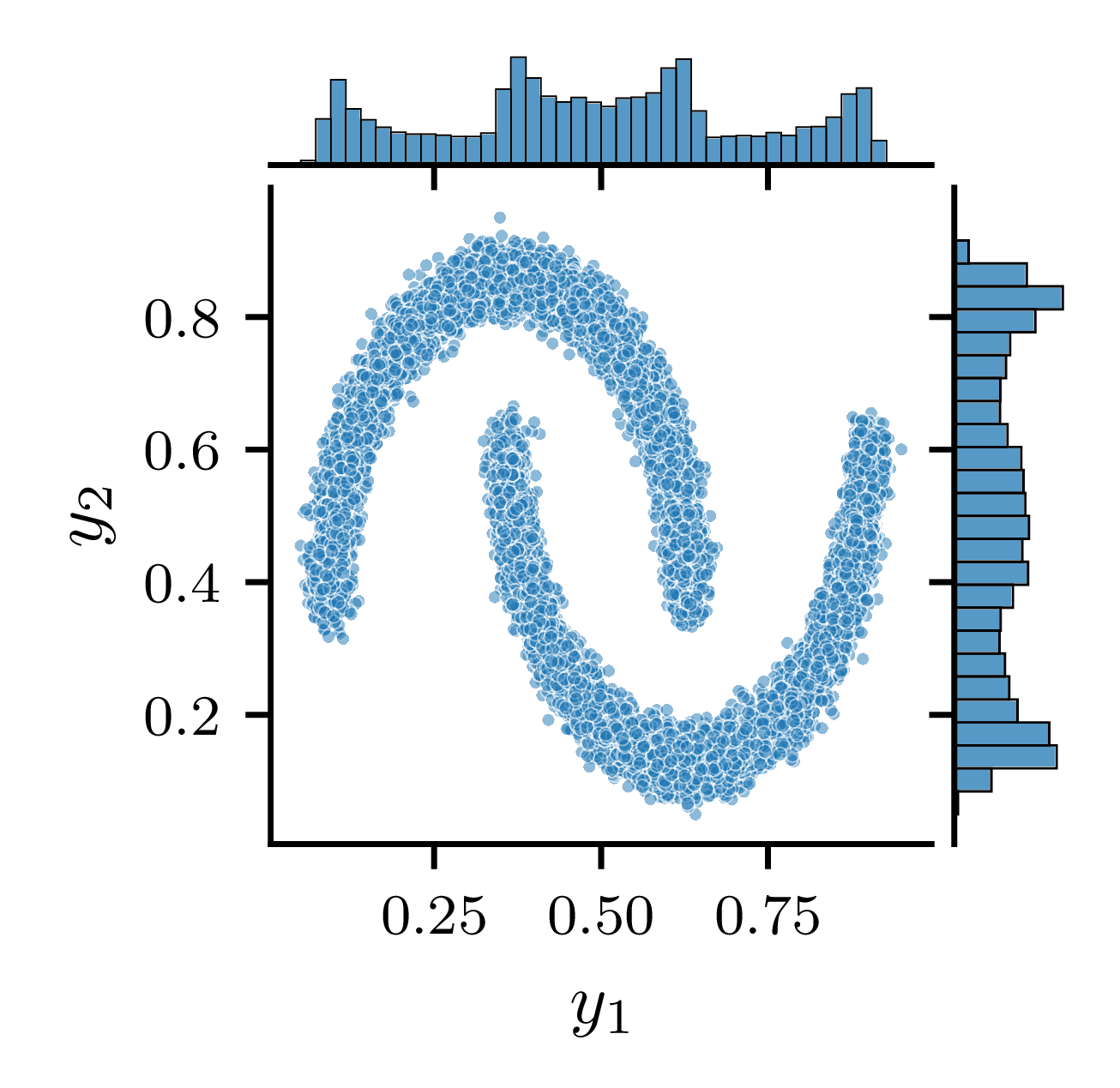}
    \caption{Original Data}
  \end{subfigure}
  \hfil%
  \begin{subfigure}[t]{0.3\linewidth}
    \centering
    \includegraphics[width=\linewidth]{moons_w.png}
    \caption{Normalized Marginals}
  \end{subfigure}
  \hfil%
  \begin{subfigure}[t]{0.3\linewidth}
    \centering
    \includegraphics[width=\linewidth]{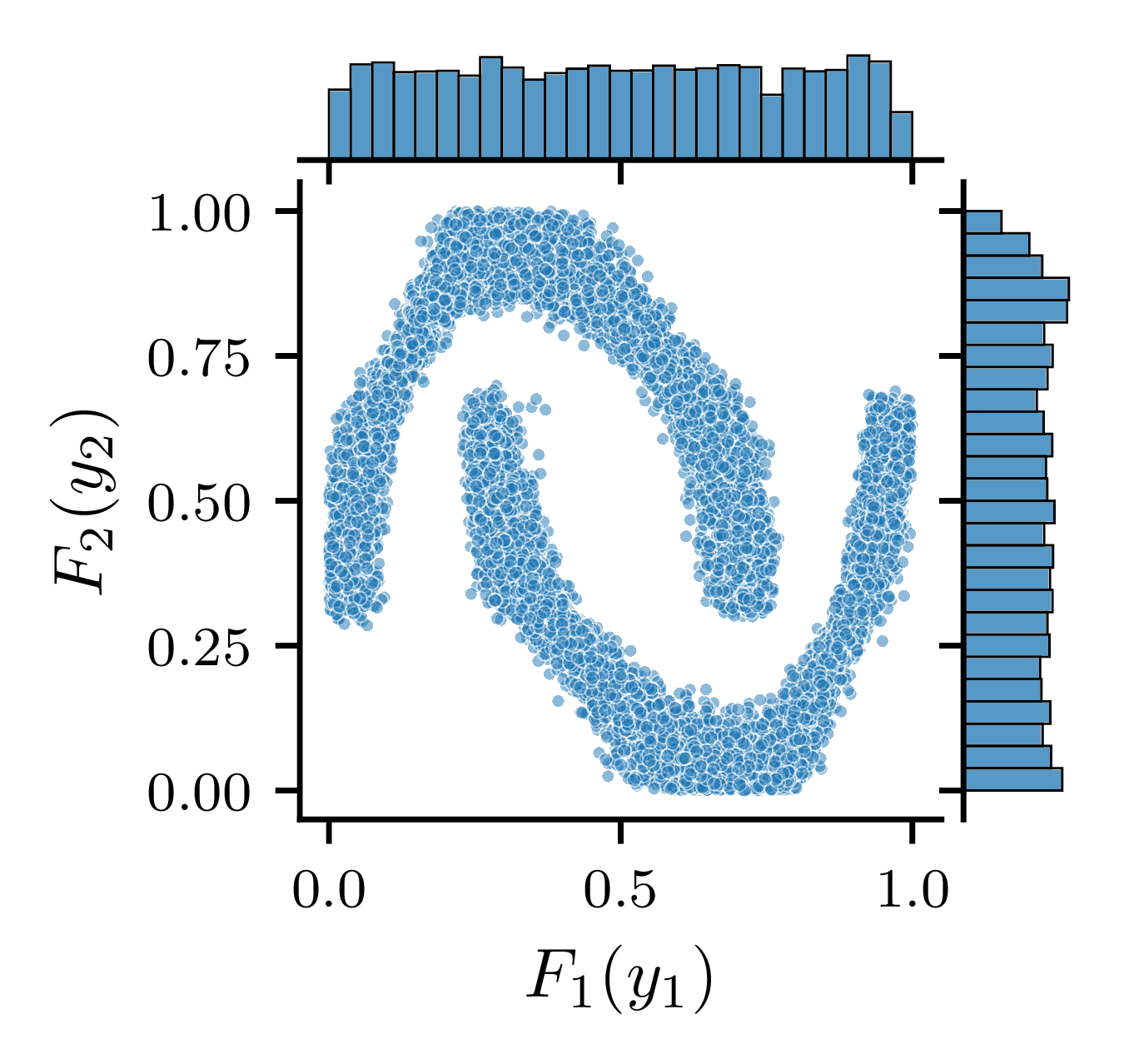}
    \caption{Uniform Marginals}
  \end{subfigure}
  \caption{Illustration of the hybrid approach on the Moons dataset.
    (a) The original data exhibits a non-linear dependency structure.
    (b) After applying $H_1$, the marginal distributions follow a normal distribution.
    (c) The autoregressive flow $H_2$ further transforms the data to obtain approximately independent uniform marginals, implicitly modeling the copula function.}
  \label{fig:copula_illustration}
\end{figure*}

Copulas model multivariate distributions by separating marginal distributions from the dependence structure.
Sklar's Theorem states that a copula function can express any \gls{CDF}. 
Let $F_{\mathbf{Y} | \mathbf{X}}(\mathbf{y} | \mathbf{x})$ be the joint \gls{CDF} of the response vector $\mathbf{Y} = (Y_1, \dots, Y_J)^\top$ given features $\mathbf{X}$.
Sklar's theorem implies a copula function $C(u_1, \dots, u_J | \mathbf{x})$ such that:
\begin{align*}
  &F_{\mathbf{Y} | \mathbf{X}}(y_1, \dots, y_J | \mathbf{x})\\
  &= C(F_{Y_1 | \mathbf{X}}(y_1 | \mathbf{x}), \dots, F_{Y_J | \mathbf{X}}(y_J | \mathbf{x}) | \mathbf{x}),
\end{align*}
where $u_j = F_{Y_j | \mathbf{X}}(y_j | \mathbf{x})$ are uniform marginal \glspl{CDF}.
The copula $C$ is a multivariate \gls{CDF} on $[0,1]^J$ with uniform marginals.
Our hybrid approach directly relates to this.
The first step ($H_1$) models marginals $F_{Y_j|\mathbf{X}}$ and transforms them to the base distribution $F_Z$.
Applying the \gls{PIT}, $u_j = F_Z(z_{1j})$, yields uniform marginals.

The copula density $c$ of $\bm{u}=(u_1, \dots, u_J)^\top$ is the ratio of the joint density and the product of the marginal densities:
\begin{equation*}
  c(\bm{u} | \mathbf{x}) = \frac{f_{\mathbf{Y} | \mathbf{X}}(F^{-1}_{Y_1 | \mathbf{X}}(u_1 | \mathbf{x}), \dots, F^{-1}_{Y_J | \mathbf{X}}(u_J | \mathbf{x}) | \mathbf{x})}{\prod_{j=1}^J f_{Y_j | \mathbf{X}}(F^{-1}_{Y_j | \mathbf{X}}(u_j | \mathbf{x}) | \mathbf{x})}.
\end{equation*}

\section{Numerical Experiments}
\label{sec:experiments}

Next, we evaluate our method using simulated and real-world datasets.
All models are implemented using \texttt{TensorFlow} (v2.15.1) and \texttt{TensorFlow Probability} (v0.23.0) and trained using Adam~\citep{Tensorflow,TensorflowProbability,Kingma2017a}.
In our experiments, hyperparameters were chosen based on our previous experience and settings suggested in related works.
However, we did some automated hyperparameter tuning (using Optuna) on some models applied to the simulated 2D data with performance monitored on a held-out validation set. This was mainly done to understand the influence of certain parameters, but it was not necessary to obtain a good model for each data set.
We found that high-order Bernstein polynomials are needed for some datasets, suggesting the fixed knot placement of Bernstein polynomials could limit their flexibility\cite[see][for an ablation to the sensitivity of the Bernstein order]{Hothorn2018}.

Splines, on the other hand, might be more suitable for representing highly nonlinear or discontinuous functions, leading to a better fit on certain datasets.
All hyperparameters used to generate the results presented are documented in the supplementary material and in the GitHub repository\footnote{\url{https://github.com/MArpogaus/hybrid_flows}}.
This includes the code to perform the hyperparameter tuning along with the defined search spaces.

\subsection{Benchmark Datasets}

We evaluate our method on five common benchmark datasets POWER, GAS, HEPMASS, MINIBOONE, and BSDS300 (see Appendix \Cref{sec:benchmark_data}).
We follow the preprocessing steps as in \citet{Papamakarios2018} and compare a \gls{MAF} and our proposed \gls{HMAF}.
Both models utilize masked autoregressive transformations.
Each \gls{MAF} layer uses a masked affine transformation, parameterized by a masked neural network followed by an invertible linear transformation (1x1 convolution) initialized with a random (non-trainable) permutation.
For the \gls{HMAF}, we use a marginal transformation step with Bernstein polynomials before the \glspl{MAF} layers.

For both \glspl{MAF} and \gls{HMAF}, we employ Rational Quadratic Splines (RQS) as the transformation functions within the \glspl{MAF} layers of $H_2$.  Hyperparameters, including the number of \glspl{MAF} layers, the number of hidden units in the \gls{MADE} networks, and the number of bins for the RQS transformations, were chosen based on the values reported in \citet{Durkan2019} and are detailed in the supplementary material and the GitHub repository.
All models were trained using the Adam optimizer with early stopping after $50$ epochs without improvements and a cosine learning rate schedule.

\begin{table*}[htb!]
  \caption{
    Test negative log-likelihood comparison against the state-of-the-art on real-world datasets (lower is better).\\
    Values are averaged over 20 trials, and their spread is reported as two standard deviations.
  }
  \label{tab:benchmark-nll}
\resizebox{0.99\textwidth}{!}{
    \begin{tabular}{l|llllll}
      \toprule
      model                          & dataset name    & bsds300              & gas                 & hepmass            & miniboone          & power              \\
      \midrule
      \multirow[c]{3}{*}{\gls{HMAF}} & test loss       & -153.663 $\pm$ 0.037 & -11.625 $\pm$ 0.209 & 18.103 $\pm$ 0.058 & 12.057 $\pm$ 0.092 & -0.527 $\pm$ 0.003 \\
                                     & train loss      & -165.274 $\pm$ 0.124 & -11.790 $\pm$ 0.230 & 17.856 $\pm$ 0.063 & 8.715 $\pm$ 0.415  & -0.573 $\pm$ 0.004 \\
                                     & validation loss & -168.769 $\pm$ 0.035 & -11.621 $\pm$ 0.210 & 18.123 $\pm$ 0.057 & 11.519 $\pm$ 0.070 & -0.537 $\pm$ 0.003 \\
      \midrule
      \multirow[c]{3}{*}{\gls{MAF}}  & test loss       & -155.057 $\pm$ 0.065 & -11.781 $\pm$ 0.032 & 18.090 $\pm$ 0.039 & 12.030 $\pm$ 0.073 & -0.541 $\pm$ 0.003 \\
                                     & train loss      & -167.694 $\pm$ 0.293 & -11.881 $\pm$ 0.032 & 17.868 $\pm$ 0.041 & 9.605 $\pm$ 0.048  & -0.606 $\pm$ 0.003 \\
                                     & validation loss & -170.188 $\pm$ 0.085 & -11.779 $\pm$ 0.031 & 18.102 $\pm$ 0.040 & 11.546 $\pm$ 0.048 & -0.551 $\pm$ 0.003 \\
      \bottomrule
    \end{tabular}
}
\end{table*}

\Cref{tab:benchmark-nll} presents the test log-likelihoods achieved by our models.
Overall, \gls{HMAF} demonstrates competitive performance compared to \glspl{MAF}.
Further investigation into more flexible marginal transformations could lead to even better performance and broader applicability of hybrid flow-based models.
See \Cref{sec:benchmark_data} for a visualization of individual model runs.

\subsection{Simulated Data}
\label{sec:simulated_data}

We start with the classical bivariate (i.e. $J=2$) \glspl{NF} datasets, \emph{moons} and \emph{circles}, exhibiting non-linear dependencies.
Each dataset has $16,384$ data points (we reserved 25\% for validation) generated using \texttt{scikit-learn}~\citep{scikit-learn}.
A binary feature $x$ based on spatial location is introduced.
For \emph{moons}, $x=1$ indicates the lower-right moon, and $x=0$ the upper-left.
For \emph{circles}, $x=1$ denotes the inner circle, and $x=0$ the outer.
This assesses the models' ability to capture $f(\mathbf{y}|x)$.
We compare the following models:
\begin{itemize}
  \item \textbf{\acrfull{MVN}} Assumes a conditional multivariate normal distribution parameterized by $x$ using a connected network (two layers, 16 units, ReLU activation).

  \item 
\textbf{\acrfull{MCTM}:} Employs Bernstein polynomials of order $M=300$ for marginal transformations and a triangular matrix $\Lambda$ for the dependency structure. In the conditional case, the marginal shift parameter $\beta_x$ and $\Lambda_x$ are obtained using Bernstein polynomials of order 6, as in \citet{Klein2022}.
%
%
        While this approach offers interpretability through \gls{SAP}, the linear dependence structure constrains its capacity.

  \item \textbf{\acrfull{CF} and \acrfull{MAF}:} These consist of two stacked coupling or masked autoregressive layers, parameterized by fully connected or masked neural networks, respectively (tree layers, 128 units, ReLU activation).
        Conditional models incorporate $x$ as input to an FCN whose output is added to the first layer's parameters. We use Bernstein polynomials of order $M=300$ (\gls{CF}~(B), \gls{MAF}~(B)) and quadratic splines with 32 bins (\gls{CF}~(S), \gls{MAF}~(S)) as transformation functions.

  \item \textbf{\acrfull{HCF}:} Combines elementwise Bernstein polynomials of order $M=300$ for marginals (similar to \glspl{MCTM}) with a single coupling layer to model dependencies.
        Again Bernstein polynomials (B) and quadratic splines (S) are compared as transformation functions in the coupling layer.
        As the data exhibits a very complex distribution, simple shift terms are not enough to model the feature effect in the marginals.
        Instead, class-specific coefficients are used to model the feature effect on the Bernstein coefficients.
\end{itemize}

All models use the Adam optimizer~\citep{Kingma2017a} with early stopping and cosine learning rate decay~\citep{Loshchilov2017}.
All chosen hyperparameters are reported in the supplementary material.

\newsavebox{\mycolumnbox}
\begin{figure}[htb!]
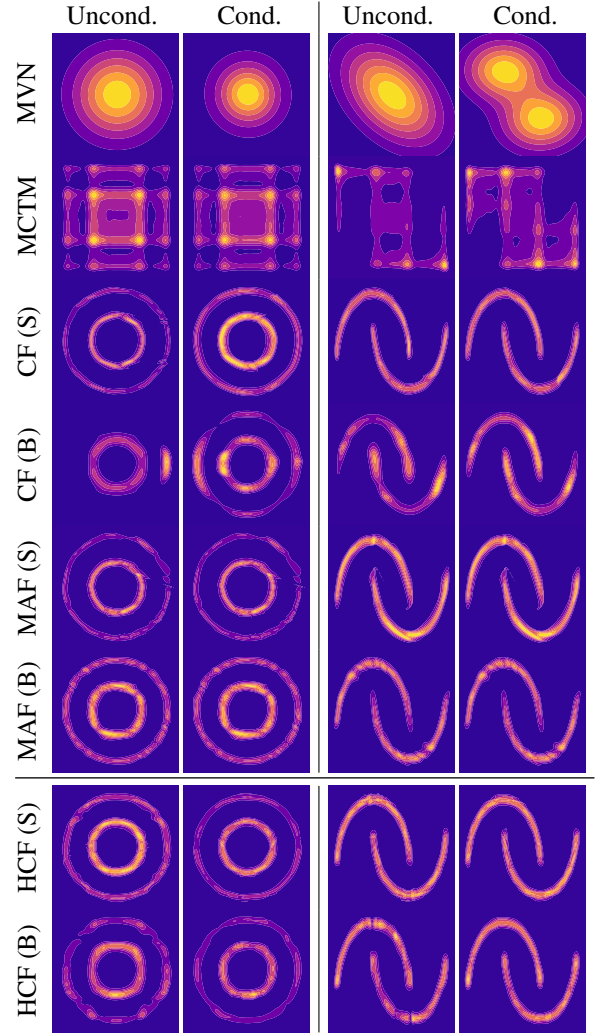

  \centering
  \setlength{\tabcolsep}{1pt}
  \begin{tabular}{>{\begin{lrbox}{\mycolumnbox}}%
               l%
               <{\end{lrbox}\rotatebox[origin=c]{90}{\textbf{\unhbox\mycolumnbox}}}%
      cc|cc}
            & Uncond.
            & Cond.
            & Uncond.
            & Cond.
    \\
    MVN     & \pdfplotcolumns{circles}{multivariate_normal_contour}
            & \pdfplotcolumns{moons}{multivariate_normal_contour}                           \\
    MCTM    & \pdfplotcolumns{circles}{multivariate_transformation_model_contour}
            & \pdfplotcolumns{moons}{multivariate_transformation_model_contour}             \\
    CF (S)  & \pdfplotcolumns{circles}{coupling_flow_quadratic_spline_contour}
            & \pdfplotcolumns{moons}{coupling_flow_quadratic_spline_contour}                \\
    CF (B)  & \pdfplotcolumns{circles}{coupling_flow_bernstein_poly_contour}
            & \pdfplotcolumns{moons}{coupling_flow_bernstein_poly_contour}                  \\
    MAF (S) & \pdfplotcolumns{circles}{masked_autoregressive_flow_quadratic_spline_contour}
            & \pdfplotcolumns{moons}{masked_autoregressive_flow_quadratic_spline_contour}   \\
    MAF (B) & \pdfplotcolumns{circles}{masked_autoregressive_flow_bernstein_poly_contour}
            & \pdfplotcolumns{moons}{masked_autoregressive_flow_bernstein_poly_contour}     \\
    \midrule
    HCF (S) & \pdfplotcolumns{circles}{hybrid_coupling_flow_quadratic_spline_contour}
            & \pdfplotcolumns{moons}{hybrid_coupling_flow_quadratic_spline_contour}         \\
    HCF (B) & \pdfplotcolumns{circles}{hybrid_coupling_flow_bernstein_poly_contour}
            & \pdfplotcolumns{moons}{hybrid_coupling_flow_bernstein_poly_contour}           \\
  \end{tabular}
  \caption{Estimated densities from different models fitted to simulated 2D data.
    Left two columns: Circles dataset, Right two columns: Moons dataset.
    Each cell shows the estimated density function, conditioned on $x$ (left) and not conditioned on $x$ (right).
    The conditional variable $x$ is 1 for the inner circle and the lower right moon, respectively.}
  \label{fig:simulated_data}
\end{figure}

\paragraph{Results}
\Cref{fig:simulated_data} visualizes the estimated densities for different models on both the \emph{circles} and \emph{moons} datasets. Each cell displays contour plots of the estimated densities, both conditional (left) and unconditional (right).
As expected, the \gls{MVN} and the \gls{MCTM} struggle with the non-linear dependencies.
The \gls{MVN} is restricted to elliptical distributions, while the \gls{MCTM}, despite capturing marginals well, fails to represent the complex joint distribution due to its linear dependence structure.
In contrast, \glspl{CF}, \glspl{MAF}, and \gls{HCF} successfully capture the datasets' non-linear shapes, demonstrating their ability to model complex distributions.
Interestingly, models using Bernstein polynomials (\gls{CF}~(B), \gls{MAF}~(B), \gls{HCF}~(B)) tend to produce slightly blurrier results, indicating a bias towards smoother distributions compared to the spline-based models.
Among the flexible models, quadratic spline transformations generally show a slight performance advantage over Bernstein polynomials for the \glspl{CF} and \glspl{MAF} architectures in terms of capturing fine-grained details of the data distribution, although the \gls{HCF} models seem to perform equally well with both transformations.

\Cref{tab:simulated_results} (in \Cref{sec:sim-nll}) provides a quantitative comparison, reporting the average test \gls{NLL} across 20 trails.
The results corroborate the visual observations from \Cref{fig:simulated_data}.
\gls{MVN} and \gls{MCTM} have significantly higher \gls{NLL} values, indicating their inadequacy for these complex distributions.
\glspl{CF} and \glspl{MAF}, especially with spline transformations, consistently achieve lower \gls{NLL}.
This difference underscores the importance of flexible dependency modeling.
The \gls{HCF} models 
perform well and demonstrate the effectiveness of the hybrid approach.
The differences of Bernstein polynomial models in \Cref{fig:simulated_data} are directly reflected in lower \gls{NLL} values, suggesting that these models lack certain flexibility to represent the data distribution.
Consistent with the visualizations, including feature information improves model performance across all models, leading to consistently lower \gls{NLL} values in the conditional setting.
This improvement highlights the models' ability to leverage the feature to better capture the conditional structure of the data.

\subsection{Malnutrition Data}

Finally, we evaluate our approach using a real-world dataset on childhood malnutrition in India, with three anthropometric indices (\texttt{stunting}, \texttt{wasting}, and \texttt{underweight}) as response variables.
The goal is to model the joint distribution conditional on the child's age (\texttt{cage}).
We follow the data preprocessing steps from \citet{Klein2022}.

We fit three models, all using Bernstein polynomials of order $M=6$ for the marginal transformations with a linear feature shift.
\begin{itemize}
  \item \gls{MCTM}: Identical to the model specification in \citet{Klein2022}, this model combines marginal transformations with a triangular matrix $\Lambda$.
  We capture the influence of \texttt{cage} on both the marginals (using linear shifts) and on the elements of $\Lambda$ using Bernstein polynomials of order $M=6$, i.e.\ the marginal shift is modeled as Bernstein polynomial $\bm{\beta}_x = \alpha(\text{cage})^\top \bm{\vartheta}$, where $\alpha$ represents the Bernstein basis, and $\bm{\vartheta}$ are the coefficients. 

  \item \acrfull{HMAF}: For our approach, the \glspl{MAF} for $H_2$ uses either Bernstein polynomials (B) or quadratic splines (S).
        We apply $H_2$ to $\mathbf{y}_{j>1}$, as $y_1$ is already normalized by $H_1$.
        Each \glspl{MAF} layer is parameterized by a masked neural network conditional on \texttt{cage}. To capture dependencies between $y_1$ and $\mathbf{y}_{j>1}$, an additional fully connected neural network is used to condition the output of the \gls{MADE} networks on $y_1$.
\end{itemize}

\paragraph{Results}\Cref{tab:malnutrition_results} presents average test \gls{NLL}.
\gls{HMAF} variants, especially \gls{HMAF}~(S), outperform \gls{MCTM}, indicating that the non-linear dependence modeling of the \glspl{MAF} is crucial.
Additionally, this is confirmed by \Cref{fig:malnutrition_samples} in \Cref{sec:malnutrition_samples} via scatter, pairwise density, and marginal density plots of the dataset as well as samples drawn from the three models.
The plots indicate that the \gls{MCTM} captures marginals well, but fails to model dependencies.
The \gls{HMAF} model shows a greater resemblance to the observed data in the pairwise density and scatter plots.

\begin{table}[htb]
  \centering
  \caption{Average test \gls{NLL} on the malnutrition data.
    Lower values indicate better performance. Values are averaged over 20 trials, and their spread is reported as two standard deviations.}
  \label{tab:malnutrition_results}
  \begin{tabular}{ll}
    \toprule
    model & test loss \\
    \midrule
    MCTM & 3.470 $\pm$ 0.026 \\
    HMAF (S) & 0.687 $\pm$ 0.206 \\
    HMAF (B) & 2.062 $\pm$ 0.038 \\
    \bottomrule
  \end{tabular}
\end{table}

\begin{figure}[htb!]
  \centering
  \includegraphics[width=\linewidth]{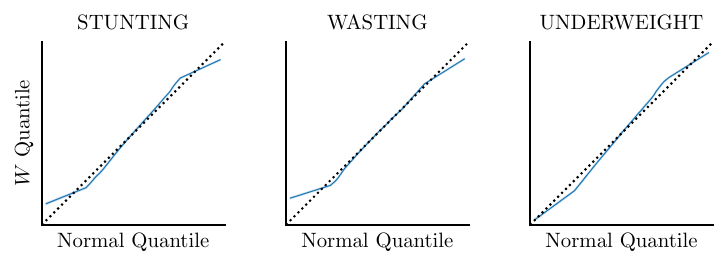}
  \caption{\gls{QQ} plots of transformed samples against a standard normal distribution.
    Deviations from the diagonal indicate non-normality.
    The solid line represents the mean, while the shaded area indicates the 95\% probability intervals obtained from 20 trials of randomly initialized models.}
  \label{fig:qq_w_base}
\end{figure}

\paragraph{Marginal distributions}
To assess whether the model captures the marginal distribution,  \citet{Hothorn2014} suggest using \gls{QQ} plots to verify $\mathbf{W}$ follows the base distribution $F_W=\mathcal{N}(0,1)$. \Cref{fig:qq_w_base} shows the empirical quantile function of marginally normalized samples $\mathbf{W}$ from the validation set plotted against the quantile function of a standard normal distribution evaluated at 200 equidistant probabilities. The results suggest a good fit of the transformation $H_1$, with slight underestimations of the marginals.

\begin{figure}[htb!]
    \centering
    \includegraphics[width=\linewidth]{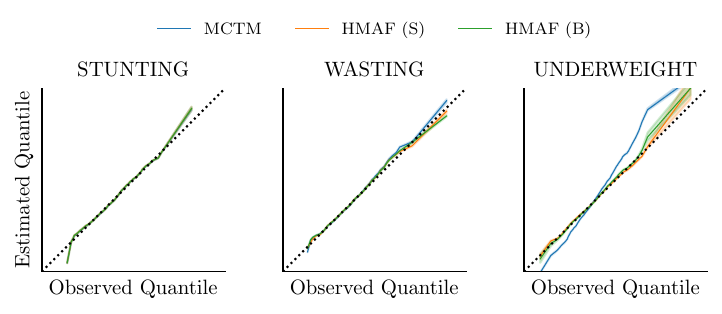}
    \caption{\gls{QQ} plot comparing empirical quantiles of the dataset with those generated by the three models.
    The solid lines represent the mean, while the shaded areas indicate the 95\% probability intervals obtained from 20 trials of randomly initialized models.}
    \label{fig:qq_data_samples}
\end{figure}

To assess the influence of $H_2$ on the marginals, we plotted the empirical quantiles of the dataset against an equal number of samples from all three models in \Cref{fig:qq_data_samples}.
All three models leave the normalized samples for \texttt{stunting} unchanged while introducing more deviations for \texttt{wasting} and \texttt{underweight}.
This deviation is especially large for the \gls{MCTM} models. The \gls{HMAF}, particularly the spline variants, generally perform well but struggle with upper tails.

\subsection{Interpretability}

A crucial aspect of our proposed hybrid approach is its ability to model marginal distributions in an interpretable manner while simultaneously capturing complex dependencies. Using the Malnutrition data, we demonstrate the model's interpretability in understanding learned relationships.

\begin{figure}[tb!]
  \centering
  \includegraphics[width=\columnwidth]{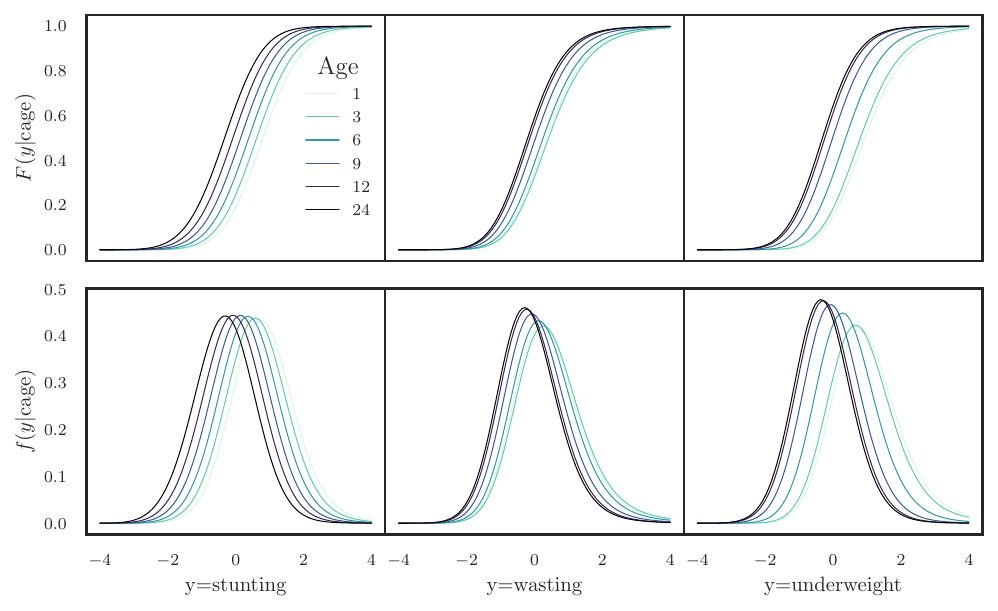}
  \caption{Comparison of marginal effect on the \glspl{PDF} and \glspl{CDF} of \texttt{stunting}, \texttt{wasting}, and \texttt{underweight} with respect to \texttt{cage}.}
  \label{fig:marginal_effects_comparison}
\end{figure}

\paragraph{Feature-induced Changes in Marginal Distribution}\Cref{fig:marginal_effects_comparison} shows the estimated marginal \glspl{CDF} $F(y_j|\text{cage})$ and \glspl{PDF} $f(y_j|\text{cage})$.
The plots indicate non-linear shifts towards lower values as the age of the child increases indicating a deteriorating nutrition status according to all three malnutrition indicators. This overall trend can be explained by the fact that most children are born with a close-to-normal nutritional status but the effects of resource-scarce environments become increasingly relevant as children age.

\begin{figure}[h!]
    \centering
    \includegraphics[width=\columnwidth]{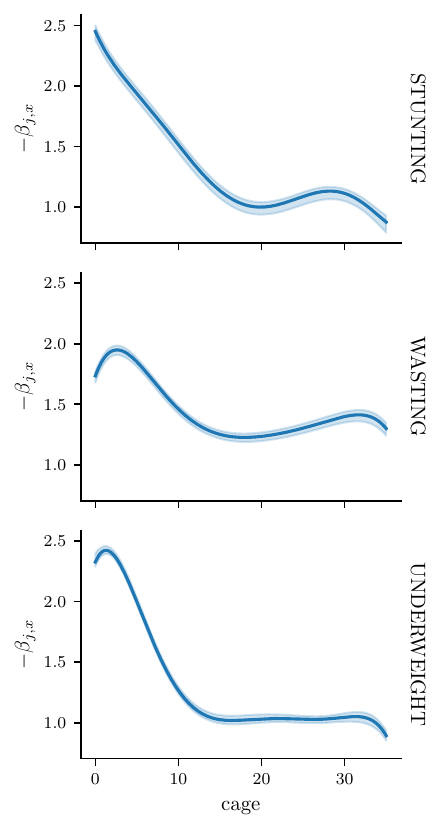}
    \caption{Inverse marginal shift $-\bm{\beta}_x$. The plot demonstrates the relationship between marginal shifts and \texttt{cage}.
    The solid line represents the mean and the shaded area the 95\% probability intervals over 20 trials of randomly initialized models.}
    \label{fig:marginal_shift}
\end{figure}

\paragraph{Inverse Marginal Shift Terms} A more nuanced interpretation can be achieved by depicting the inverse marginal shift terms $\bm{\beta}_x$ as in \Cref{fig:marginal_shift}.

The plot shows the complex non-linear change in the nutritional status with increasing age, but also highlights that the change is more pronounced for stunting and underweight. This coincides with the intuition that acute malnutrition (as measured by the stunting indicator) materializes more quickly than chronic malnutrition (as measured by the wasting indicator). Underweight represents a mixture of both acute and chronic malnutrition, which again fits with the estimated shift term.

\section{Discussion}
\label{sec:discussion}
We introduce a hybrid approach for density regression that combines the strengths of~\glspl{MCTM} and autoregressive~\glspl{NF}.
In the first step of our \gls{HMAF} approach, we model the marginal distributions using a transformation model with interpretable structured additive predictor.
This enables transparent modeling of feature effects on the marginal distribution.
In the second step, we use an autoregressive \gls{NF} to effectively capture complex, non-linear dependencies between the response variables while preserving the marginals.
Our results on simulated and real-world datasets showed that our method is competitive with state-of-the-art methods.

The hybrid approach offers 1) interpretability, by
  transparently modeling of feature effects using structured additive predictors, 2) flexibility, by
  capturing complex non-linear dependencies using an autoregressive flow, and 3) efficiency, by 
  offering fast computation of log-likelihoods and gradients due to the \gls{MADE} architecture.


\begin{acknowledgements} 
During the finalization of this paper, large language models (LLMs) were used to optimize language, and grammar, and restructure some parts of the document.

This research was funded by the Carl-Zeiss-Stiftung in the project ”DeepCarbPlanner” (grant no. P2021-08-007).

We thank the DACHS data analysis cluster, hosted at Hochschule Esslingen and co-funded by the MWK within the DFG's ,,Großgeräte der Länder'' program, for providing the computational resources necessary for this research\footnote{https://wiki.bwhpc.de/e/DACHS}.
\end{acknowledgements}


\printbibliography

\newpage

\onecolumn

\title{Hybrid Bernstein Normalizing Flows for Flexible Multivariate Density
Regression with Interpretable Marginal\\(Supplementary Material)}
\maketitle

\appendix

\section{Extended Background}
\label{sec:extended_background}

\subsection{Bernstein Polynomials}
\label{sec:bernstein_poly}
Bernstein polynomials of order $M$ are defined as
\begin{equation}
h(z) = \frac{1}{M+1}\sum_{i=0}^M \operatorname{Be}_i(z) \vartheta_i,
\end{equation}
where $\operatorname{Be}_i(z)$ is the density of a Beta distribution with parameters $i+1$ and $M-i+1$, and $\vartheta_0,\ldots,\vartheta_M$ are the Bernstein coefficients~\cite{Farouki2012}.
As $M$ increases, Bernstein polynomials become a \emph{universal approximator} for smooth functions in $[0,1]$ \cite{Farouki1988}.
In practice, $M \ge 10$ is often sufficient~\cite{Hothorn2018}.

Bernstein polynomials are defined for values of $z$ within the range $[0,1]$.  Outside this interval, linear extrapolation is performed.
To guarantee invertibility, transformation functions must be bijective, achieved by strict monotonicity. When using Bernstein polynomials, monotonicity is enforced by constraining the Bernstein coefficients $\vartheta_0,\ldots,\vartheta_M$ to be increasing. This is achieved by recursively applying a strictly positive function like $\softplus$ to an unconstrained vector $\tilde\vartheta_0,\ldots,\tilde\vartheta_M$, such that $\vartheta_0=\tilde\vartheta_0$ and $\vartheta_{k}=\vartheta_{k-1} + \softplus(\tilde\vartheta_k)$ for $k=1,\ldots,M$.
Parameter estimation can depend on initialization, as $\vartheta_0$ is directly derived from unconstrained parameters.
Monotonicity of $h$ is enforced by constraining the coefficients to be increasing, e.g.
by $\vartheta_0=\tilde\vartheta_0$ and $\vartheta_{k}=\vartheta_{k-1} + \softplus(\tilde\vartheta_k)$ for $k=1,\ldots,M$~\cite{Sick2021}.
The inverse transformation can be found using a root-finding algorithm~\cite{Chandrupatla1997}.

We require transformations to cover at least the range $[-3,3]$ (e.g., $\pm 3\sigma$ of a standard Gaussian). Since Bernstein polynomials' boundaries are defined by its first and last coefficients ($f(0)=\vartheta_{0}$ and $f(1)=\vartheta_{M}$), we determine these from unrestricted parameters $\tilde{\vartheta}_0$ and $\tilde{\vartheta}_{M+1}$ via $\vartheta_0 = -\softplus(\tilde{\vartheta}_0) - 3.0 \le -3$ and $\vartheta_M =  \softplus(\tilde{\vartheta}_{M+1}) + 3.0 \ge 3$.
To ensure $\sum_{k=1}^M{(\vartheta_k - \vartheta_{k-1})}=\vartheta_M - \vartheta_0 =: \Delta$, the remaining coefficients $\vartheta_k$ for $k=1,\ldots,M$ are calculated as:
\begin{equation}
\vartheta_k = \vartheta_{k-1} + \Delta\cdot\softmax\left(\left[\tilde\vartheta_1, \tilde\vartheta_3,\ldots, \tilde\vartheta_M\right]\right)_{k-1}
\end{equation}
Since $\Delta$ and all $\softmax$ components are non-negative, $\vartheta_k - \vartheta_{k-1} \ge 0$, ensuring monotonicity.

\subsection{Maximum Likelihood Estimation for Parameter Inference}
Parameter estimation usually involves minimizing a divergence between the true distribution $p_{y}(y)$ and the transformation model $p_y(\mathbf{y}|\theta)$ regarding the model's parameters $\theta$ \citep{Papamakarios2021}.
Since $p_y(y)$ is unknown, we minimize the negative log-likelihood of the empirical distribution $p_{\mathcal{D}}$ obtained from a finite set $\mathcal{D}$ of $N$ i.i.d. observations $\mathbb{Y} = \{\mathbf{y}_1,\ldots,\mathbf{y}_m\}$:
\begin{equation}\label{eq:nll}
\nll = - \sum_{y \in \mathcal{D}}\log \left(p_y(y;\theta)\right)
\end{equation}
which is equivalent to minimizing the KL divergence between $p_\mathcal{D}$ and the flow-based model $p_y (y; \theta)$ \citep{Papamakarios2021}.

\subsection{Multivariate Conditional Transformation Models}
\label{sec:mctm}

\gls{MCTM} \citep{Klein2022} use element-wise transformations $\tilde{h}_{j}\left(y_{j}\right), j=1, \ldots, J$ and a linear triangular $(J \times J)$ matrix $\Lambda$ to model correlations:
\begin{equation}
h_j(y_1,\ldots,y_j) = \lambda_{j 1} \tilde{h}_{1}(y_1)+\ldots+\lambda_{j,j-1} \tilde{h}_{j-1}(y_{j-1}) + \tilde{h}_j(y_j)
\end{equation}
For conditional distributions, $\Lambda$ and transformation parameters $\theta$ can depend on covariates $\mathbf{x}$:
\begin{equation}
h_j(\mathbf{y}|\mathbf{x}) = \sum_{i=1}^{j-1}\lambda_{ji}(\mathbf{x})\tilde{h}_i(y_i;\theta_i(\mathbf{x})) + \tilde{h}_j(y_j;\theta_j(\mathbf{x}))
\end{equation}

\subsection{Autoregressive Transformation Models}
\label{sec:ar_models}
Autoregressive flows factorize multivariate distributions based on the chain rule of probability:
\begin{equation}
p_y(\mathbf{y}) = \prod_{i=1}^D p_y(y_i|\mathbf{y}_{<i})
\end{equation}
Applying the change of variables formula yields:
\begin{equation}
p_y(\mathbf{y}) = p_z\left(h(y)\right) \left|\det\nabla{h}(y)\right| = \prod_{i=1}^D p_z\left(h_i(y_{i},\mathbf{y}_{<i})\right)\left|\det\nabla{h}_i(y_{i},\mathbf{y}_{<i})\right|
\end{equation}
where $h_i$ is a diffeomorphism applied to the $i$-th element of $\mathbf{y}$, conditioned on preceding elements $\mathbf{y}_{<i}$.
The lower triangular Jacobian determinant is the product of its diagonal elements:
\begin{equation}
\det\nabla{h}(y) = \prod_{i=1}^D \frac{\partial F_i}{\partial y_i}
\end{equation}
This is implemented as $z_i=h_i(y_i, \theta_i)$ with $\theta_i=c(\mathbf{y}_{<i})$, where the conditioner $c_i$, often a neural network, captures dependencies and enforces the autoregressive property \citep{Papamakarios2021}. New samples are obtained via the inverse transformation $y_i=T_i^{-1}(z_i, \theta_i)$.

Two common architectures for the conditioner are coupling layers and masked autoregressive networks. Coupling flows \citep{Dinh2017} split the response $\mathbf{y}\in\mathbb{R}^D$ into subsets $(\mathbf{y}_A, \mathbf{y}_B) \in (\mathbb{R}^{d},\mathbb{R}^{D-d})$, with one subset conditioning the transformation of the other:
\begin{equation}
\mathbf{y} = \begin{cases}
\mathbf{y}_A = \mathbf{y}_A \\
\mathbf{y}_B = h(\mathbf{y}_B; \Theta(\mathbf{y}_A))
\end{cases}
\end{equation}
\gls{MADE} \citep{Germain2015}, as used in \glspl{MAF} \citep{Papamakarios2018} and \glspl{IAF} \citep{Kingma2016}, generalizes this using masked neural networks to enforce autoregressive constraints.

In \gls{MADE}, the output $d$ depends only on inputs $\mathbf{y}_{<d}$. This is achieved by element-wise multiplying weight matrices by binary masks, zeroing connections that violate the autoregressive property. Whether the masked network uses $\mathbf{y}$ or the latent representation $\mathbf{z}$ as input affects only whether inference or sampling is iterative \citep{Papamakarios2021}.

\subsection{Structured Additive Predictors for Enhanced Flexibility}
\label{sec:sap}

Structured Additive Predictors \citep[\gls{SAP};][]{Fahrmeir2013,Fahrmeir2009} allow for (non-)linear effects, interactions, and other structured terms. An exemplary predictor is given by
\begin{equation}
\eta(\mathbf{x}) = \beta_0 + \sum_{u=1}^{U} f_u(x_u) + \sum_{u < v} f_{uv}(x_u, x_v) + \dots 
\end{equation}
Here, $f_u(x_u)$ is a linear or non-linear function, where the latter is usually specified using regression splines to stay in the context of parametric regression. $f_{uv}(x_u,x_v)$ are linear or smooth interaction effects. The order of interaction determines the degree of interpretability. In our hybrid approach, \gls{SAP} can be used to parameterize marginal shifts $\beta_j(\mathbf{x})$ in $H_1$.

\clearpage

\section{Extended Results}

\subsection{negative log-likelihood on 2D simluation datasets}
\label{sec:sim-nll}

\begin{table*}[ht!]
\centering
\begin{tabular}{lllll}
  \toprule
  dataset name & \multicolumn{2}{c}{circles} & \multicolumn{2}{c}{moons}                                           \\
  conditional  & False                       & True                      & False              & True               \\
  model        &                             &                           &                    &                    \\
  \midrule
  MVN          & -0.204 $\pm$ 0.000          & -0.423 $\pm$ 0.002        & -0.151 $\pm$ 0.002 & -0.704 $\pm$ 0.004 \\
  MCTM         & -0.490 $\pm$ 0.002          & -0.489 $\pm$ 0.002        & -0.536 $\pm$ 0.000 & -1.046 $\pm$ 0.006 \\
  MAF (S)      & -1.123 $\pm$ 0.022          & -1.123 $\pm$ 0.022        & -1.611 $\pm$ 0.042 & -1.611 $\pm$ 0.042 \\
  MAF (B)      & -1.179 $\pm$ 0.014          & -1.179 $\pm$ 0.014        & -1.625 $\pm$ 0.016 & -1.625 $\pm$ 0.016 \\
  CF (S)       & -1.045 $\pm$ 0.132          & -1.861 $\pm$ 0.056        & -1.587 $\pm$ 0.052 & -2.306 $\pm$ 0.050 \\
  CF (B)       & -0.651 $\pm$ 0.202          & -1.657 $\pm$ 0.038        & -1.350 $\pm$ 0.106 & -2.186 $\pm$ 0.116 \\
  \midrule
  HCF (S)      & -1.175 $\pm$ 0.012          & -1.870 $\pm$ 0.018        & -1.628 $\pm$ 0.018 & -2.332 $\pm$ 0.014 \\
  HCF (B)      & -1.071 $\pm$ 0.024          & -1.826 $\pm$ 0.078        & -1.583 $\pm$ 0.042 & -2.332 $\pm$ 0.032 \\
  \bottomrule
\end{tabular}
\caption{
    Test negative log-likelihood on 2D simluation datasets (lower is better).
    Log-likelihoods are averaged over 20 trials, and their spread is reported as two standard deviations.
}
\label{tab:simulated_results}
\end{table*}

\clearpage

\subsection{Scatter plots of Samples from Malnutrition Models}
\label{sec:malnutrition_samples}

\begin{figure}[h!]
\centering
\begin{subfigure}[t]{0.5\textwidth}
\centering
\includegraphics[width=\textwidth]{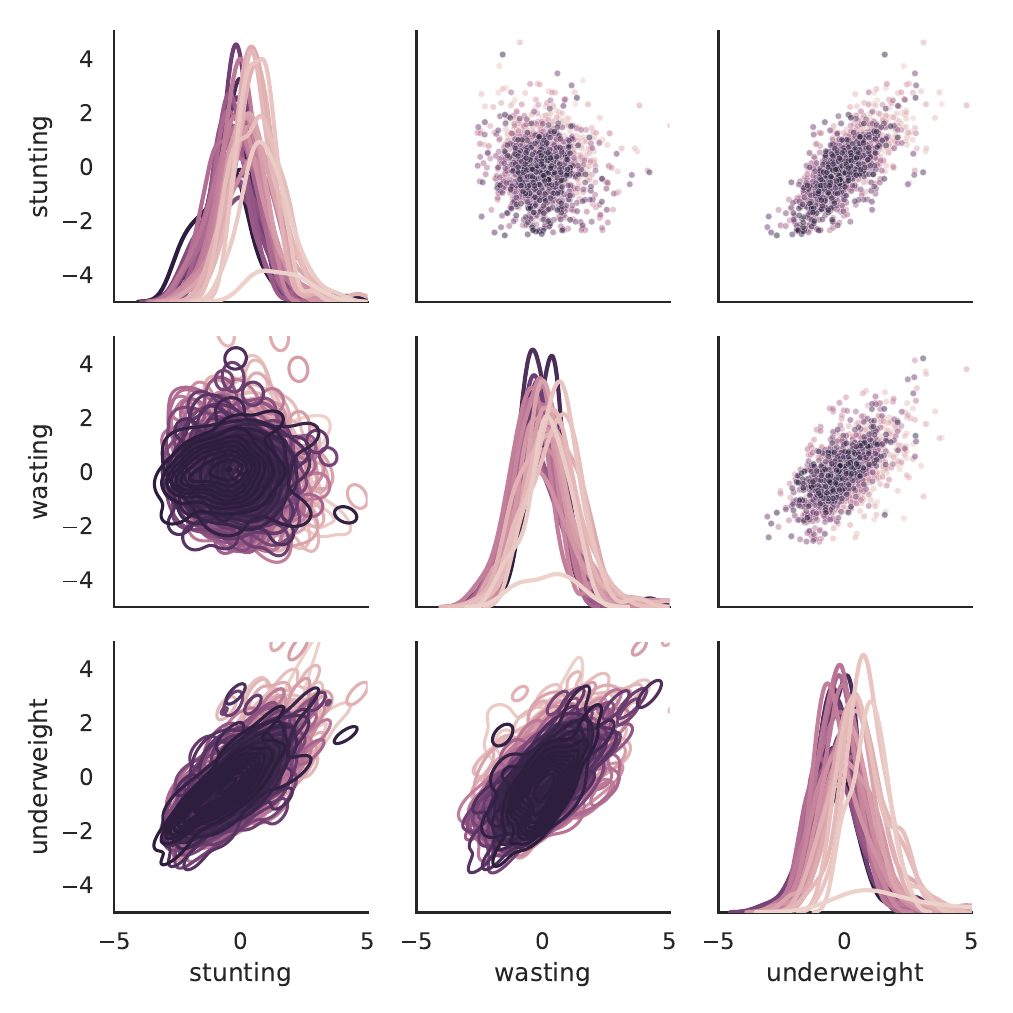}
\caption{Random samples from the validation dataset.}
\end{subfigure}%
~
\begin{subfigure}[t]{0.5\textwidth}
\centering
\includegraphics[width=\textwidth]{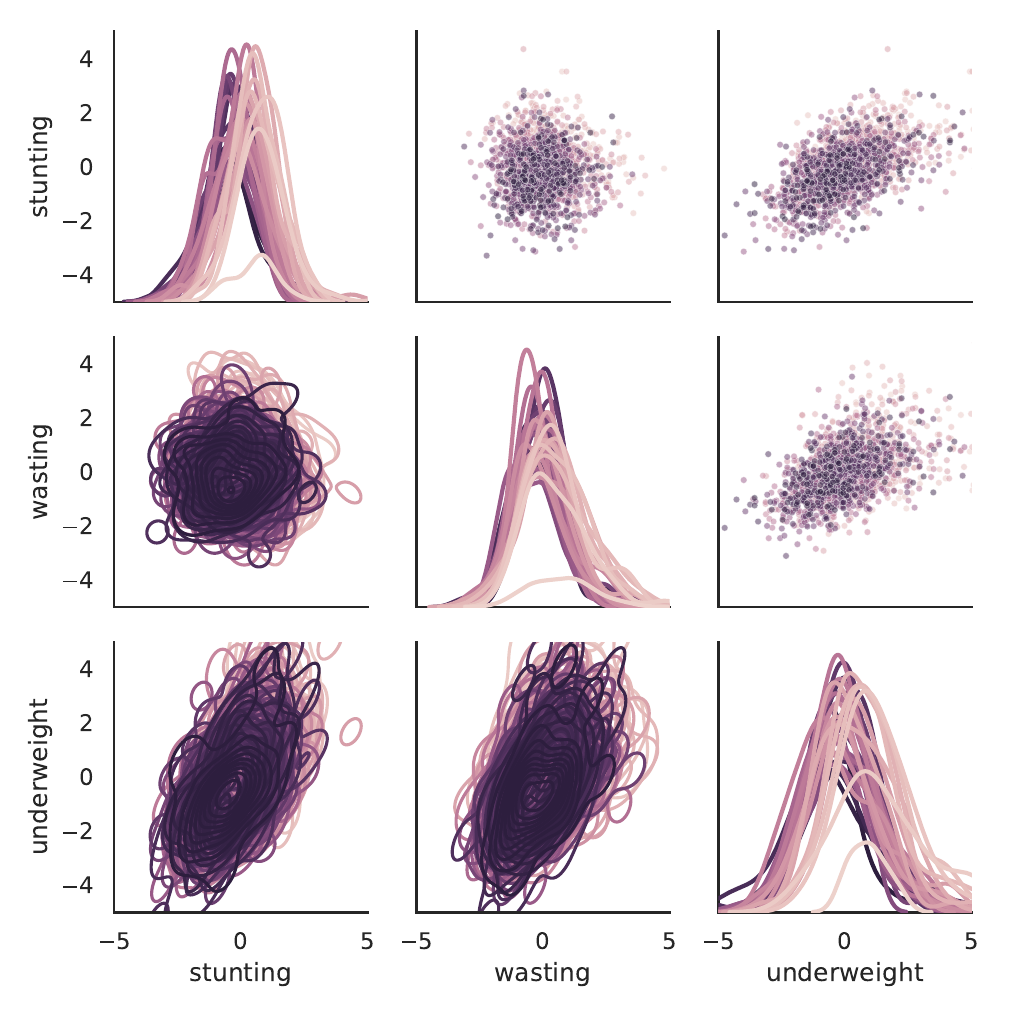}
\caption{Random samples from the \gls{MCTM} model.}
\end{subfigure}
\\
\begin{subfigure}[t]{0.5\textwidth}
\centering
\includegraphics[width=\textwidth]{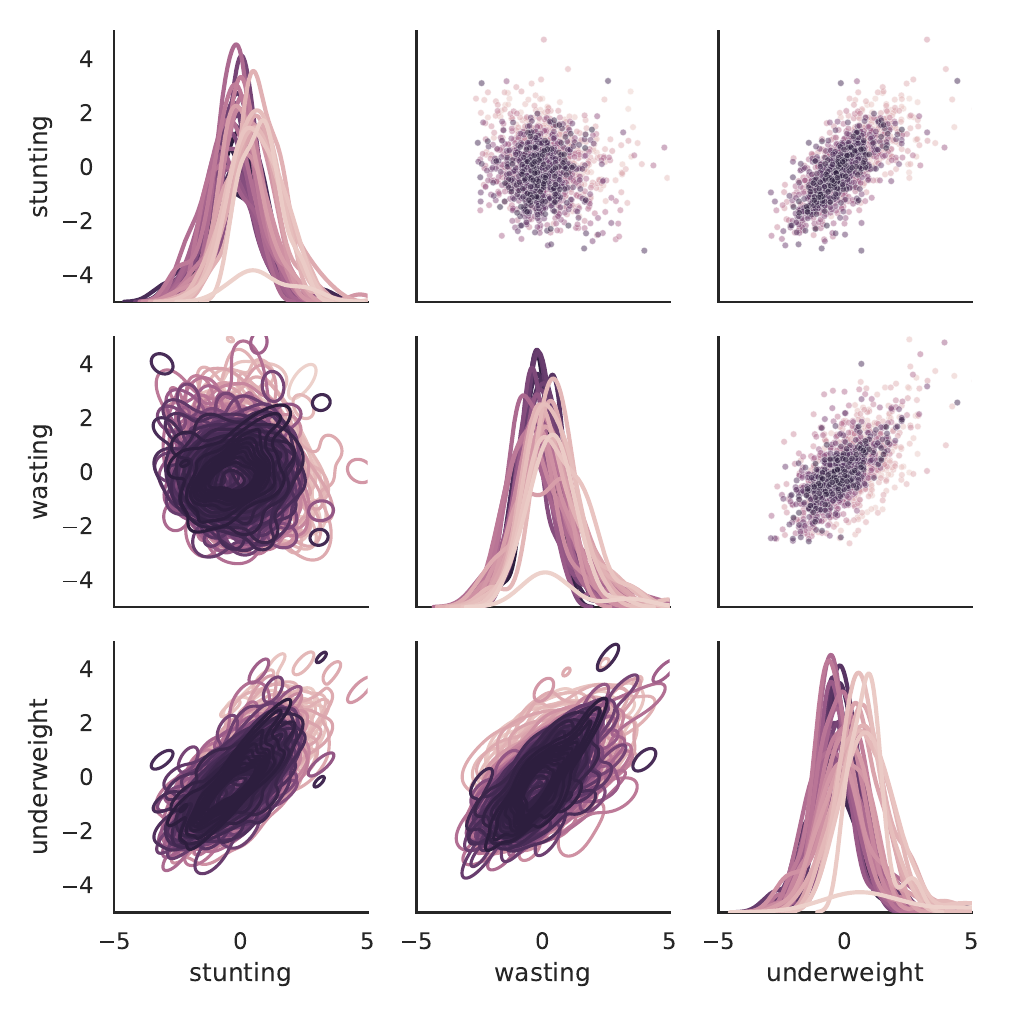}
\caption{Random samples from the hybrid model using quadratic splines as the transformation function.}
\end{subfigure}%
~
\begin{subfigure}[t]{0.5\textwidth}
\centering
\includegraphics[width=\textwidth]{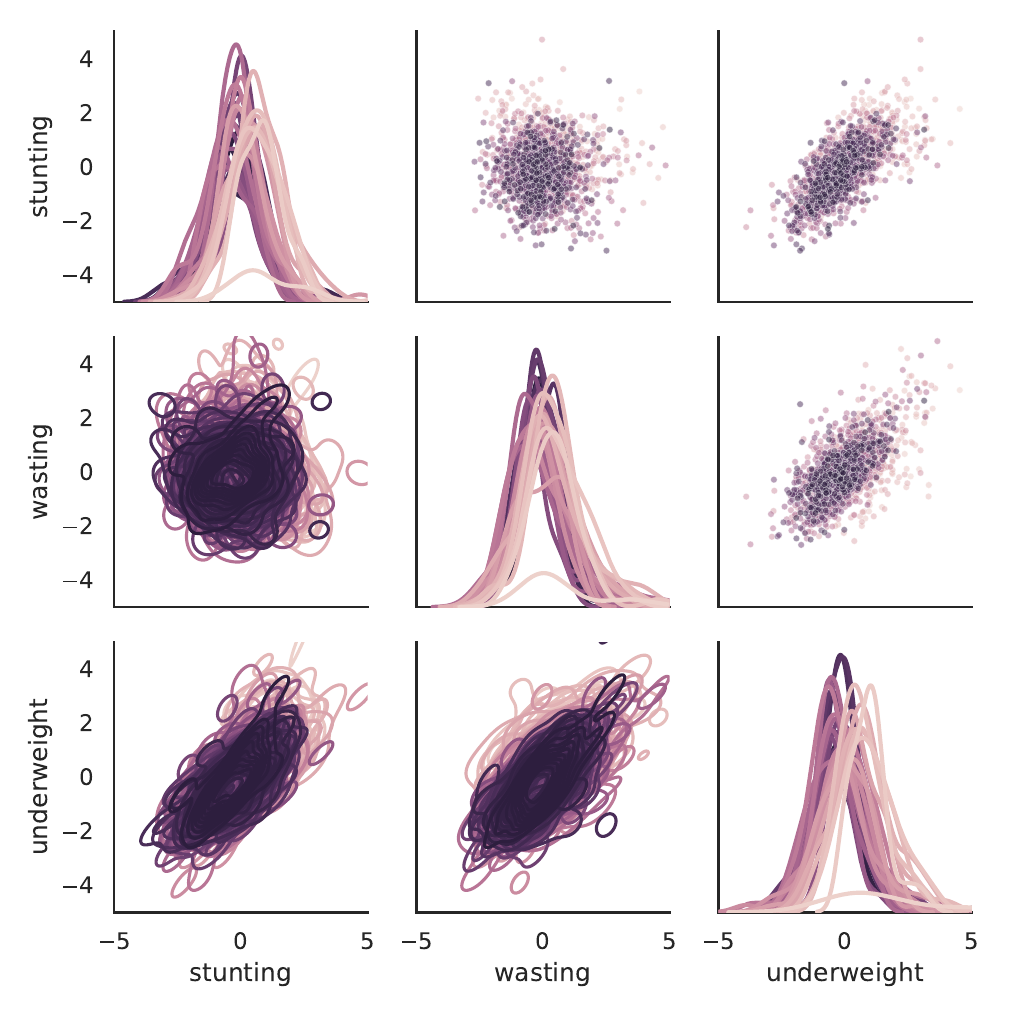}
\caption{Random samples from the hybrid model using Bernstein polynomials as the transformation function.}
\end{subfigure}
\caption{Scatter plots of the three target variables—\texttt{stunting}, \texttt{wasting}, and \texttt{underweight}.
         In the upper triangular section, random data points are illustrated, either from the validation dataset or sampled from the three models.
         The diagonal displays the marginal kernel density estimation (KDE) plots, while the lower triangular region contains the two-dimensional KDE plots.
         The data is categorized by the variable \texttt{cage}.}
\label{fig:malnutrition_samples}
\end{figure}

\autoref{fig:malnutrition_samples} shows that \gls{MCTM} captures marginals well but fails to model dependencies.
The \gls{HMAF} models, especially with spline transformations (\gls{HMAF}~(S)), show a greater resemblance to the observed data in the pairwise density plots.

\subsection{Interpretability of Covariate Effects}
\label{sec:interpretation_details}

Interpretable covariate effects are important in many regression tasks. \glspl{CTM} offer a balance between interpretability and flexibility \citep{Hothorn2014}. The base distribution and transformation function parameterization determine the interpretability.

In a univariate \gls{CTM} with covariate $x$:
\begin{equation}
\mathbb{P}(Y \leq y | X = x) = F_Z(h(y | \theta_x)),
\end{equation}
where $h(y | \theta_x) = \alpha(y)^\top \bm{\vartheta} + \beta x$, with $\beta$ is the covariate effect on the transformed response, and  $\alpha(y)^\top \bm{\vartheta}$ controlling the distribution's shape.

\begin{itemize}
\item \textbf{Logistic Distribution:} $\beta$ is the log-odds ratio, quantifying the change in the odds of $Y \le y$ associated with a unit increase in $x$.
\item \textbf{Minimum Extreme Value Distribution:} $\beta$ is the log-hazard ratio, representing the influence of $x$ on the instantaneous risk of an event.
\item \textbf{Gaussian Distribution:} With a linear transformation $h(y|\theta_x)$, $\beta$ is the change in $Y$'s conditional mean per unit change in $x$ (scaled by the standard deviation).
With non-linear $h$, $\beta$'s interpretation is less direct, affecting multiple moments.
\end{itemize}

\subsection{Interpretable Covariate Effect on the Dependency Structure in MCTMs}

As described in \cite{Klein2022}, the covariate effect on the dependency in the \gls{MCTM} model can be interpreted as Spearman's rank correlation from the covariance matrix

\begin{equation}
\Sigma = \Lambda^\top \Lambda^{-\top}
\end{equation}

via the Pearson correlation coefficient,

\begin{equation}
\rho_{ij} = \frac{\Sigma_{ij}}{\sigma_{i} \sigma_{j}} = \frac{\Sigma_{ij}}{\sqrt{\Sigma_{ii} \Sigma_{jj}}}
\end{equation}

where $\Sigma_{ij}$ represents the covariance between variables $y_i$ and $y_j$ and $\Sigma_{ii}$ the variance $\sigma_i^2$. To convert Pearson correlations to Spearman's rank correlation, the following transformation is applied:

\begin{equation}
\rho_s = \frac{6}{\pi} \arcsin\left(\frac{\rho}{2}\right)
\end{equation}

\begin{figure}[htb!]
\centering
\includegraphics{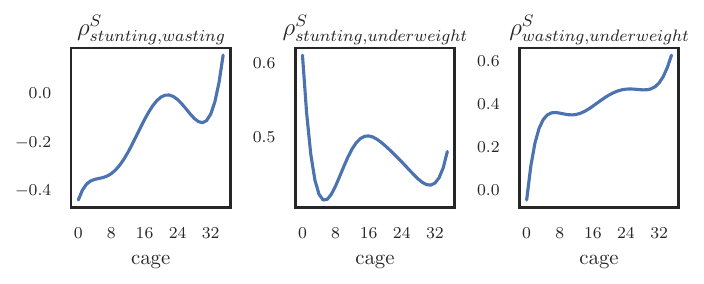}
\caption{Comparison of Spearman's rank correlation $\rho^S$ estimates between \texttt{stunting}, \texttt{wasting}, and \texttt{underweight} with respect to \texttt{cage}.}
\label{fig:rank_correlation_comparison}
\end{figure}

The resulting covariate-dependent rank correlations are shown in \autoref{fig:rank_correlation_comparison}.

\subsection{Benchmark Datasets}
\label{sec:benchmark_data}

We evaluate our method on five common benchmark datasets:
POWER\footnotehyperlink{https://archive.ics.uci.edu/dataset/235},
GAS\footnotehyperlink{https://archive.ics.uci.edu/dataset/322}.
HEPMASS\footnotehyperlink{https://archive.ics.uci.edu/dataset/347},
MINIBOONE\footnotehyperlink{https://archive.ics.uci.edu/dataset/199/},
BSDS300\footnotehyperlink{https://www2.eecs.berkeley.edu/Research/Projects/CS/vision/bsds/}.
We follow the preprocessing steps from \citet{Papamakarios2018}\footnote{\url{https://github.com/francois-rozet/uci-datasets}}.

\begin{figure}[hb!]
\centering
\includegraphics[height=0.6\textheight]{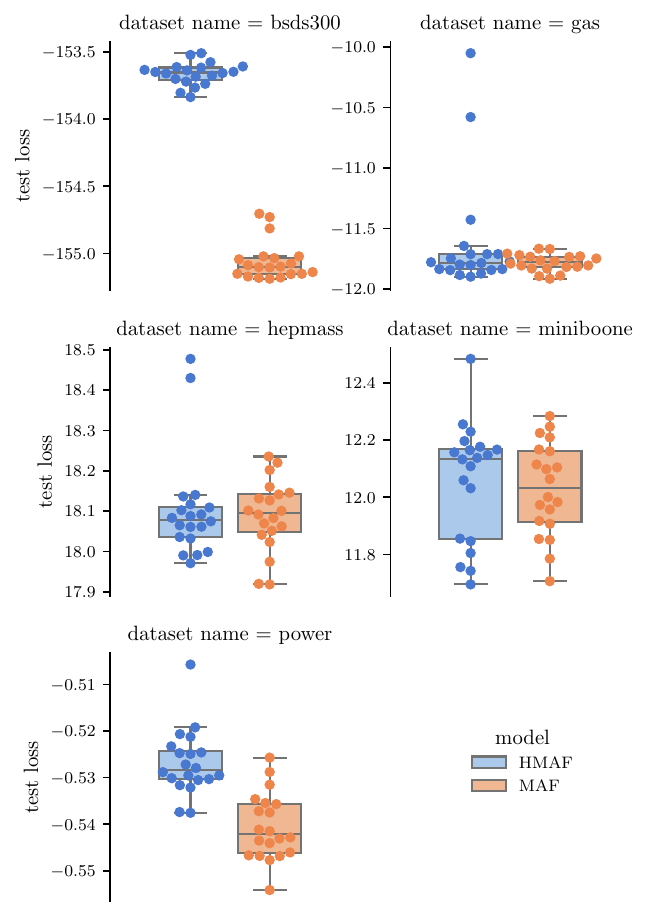}
\caption{Distribution of negative log-likelihood (NLL) scores for the \gls{HMAF} and \gls{MAF} models, resulting from 20 runs with different random weight initializations.
         The box plots show the median, interquartile range (IQR), and the data range of 1.5 times the IQR, outside the IQR.
         The swarm plots above the boxes show the individual NLL scores for each run.}
\label{tab:benchmark-nll-box-plot}
\end{figure}

\autoref{tab:benchmark-nll-box-plot} shows the distribution of \gls{NLL} scores for \gls{HMAF} and \gls{MAF} across 20 different random initializations.
The figure visualizes performance robustness.
Consistent lower \gls{NLL} values indicate better performance, smaller IQRs suggest less sensitivity to initialization, and a lower spread of outliers suggests more stable training.

\section{Complexity Estimates (Runtime and Number of Parameters)}

\subsection{Runtime of Model Variants}

The tables below present the runtime for training and evaluating our models on the HPC Cluster at the University of Applied Sciences Esslingen, utilizing NVIDIA L40S GPUs with 48 GB VRAM. 
Note that our code is not fully optimized for runtime, and variations in training and inference times are significantly influenced by hyperparameters such as the number of epochs and the use of early stopping.

\begin{table}[h!]
  \centering
  \caption{Runtime in Minutes for training and evaluation of models on benchmark data.
    Variance resulting deviations from 20 runs reported as standard deviation.}
  \begin{tabular}{llll}
    \toprule
    model                    & dataset name & train                  & evaluation          \\
    \midrule
    \multirow[c]{5}{*}{HMAF} & bsds300      & 1191.992 $\pm$ 537.944 & 481.991 $\pm$ 0.650 \\
                             & gas          & 319.809 $\pm$ 132.472  & 14.931 $\pm$ 0.043  \\
                             & hepmass      & 229.736 $\pm$ 155.484  & 15.094 $\pm$ 0.047  \\
                             & miniboone    & 82.882 $\pm$ 58.692    & 3.933 $\pm$ 0.012   \\
                             & power        & 437.108 $\pm$ 63.707   & 11.321 $\pm$ 0.091  \\
    \midrule
    \multirow[c]{5}{*}{MAF}  & bsds300      & 261.977 $\pm$ 70.957   & 16.716 $\pm$ 0.022  \\
                             & gas          & 68.993 $\pm$ 0.073     & 1.858 $\pm$ 0.005   \\
                             & hepmass      & 34.774 $\pm$ 0.003     & 1.540 $\pm$ 0.004   \\
                             & miniboone    & 16.486 $\pm$ 1.404     & 0.279 $\pm$ 0.001   \\
                             & power        & 136.796 $\pm$ 0.004    & 4.979 $\pm$ 0.120   \\
    \bottomrule
  \end{tabular}
\end{table}

\begin{table}[h!]
  \centering
  \caption{Mean runtime in seconds for training and evaluation of models on simulated data.
    Variance resulting deviations from 20 runs reported as standard deviation.}
  \begin{tabular}{lllll}
    \toprule
                                 &                           &         & train                 & evaluation         \\
    dataset name                 & conditional               & model   &                       &                    \\
    \midrule
    \multirow[c]{16}{*}{circles} & \multirow[c]{8}{*}{False} & CF (B)  & 35.043 $\pm$ 15.008   & 16.782 $\pm$ 0.238 \\
                                 &                           & CF (S)  & 317.228 $\pm$ 233.129 & 17.428 $\pm$ 0.181 \\
                                 &                           & HCF (B) & 110.039 $\pm$ 11.230  & 57.759 $\pm$ 0.830 \\
                                 &                           & HCF (S) & 76.960 $\pm$ 37.909   & 29.489 $\pm$ 0.191 \\
                                 &                           & MAF (B) & 91.213 $\pm$ 0.354    & 19.562 $\pm$ 2.876 \\
                                 &                           & MAF (S) & 145.815 $\pm$ 36.820  & 17.369 $\pm$ 1.585 \\
                                 &                           & MCTM    & 42.597 $\pm$ 0.724    & 29.009 $\pm$ 0.375 \\
                                 &                           & MVN     & 44.892 $\pm$ 0.099    & 10.722 $\pm$ 0.067 \\
    \cline{2-5}
                                 & \multirow[c]{8}{*}{True}  & CF (B)  & 41.655 $\pm$ 20.906   & 20.423 $\pm$ 0.223 \\
                                 &                           & CF (S)  & 329.650 $\pm$ 211.377 & 17.864 $\pm$ 0.267 \\
                                 &                           & HCF (B) & 66.384 $\pm$ 40.931   & 73.782 $\pm$ 1.620 \\
                                 &                           & HCF (S) & 171.494 $\pm$ 107.205 & 31.913 $\pm$ 0.128 \\
                                 &                           & MAF (B) & 91.153 $\pm$ 0.337    & 20.306 $\pm$ 2.840 \\
                                 &                           & MAF (S) & 145.765 $\pm$ 36.843  & 17.912 $\pm$ 1.554 \\
                                 &                           & MCTM    & 304.875 $\pm$ 112.109 & 31.375 $\pm$ 2.577 \\
                                 &                           & MVN     & 56.360 $\pm$ 36.401   & 11.708 $\pm$ 0.055 \\
    \midrule
    \multirow[c]{16}{*}{moons}   & \multirow[c]{8}{*}{False} & CF (B)  & 33.991 $\pm$ 12.603   & 16.842 $\pm$ 0.145 \\
                                 &                           & CF (S)  & 443.163 $\pm$ 80.343  & 17.129 $\pm$ 0.312 \\
                                 &                           & HCF (B) & 184.718 $\pm$ 71.683  & 67.237 $\pm$ 2.418 \\
                                 &                           & HCF (S) & 76.214 $\pm$ 39.760   & 29.368 $\pm$ 0.106 \\
                                 &                           & MAF (B) & 91.252 $\pm$ 0.538    & 19.872 $\pm$ 3.607 \\
                                 &                           & MAF (S) & 134.682 $\pm$ 62.895  & 14.719 $\pm$ 3.458 \\
                                 &                           & MCTM    & 42.544 $\pm$ 0.742    & 28.949 $\pm$ 0.141 \\
                                 &                           & MVN     & 44.823 $\pm$ 0.024    & 11.497 $\pm$ 0.046 \\
    \cline{2-5}
                                 & \multirow[c]{8}{*}{True}  & CF (B)  & 44.995 $\pm$ 20.047   & 20.477 $\pm$ 0.313 \\
                                 &                           & CF (S)  & 368.668 $\pm$ 250.469 & 17.644 $\pm$ 0.415 \\
                                 &                           & HCF (B) & 48.243 $\pm$ 17.635   & 40.735 $\pm$ 0.516 \\
                                 &                           & HCF (S) & 169.016 $\pm$ 62.618  & 31.744 $\pm$ 0.146 \\
                                 &                           & MAF (B) & 91.215 $\pm$ 0.529    & 20.623 $\pm$ 3.563 \\
                                 &                           & MAF (S) & 135.603 $\pm$ 68.688  & 15.308 $\pm$ 3.487 \\
                                 &                           & MCTM    & 475.688 $\pm$ 1.217   & 35.501 $\pm$ 7.524 \\
                                 &                           & MVN     & 48.118 $\pm$ 18.322   & 11.030 $\pm$ 0.071 \\
    \bottomrule
  \end{tabular}
\end{table}

\begin{table}[h!]
  \centering
\caption{Mean runtime in seconds for training and evaluation of models on malnutrition data.
Variance resulting deviations from 20 runs reported as standard deviation.}
\begin{tabular}{lll}
\toprule
 & train & evaluation \\
model &  &  \\
\midrule
HMAF (B) & 260.752 $\pm$ 121.895 & 20.823 $\pm$ 0.535 \\
HMAF (S) & 1993.317 $\pm$ 717.933 & 19.649 $\pm$ 0.110 \\
MCTM & 4106.187 $\pm$ 725.136 & 16.877 $\pm$ 0.847 \\
\bottomrule
\end{tabular}
\end{table}


\end{document}


\onecolumn

\title{Hybrid Bernstein Normalizing Flows for Flexible Multivariate Density Regression with Interpretable Marginals\\(Supplementary Material)}
\maketitle

\section{Model Details}
\label{sec:model_details}

In the following section, we provide an overview of the most important hyper parameters used to produce our results.
All models are implemented using \texttt{TensorFlow} (v2.15.1) and \texttt{TensorFlow Probability} (v0.23.0) and are trained using Adam~\citep{Tensorflow,TensorflowProbability,Kingma2017a}.
The implementation is available on GitHub\footnote{\url{https://github.com/MArpogaus/hybrid_flows}}.

\subsection{Simulation Data}
The following tables contain all hyper parameters of the models trained on the bivariate simulation data.


\begin{table}[h!]
  \centering
  \begin{tabular}{ll}
\toprule
 & CF (B) \\
\midrule
JIT Compile & True \\
Epochs & 400 \\
Batch Size & 512 \\
Learning Rate Decay Steps & epochs \\
Initial Learning Rate & 0.001 \\
Learning Rate Scheduler & cosine\_decay \\
Early Stopping & 50 \\
Validation Split & 0.250 \\
Output Distribution & coupling\_flow \\
Number of Flows & 2 \\
Bijector & BernsteinPolynomial \\
Bernstein Bijector Extrapolation & False \\
Bernstein Bijector Domain Min & 0 \\
Bernstein Bijector Domain Max & 1 \\
Invert Bijector & True \\
Allow Flexible Bounds for Parameters & False \\
Bounds constraints for Parameters & linear \\
Nested Bernstein Bijector Codomain Min & 0 \\
Nested Bernstein Bijector Codomain Max & 1 \\
Number of Parameters & 300 \\
Low Constraint for Layer Overwrite -2 & -5 \\
High Constraint for Layer Overwrite -2 & 5 \\
Low Constraint for Layer Overwrite -1 & -5 \\
High Constraint for Layer Overwrite -1 & 5 \\
Activation Function & relu \\
Batch Normalization & False \\
Dropout Rate & 0 \\
First Hidden Units & 128 \\
Second Hidden Units & 128 \\
Third Hidden Units & 128 \\
\bottomrule
\end{tabular}

  \caption{Common Hyperparameters of \gls{CF} (B) models trained on simulated 2D data.}
\end{table}
\begin{table}[h!]
  \centering
  \begin{tabular}{lllll}
\toprule
Dataset Name & \multicolumn{2}{c}{moons} & \multicolumn{2}{c}{circles} \\
Conditional & True & False & True & False \\
\midrule
Conditional & True & -- & True & -- \\
Conditional Event Shape & 1 & -- & 1 & -- \\
\bottomrule
\end{tabular}

  \caption{Specific Hyperparameters of \gls{CF} (B) models trained on simulated 2D data.}
\end{table}

\begin{table}[h!]
  \centering
  \begin{tabular}{ll}
\toprule
 & CF (S) \\
\midrule
JIT Compile & True \\
Epochs & 400 \\
Batch Size & 512 \\
Learning Rate Decay Steps & epochs \\
Initial Learning Rate & 0.001 \\
Learning Rate Scheduler & cosine\_decay \\
Early Stopping & 50 \\
Validation Split & 0.250 \\
Output Distribution & coupling\_flow \\
Number of Flows & 2 \\
Bijector & RationalQuadraticSpline \\
Number of Parameters & 95 \\
Activation Function & relu \\
Batch Normalization & False \\
Dropout Rate & 0 \\
First Hidden Units & 128 \\
Second Hidden Units & 128 \\
Third Hidden Units & 128 \\
Spline Bijector Domain Min & -5 \\
Interval Width for Nested Spline Bijector & 10 \\
Minimum Slope for Nested Spline Bijector & 0.001 \\
Minimum Bin Width for Nested Spline Bijector & 0.001 \\
Number of Bins for Nested Spline Bijector & 32 \\
\bottomrule
\end{tabular}

  \caption{Common Hyperparameters of \gls{CF} (S) models trained on simulated 2D data.}
\end{table}
\begin{table}[h!]
  \centering
  \begin{tabular}{lllll}
\toprule
Dataset Name & \multicolumn{2}{c}{moons} & \multicolumn{2}{c}{circles} \\
Conditional & True & False & True & False \\
\midrule
Conditional & True & -- & True & -- \\
Conditional Event Shape & 1 & -- & 1 & -- \\
\bottomrule
\end{tabular}

  \caption{Specific Hyperparameters of \gls{CF} (S) models trained on simulated 2D data.}
\end{table}

\begin{table}[h!]
  \centering
  \begin{tabular}{ll}
\toprule
 & HCF (B) \\
\midrule
Two-Stage Training & True \\
First Marginal Bijector & BernsteinPolynomial \\
Invert First Marginal Bijector & True \\
Marginal Bijector Domain Min & 0 \\
Marginal Bijector Domain Max & 1 \\
Marginal Bijector Extrapolation & False \\
First Marginal Bijector Parameter Constraint Function & mctm.activations.get\_thetas\_constrain\_fn \\
Allow Flexible Bounds for First Marginal Bijector & False \\
Bounds constraints for First Marginal Bijector & linear \\
First Marginal Bijector Codomain Max & 5 \\
First Marginal Bijector Codomain Min & -5 \\
First Marginal Bijector Parameter Shape 0 & 2 \\
First Marginal Bijector Parameter Shape 1 & 300 \\
First Joint Bijector & RealNVP \\
Parameter Shape 0 for Joint Bijector & 1 \\
Marginal Epochs & 1000 \\
Marginal Batch Size & 1024 \\
Marginal Learning Rate & 0.050 \\
Marginal Reduce LR on Plateau & False \\
Marginal Early Stopping & 50 \\
Marginal Monitor & val\_loss \\
Marginal Verbosity & True \\
Marginal Validation Split & 0.250 \\
Joint Epochs & 400 \\
Joint Batch Size & 512 \\
Joint Learning Rate Decay Steps & epochs \\
Joint Initial Learning Rate & 0.001 \\
Joint Learning Rate Scheduler & cosine\_decay \\
Joint Early Stopping & 50 \\
Joint Validation Split & 0.250 \\
Masked Dimensions & 1 \\
First Nested Bijector & BernsteinPolynomial \\
Invert Nested Bijector & True \\
Nested Bernstein Bijector Domain Min & -5 \\
Nested Bernstein Bijector Domain Max & 5 \\
Nested Bernstein Bijector Extrapolation & False \\
Nested Bijector Parameter Constraint Function & mctm.activations.get\_thetas\_constrain\_fn \\
Bounds constraints of Nested Bernstein Bijector & linear \\
Nested Bernstein Bijector Codomain Max & 5 \\
Nested Bernstein Bijector Codomain Min & -5 \\
Activation Function for Joint Bijector & relu \\
Batch Normalization for Joint Bijector & False \\
Dropout Rate for Joint Bijector & False \\
First Hidden Units for Joint Bijector & 512 \\
Input Shape 0 for Joint Bijector & 1 \\
\bottomrule
\end{tabular}

  \caption{Common Hyperparameters of \gls{HCF} (B) models trained on simulated 2D data.}
\end{table}
\begin{table}[h!]
  \centering
  \resizebox{\linewidth}{!}{\begin{tabular}{lllll}
\toprule
Dataset Name & \multicolumn{2}{c}{moons} & \multicolumn{2}{c}{circles} \\
Conditional & True & False & True & False \\
\midrule
JIT Compile & True & False & True & False \\
Parameters Function for First Marginal Bijector & bernstein\_polynomial & parameter\_vector & bernstein\_polynomial & parameter\_vector \\
Joint Bijector Conditional Event Shape & 1 & -- & 1 & -- \\
Extrapolation for First Marginal Bijector & True & -- & True & -- \\
First Marginal Bijector Conditional Event Shape & 1 & -- & 1 & -- \\
Polynomial Order for First Marginal Bijector & 1 & -- & 1 & -- \\
Second Hidden Units for Joint Bijector & 32 & 128 & 512 & 32 \\
Third Hidden Units for Joint Bijector & 512 & 1024 & 32 & 1024 \\
Parameter Shape 1 for Joint Bijector & 300 & 2000 & 1500 & 1500 \\
Is Conditional for Joint Bijector & True & -- & True & -- \\
\bottomrule
\end{tabular}
}
  \caption{Specific Hyperparameters of \gls{HCF} (B) models trained on simulated 2D data.}
\end{table}

\begin{table}[h!]
  \centering
  \begin{tabular}{ll}
\toprule
 & HCF (S) \\
\midrule
Two-Stage Training & True \\
First Marginal Bijector & BernsteinPolynomial \\
Invert First Marginal Bijector & True \\
Marginal Bijector Domain Min & 0 \\
Marginal Bijector Domain Max & 1 \\
Marginal Bijector Extrapolation & False \\
First Marginal Bijector Parameter Constraint Function & mctm.activations.get\_thetas\_constrain\_fn \\
Allow Flexible Bounds for First Marginal Bijector & False \\
Bounds constraints for First Marginal Bijector & linear \\
First Marginal Bijector Codomain Max & 5 \\
First Marginal Bijector Codomain Min & -5 \\
First Marginal Bijector Parameter Shape 0 & 2 \\
First Marginal Bijector Parameter Shape 1 & 300 \\
First Joint Bijector & RealNVP \\
Parameter Shape 0 for Joint Bijector & 1 \\
Marginal Epochs & 1000 \\
Marginal Batch Size & 1024 \\
Marginal Learning Rate & 0.050 \\
Marginal Reduce LR on Plateau & False \\
Marginal Early Stopping & 50 \\
Marginal Monitor & val\_loss \\
Marginal Verbosity & True \\
Marginal Validation Split & 0.250 \\
Joint Epochs & 400 \\
Joint Batch Size & 512 \\
Joint Learning Rate Decay Steps & epochs \\
Joint Initial Learning Rate & 0.001 \\
Joint Learning Rate Scheduler & cosine\_decay \\
Joint Early Stopping & 50 \\
Joint Validation Split & 0.250 \\
Masked Dimensions & 1 \\
First Nested Bijector & RationalQuadraticSpline \\
Activation Function for Joint Bijector & relu \\
Batch Normalization for Joint Bijector & False \\
Dropout Rate for Joint Bijector & False \\
First Hidden Units for Joint Bijector & 128 \\
Second Hidden Units for Joint Bijector & 128 \\
Third Hidden Units for Joint Bijector & 128 \\
Input Shape 0 for Joint Bijector & 1 \\
Parameter Shape 1 for Joint Bijector & 95 \\
Nested Spline Bijector Range Min & -5 \\
Interval Width for Nested Spline Bijector & 10 \\
Minimum Slope for Nested Spline Bijector & 0.001 \\
Minimum Bin Width for Nested Spline Bijector & 0.001 \\
Number of Bins for Nested Spline Bijector & 32 \\
\bottomrule
\end{tabular}

  \caption{Common Hyperparameters of \gls{HCF} (S) models trained on simulated 2D data.}
\end{table}
\begin{table}[h!]
  \centering
  \resizebox{\linewidth}{!}{\begin{tabular}{lllll}
\toprule
Dataset Name & \multicolumn{2}{c}{moons} & \multicolumn{2}{c}{circles} \\
Conditional & True & False & True & False \\
\midrule
JIT Compile & True & False & True & False \\
Parameters Function for First Marginal Bijector & bernstein\_polynomial & parameter\_vector & bernstein\_polynomial & parameter\_vector \\
Joint Bijector Conditional Event Shape & 1 & -- & 1 & -- \\
Extrapolation for First Marginal Bijector & True & -- & True & -- \\
First Marginal Bijector Conditional Event Shape & 1 & -- & 1 & -- \\
Polynomial Order for First Marginal Bijector & 1 & -- & 1 & -- \\
Is Conditional for Joint Bijector & True & -- & True & -- \\
\bottomrule
\end{tabular}
}
  \caption{Specific Hyperparameters of \gls{HCF} (S) models trained on simulated 2D data.}
\end{table}

\begin{table}[h!]
  \centering
  \begin{tabular}{ll}
\toprule
 & MAF (B) \\
\midrule
JIT Compile & False \\
Epochs & 400 \\
Batch Size & 512 \\
Learning Rate Decay Steps & epochs \\
Initial Learning Rate & 0.001 \\
Learning Rate Scheduler & cosine\_decay \\
Early Stopping & 50 \\
Validation Split & 0.250 \\
Output Distribution & masked\_autoregressive\_flow \\
Number of Flows & 2 \\
Bijector & BernsteinPolynomial \\
Bernstein Bijector Extrapolation & False \\
Bernstein Bijector Domain Min & -5 \\
Bernstein Bijector Domain Max & 5 \\
Invert Bijector & True \\
Allow Flexible Bounds for Parameters & False \\
Bounds constraints for Parameters & linear \\
Nested Bernstein Bijector Codomain Min & -5 \\
Nested Bernstein Bijector Codomain Max & 5 \\
Number of Parameters & 128 \\
Activation Function & relu \\
First Hidden Units & 128 \\
Second Hidden Units & 128 \\
Third Hidden Units & 128 \\
Layer Overwrite 0 Domain Min & 0 \\
Layer Overwrite 0 Domain Max & 1 \\
\bottomrule
\end{tabular}

  \caption{Common Hyperparameters of \gls{MAF} (B) models trained on simulated 2D data.}
\end{table}
\begin{table}[h!]
  \centering
  \begin{tabular}{lllll}
\toprule
Dataset Name & \multicolumn{2}{c}{moons} & \multicolumn{2}{c}{circles} \\
Conditional & True & False & True & False \\
\midrule
Conditional & True & -- & True & -- \\
Conditional Event Shape & 1 & -- & 1 & -- \\
\bottomrule
\end{tabular}

  \caption{Specific Hyperparameters of \gls{MAF} (B) models trained on simulated 2D data.}
\end{table}

\begin{table}[h!]
  \centering
  \begin{tabular}{ll}
\toprule
 & MAF (S) \\
\midrule
JIT Compile & False \\
Epochs & 400 \\
Batch Size & 512 \\
Learning Rate Decay Steps & epochs \\
Initial Learning Rate & 0.001 \\
Learning Rate Scheduler & cosine\_decay \\
Early Stopping & 50 \\
Validation Split & 0.250 \\
Output Distribution & masked\_autoregressive\_flow \\
Number of Flows & 2 \\
Bijector & RationalQuadraticSpline \\
Number of Parameters & 95 \\
Activation Function & relu \\
First Hidden Units & 128 \\
Second Hidden Units & 128 \\
Third Hidden Units & 128 \\
Spline Bijector Domain Min & -5 \\
Interval Width for Nested Spline Bijector & 10 \\
Minimum Slope for Nested Spline Bijector & 0.001 \\
Minimum Bin Width for Nested Spline Bijector & 0.001 \\
Number of Bins for Nested Spline Bijector & 32 \\
\bottomrule
\end{tabular}

  \caption{Common Hyperparameters of \gls{MAF} (S) models trained on simulated 2D data.}
\end{table}
\begin{table}[h!]
  \centering
  \begin{tabular}{lllll}
\toprule
Dataset Name & \multicolumn{2}{c}{moons} & \multicolumn{2}{c}{circles} \\
Conditional & True & False & True & False \\
\midrule
Conditional & True & -- & True & -- \\
Conditional Event Shape & 1 & -- & 1 & -- \\
\bottomrule
\end{tabular}

  \caption{Specific Hyperparameters of \gls{MAF} (S) models trained on simulated 2D data.}
\end{table}

\begin{table}[h!]
  \centering
  \begin{tabular}{ll}
\toprule
 & MCTM \\
\midrule
Epochs & 100 \\
Batch Size & 512 \\
Learning Rate Decay Steps & 100 \\
Initial Learning Rate & 0.010 \\
Learning Rate Scheduler & cosine\_decay \\
Validation Split & 0.250 \\
Two-Stage Training & True \\
First Marginal Bijector & BernsteinPolynomial \\
Invert First Marginal Bijector & True \\
Marginal Bijector Domain Min & 0 \\
Marginal Bijector Domain Max & 1 \\
Marginal Bijector Extrapolation & True \\
First Marginal Bijector Parameter Constraint Function & mctm.activations.get\_thetas\_constrain\_fn \\
Allow Flexible Bounds for First Marginal Bijector & False \\
Bounds constraints for First Marginal Bijector & smooth \\
First Marginal Bijector Codomain Max & 4 \\
First Marginal Bijector Codomain Min & -4 \\
Parameters Function for First Marginal Bijector & parameter\_vector \\
First Marginal Bijector Parameter Shape 0 & 2 \\
First Marginal Bijector Parameter Shape 1 & 300 \\
First Joint Bijector & ScaleMatvecLinearOperator \\
Invert First Joint Bijector & True \\
Parameter Shape 0 for Joint Bijector & 1 \\
\bottomrule
\end{tabular}

  \caption{Common Hyperparameters of \gls{MCTM} models trained on simulated 2D data.}
\end{table}
\begin{table}[h!]
  \centering
  \resizebox{\linewidth}{!}{\begin{tabular}{lllll}
\toprule
Dataset Name & \multicolumn{2}{c}{moons} & \multicolumn{2}{c}{circles} \\
Conditional & True & False & True & False \\
\midrule
Early Stopping & False & False & 30 & False \\
Second Marginal Bijector & Shift & -- & Shift & -- \\
Invert Second Marginal Bijector & True & -- & True & -- \\
Parameters Function for Second Marginal Bijector & bernstein\_polynomial & -- & bernstein\_polynomial & -- \\
Second Marginal Bijector Conditional Event Shape & 1 & -- & 1 & -- \\
Extrapolation for Second Marginal Bijector & True & -- & True & -- \\
Second Marginal Bijector Parameter Shape 0 & 2 & -- & 2 & -- \\
Polynomial Order for Second Marginal Bijector & 6 & -- & 6 & -- \\
Joint Bijector Conditional Event Shape & 1 & -- & 1 & -- \\
Extrapolation for Joint Bijector & True & -- & True & -- \\
Polynomial Order for Joint Bijector & 3 & -- & 3 & -- \\
Parameter Shape 1 for Joint Bijector & -- & 1 & -- & 1 \\
\bottomrule
\end{tabular}
}
  \caption{Specific Hyperparameters of \gls{MCTM} models trained on simulated 2D data.}
\end{table}

\begin{table}[h!]
  \centering
  \begin{tabular}{ll}
\toprule
 & MVN \\
\midrule
JIT Compile & False \\
Epochs & 200 \\
Batch Size & 256 \\
Learning Rate Decay Steps & 200 \\
Initial Learning Rate & 0.010 \\
Learning Rate Scheduler & cosine\_decay \\
Early Stopping & False \\
Validation Split & 0.100 \\
Output Distribution & multivariate\_normal \\
\bottomrule
\end{tabular}

  \caption{Common Hyperparameters of \gls{MVN} models trained on simulated 2D data.}
\end{table}
\begin{table}[h!]
  \centering
  \begin{tabular}{lllll}
\toprule
Dataset Name & \multicolumn{2}{c}{moons} & \multicolumn{2}{c}{circles} \\
Conditional & True & False & True & False \\
\midrule
Activation Function & relu & -- & relu & -- \\
Batch Normalization & True & -- & True & -- \\
Dropout Rate & 0 & -- & 0 & -- \\
First Hidden Units & 16 & -- & 16 & -- \\
Second Hidden Units & 16 & -- & 16 & -- \\
Conditional & True & False & True & False \\
Conditional Event Shape & 1 & -- & 1 & -- \\
\bottomrule
\end{tabular}

  \caption{Specific Hyperparameters of \gls{MVN} models trained on simulated 2D data.}
\end{table}

\clearpage
\subsection{Benchmark Data}

The following tables contain all hyper parameters of the models traind on the multivariate benchmark data.

\begin{table}[h!]
  \centering
  \begin{tabular}{ll}
\toprule
 & HMAF (S) \\
\midrule
Two-Stage Training & True \\
First Marginal Bijector & BernsteinPolynomial \\
Invert First Marginal Bijector & True \\
Marginal Bijector Extrapolation & True \\
First Marginal Bijector Parameter Constraint Function & mctm.activations.get\_thetas\_constrain\_fn \\
Allow Flexible Bounds for First Marginal Bijector & False \\
Bounds constraints for First Marginal Bijector & smooth \\
First Marginal Bijector Codomain Max & 4 \\
First Marginal Bijector Codomain Min & -4 \\
Parameters Function for First Marginal Bijector & parameter\_vector \\
First Marginal Bijector Parameter Shape 1 & 2048 \\
Marginal Reduce LR on Plateau & False \\
Marginal Early Stopping & 30 \\
Marginal Monitor & val\_loss \\
Marginal Verbosity & True \\
Joint Epochs & 200 \\
Joint Learning Rate Decay Steps & 200 \\
Joint Learning Rate Scheduler & cosine\_decay \\
Joint Early Stopping & 30 \\
Type of Joint Flow & masked\_autoregressive\_flow\_first\_dim\_masked \\
Number of Flows & 10 \\
Use Invertible Linear Transformations & True \\
Random Permutation Seed & 1 \\
Joint Bijector & RationalQuadraticSpline \\
Joint Bijector Range Min & -4 \\
Interval Width for Joint Bijector & 8 \\
Minimum Slope for Joint Bijector & 0.001 \\
Minimum Bin Width for Joint Bijector & 0.001 \\
Number of Bins for Joint Bijector & 8 \\
Number of Parameters for Joint Bijectors & 23 \\
Activation Function for MAF & relu \\
Activation Function for X0 Parameters & relu \\
Batch Normalization for X0 Parameters & False \\
Dropout Rate for X0 Parameters & False \\
\bottomrule
\end{tabular}

  \caption{Common Hyperparameters of \gls{HMAF} (S) models trained on benchmark data.}
\end{table}

\begin{table}[h!]
  \centering
  \begin{tabular}{llllll}
\toprule
Dataset Name & bsds300 & gas & hepmass & miniboone & power \\
Conditional & False & False & False & False & False \\
\midrule
First Marginal Bijector Parameter Shape 0 & 63 & 8 & 21 & 43 & 6 \\
Marginal Epochs & 1000 & 1000 & 2000 & 2000 & 1000 \\
Marginal Batch Size & 16384 & 16384 & 16384 & 512 & 16384 \\
Marginal Learning Rate & 0.100 & 0.100 & 0.100 & 0.010 & 0.100 \\
Joint Batch Size & 4096 & 4096 & 4096 & 512 & 4096 \\
Joint Initial Learning Rate & 0.001 & 0.001 & 0.001 & 0.001 & 0.001 \\
First Hidden Units for MAF & 512 & 512 & 64 & 64 & 256 \\
Second Hidden Units for MAF & 512 & 512 & 64 & 64 & 256 \\
First Hidden Units for X0 Network & 512 & 512 & 64 & 64 & 256 \\
Second Hidden Units for X0 Network & 512 & 512 & 64 & 64 & 256 \\
\bottomrule
\end{tabular}

  \caption{Specific Hyperparameters of \gls{HMAF} (S) models trained on benchmark data.}
\end{table}

\begin{table}[h!]
  \centering
  \begin{tabular}{ll}
\toprule
 & MAF (S) \\
\midrule
Epochs & 200 \\
Learning Rate Decay Steps & 200 \\
Learning Rate Scheduler & cosine\_decay \\
Early Stopping & 30 \\
Output Distribution & masked\_autoregressive\_flow \\
Number of Flows & 10 \\
Bijector & RationalQuadraticSpline \\
Activation Function & relu \\
Spline Bijector Domain Min & -3 \\
Interval Width for Nested Spline Bijector & 6 \\
Minimum Slope for Nested Spline Bijector & 0.001 \\
Minimum Bin Width for Nested Spline Bijector & 0.001 \\
Use Invertible Linear Transformations & True \\
\bottomrule
\end{tabular}

  \caption{Common Hyperparameters of \gls{MAF} (S) models trained on benchmark data.}
\end{table}
\begin{table}[h!]
  \centering
  \begin{tabular}{llllll}
\toprule
Dataset Name & bsds300 & gas & hepmass & miniboone & power \\
Conditional & False & False & False & False & False \\
\midrule
Batch Size & 4096 & 4096 & 4096 & 512 & 4096 \\
Initial Learning Rate & 0.001 & 0.001 & 0.001 & 0.000 & 0.001 \\
Number of Parameters & 23 & 23 & 23 & 11 & 11 \\
First Hidden Units & 512 & 256 & 64 & 64 & 512 \\
Second Hidden Units & 512 & 256 & 64 & 64 & 512 \\
Number of Bins for Nested Spline Bijector & 8 & 8 & 8 & 4 & 4 \\
\bottomrule
\end{tabular}

  \caption{Specific Hyperparameters of \gls{MAF} (S) models trained on benchmark data.}
\end{table}

\clearpage
\section{Malnutrition Data}

The following tables contain all hyper parameters of the models traind on the trivariate malnutrition data.

\begin{table}[h!]
  \centering
  \begin{tabular}{ll}
\toprule
 & malnutrition \\
\midrule
Two-Stage Training & True \\
First Marginal Bijector & BernsteinPolynomial \\
Invert First Marginal Bijector & True \\
Marginal Bijector Domain Min & -5 \\
Marginal Bijector Domain Max & 5 \\
Marginal Bijector Extrapolation & True \\
First Marginal Bijector Parameter Constraint Function & mctm.activations.get\_thetas\_constrain\_fn \\
Allow Flexible Bounds for First Marginal Bijector & False \\
Bounds constraints for First Marginal Bijector & linear \\
First Marginal Bijector Codomain Max & 5 \\
First Marginal Bijector Codomain Min & -5 \\
Parameters Function for First Marginal Bijector & parameter\_vector \\
First Marginal Bijector Parameter Shape 0 & 3 \\
First Marginal Bijector Parameter Shape 1 & 6 \\
Second Marginal Bijector & Shift \\
Invert Second Marginal Bijector & True \\
Parameters Function for Second Marginal Bijector & bernstein\_polynomial \\
Second Marginal Bijector Conditional Event Shape & 1 \\
Extrapolation for Second Marginal Bijector & True \\
Second Marginal Bijector Parameter Shape 0 & 3 \\
Polynomial Order for Second Marginal Bijector & 6 \\
Marginal Epochs & 1000 \\
Marginal Batch Size & 1024 \\
Marginal Learning Rate & 0.050 \\
Marginal Reduce LR on Plateau & False \\
Marginal Early Stopping & 50 \\
Marginal Monitor & val\_loss \\
Marginal Verbosity & True \\
Joint Epochs & 400 \\
Joint Batch Size & 512 \\
Joint Learning Rate Decay Steps & epochs \\
Joint Learning Rate Scheduler & cosine\_decay \\
Joint Early Stopping & 50 \\
Second Marginal Bijector Domain Min & 0 \\
Second Marginal Bijector Domain Max & 35 \\
\bottomrule
\end{tabular}

  \caption{Common Hyperparameters of models trained on malnutrition data.}
\end{table}

\begin{table}[h!]
  \centering
  \resizebox{\linewidth}{!}{\begin{tabular}{llll}
\toprule
Model & MCTM & HMAF (B) & HMAF (S) \\
\midrule
JIT Compile & -- & True & True \\
First Joint Bijector & ScaleMatvecLinearOperator & -- & -- \\
Invert First Joint Bijector & True & -- & -- \\
Joint Bijector Conditional Event Shape & 1 & -- & -- \\
Extrapolation for Joint Bijector & True & -- & -- \\
Parameter Shape 0 for Joint Bijector & 3 & -- & -- \\
Polynomial Order for Joint Bijector & 6 & -- & -- \\
Joint Initial Learning Rate & 0.001 & 0.001 & 0.001 \\
Type of Joint Flow & -- & masked\_autoregressive\_flow\_first\_dim\_masked & masked\_autoregressive\_flow\_first\_dim\_masked \\
Number of Flows & -- & 1 & 1 \\
Joint Bijector & -- & BernsteinPolynomial & RationalQuadraticSpline \\
Joint Bijector Range Min & -- & -- & -5 \\
Interval Width for Joint Bijector & -- & -- & 10 \\
Minimum Slope for Joint Bijector & -- & -- & 0.001 \\
Minimum Bin Width for Joint Bijector & -- & -- & 0.001 \\
Number of Bins for Joint Bijector & -- & -- & 16 \\
Number of Parameters for Joint Bijectors & -- & 32 & 47 \\
Activation Function for MAF & -- & relu & relu \\
First Hidden Units for MAF & -- & 512 & 512 \\
Second Hidden Units for MAF & -- & 512 & 512 \\
Activation Function for X0 Parameters & -- & relu & relu \\
Batch Normalization for X0 Parameters & -- & False & False \\
Dropout Rate for X0 Parameters & -- & False & False \\
First Hidden Units for X0 Network & -- & 512 & 512 \\
Second Hidden Units for X0 Network & -- & 512 & 512 \\
Nested Bernstein Bijector Domain Min & 0 & -- & -- \\
Nested Bernstein Bijector Domain Max & 35 & -- & -- \\
Joint Bijector Extrapolation & -- & False & -- \\
Nested Bernstein Bijector Domain Min & -- & -5 & -- \\
Nested Bernstein Bijector Domain Max & -- & 5 & -- \\
Invert Joint Bijector & -- & True & -- \\
Joint Bijector Codomain Min & -- & -5 & -- \\
Joint Bijector Codomain Max & -- & 5 & -- \\
Third Hidden Units for MAF & -- & 512 & 512 \\
Is Conditional for MAF & -- & True & True \\
MAF Conditional Event Shape & -- & 1 & 1 \\
Third Hidden Units for X0 Network & -- & 512 & 512 \\
Is Conditional for X0 Parameters & -- & True & True \\
X0 Parameters Conditional Event Shape & -- & 1 & 1 \\
\bottomrule
\end{tabular}
}
  \caption{Specific Hyperparameters of models trained on malnutrition data.}
\end{table}

\clearpage

\begin{landscape}
  \section{Description of Hyperparameters}
  The following table gives a brief description of the hyperparameters and maps their table names to the canonical path inside the configurations files in the git repository\footnote{\url{https://github.com/MArpogaus/hybrid_flows}}.
  \begin{longtable}{lp{0.6\linewidth}}
\toprule
Table Name & Parameter / Description \\
\midrule
\endfirsthead
\toprule
Table Name & Parameter / Description \\
\midrule
\endhead
\midrule
\multicolumn{2}{r}{Continued on next page} \\
\midrule
\endfoot
\bottomrule
\endlastfoot
\multirow[t]{2}{*}{JIT Compile} & compile\_kwargs.jit\_compile \\
 & Defines whether to compile the model using Just-In-Time compilation for performance \\
\cline{1-2}
\multirow[t]{2}{*}{Epochs} & fit\_kwargs.epochs \\
 & Maximum number of epochs \\
\cline{1-2}
\multirow[t]{2}{*}{Batch Size} & fit\_kwargs.batch\_size \\
 & Batch size used for training \\
\cline{1-2}
\multirow[t]{2}{*}{Learning Rate Decay Steps} & fit\_kwargs.learning\_rate.scheduler\_kwargs.decay\_steps \\
 & Steps after which to decay the learning rate with the scheduler \\
\cline{1-2}
\multirow[t]{2}{*}{Initial Learning Rate} & fit\_kwargs.learning\_rate.scheduler\_kwargs.initial\_learning\_rate \\
 & Initial learning rate for the learning rate scheduler \\
\cline{1-2}
\multirow[t]{2}{*}{Learning Rate Scheduler} & fit\_kwargs.learning\_rate.scheduler\_name \\
 & Name of the learning rate scheduler \\
\cline{1-2}
\multirow[t]{2}{*}{Early Stopping} & fit\_kwargs.early\_stopping \\
 & Enable early stopping \\
\cline{1-2}
\multirow[t]{2}{*}{Validation Split} & fit\_kwargs.validation\_split \\
 & Fraction of training data to be used as validation data \\
\cline{1-2}
\multirow[t]{2}{*}{Output Distribution} & model\_kwargs.distribution \\
 & Specifies the type of output distribution used in the model \\
\cline{1-2}
\multirow[t]{2}{*}{Number of Flows} & model\_kwargs.num\_layers \\
 & Total number of layers in the model \\
\cline{1-2}
\multirow[t]{2}{*}{Bijector} & model\_kwargs.bijector \\
 & Specifies the bijector type to be used in the model \\
\cline{1-2}
\multirow[t]{2}{*}{Bernstein Bijector Extrapolation} & model\_kwargs.bijector\_kwargs.extrapolation \\
 & Defines whether to allow extrapolation in the bijector \\
\cline{1-2}
\multirow[t]{2}{*}{Bernstein Bijector Domain Min} & model\_kwargs.bijector\_kwargs.domain.0 \\
 & Lower bound of the Bernstein polynomial bijector domain \\
\cline{1-2}
\multirow[t]{2}{*}{Bernstein Bijector Domain Max} & model\_kwargs.bijector\_kwargs.domain.1 \\
 & Upper bound of the Bernstein polynomial bijector domain \\
\cline{1-2}
\multirow[t]{2}{*}{Invert Bijector} & model\_kwargs.invert \\
 & Indicates if the bijector should be inverted \\
\cline{1-2}
\multirow[t]{2}{*}{Allow Flexible Bounds for Parameters} & model\_kwargs.parameters\_constraint\_fn\_kwargs.allow\_flexible\_bounds \\
 & Defines if flexible bounds are allowed for the parameters \\
\cline{1-2}
\multirow[t]{2}{*}{Bounds constraints for Parameters} & model\_kwargs.parameters\_constraint\_fn\_kwargs.bounds \\
 & Defining the type of constraints on the bounds for the parameters \\
\cline{1-2}
\multirow[t]{2}{*}{Nested Bernstein Bijector Codomain Min} & model\_kwargs.parameters\_constraint\_fn\_kwargs.low \\
 & Defines the lower bound for the corresponding parameters \\
\cline{1-2}
\multirow[t]{2}{*}{Nested Bernstein Bijector Codomain Max} & model\_kwargs.parameters\_constraint\_fn\_kwargs.high \\
 & Defines the upper bound for the corresponding parameters \\
\cline{1-2}
\multirow[t]{2}{*}{Number of Parameters} & model\_kwargs.num\_parameters \\
 & Total number of parameters in the model \\
\cline{1-2}
\multirow[t]{2}{*}{Low Constraint for Layer Overwrite -2} & model\_kwargs.layer\_overwrites.-2.parameters\_constraint\_fn\_kwargs.low \\
 & Lower constraint for parameters in layer overwrite -2 \\
\cline{1-2}
\multirow[t]{2}{*}{High Constraint for Layer Overwrite -2} & model\_kwargs.layer\_overwrites.-2.parameters\_constraint\_fn\_kwargs.high \\
 & Upper constraint for parameters in layer overwrite -2 \\
\cline{1-2}
\multirow[t]{2}{*}{Low Constraint for Layer Overwrite -1} & model\_kwargs.layer\_overwrites.-1.parameters\_constraint\_fn\_kwargs.low \\
 & Lower constraint for parameters in layer overwrite -1 \\
\cline{1-2}
\multirow[t]{2}{*}{High Constraint for Layer Overwrite -1} & model\_kwargs.layer\_overwrites.-1.parameters\_constraint\_fn\_kwargs.high \\
 & Upper constraint for parameters in layer overwrite -1 \\
\cline{1-2}
\multirow[t]{2}{*}{Activation Function} & model\_kwargs.parameters\_fn\_kwargs.activation \\
 & Activation function of the used neural network \\
\cline{1-2}
\multirow[t]{2}{*}{Batch Normalization} & model\_kwargs.parameters\_fn\_kwargs.batch\_norm \\
 & Specifies if batch normalization should be enabled in the used neural network \\
\cline{1-2}
\multirow[t]{2}{*}{Dropout Rate} & model\_kwargs.parameters\_fn\_kwargs.dropout \\
 & Dropout rate for the neural network \\
\cline{1-2}
\multirow[t]{2}{*}{First Hidden Units} & model\_kwargs.parameters\_fn\_kwargs.hidden\_units.0 \\
 & Number of units in the first hidden layer of the neural network \\
\cline{1-2}
\multirow[t]{2}{*}{Second Hidden Units} & model\_kwargs.parameters\_fn\_kwargs.hidden\_units.1 \\
 & Number of units in the second hidden layer of the neural network \\
\cline{1-2}
\multirow[t]{2}{*}{Third Hidden Units} & model\_kwargs.parameters\_fn\_kwargs.hidden\_units.2 \\
 & Number of units in the third hidden layer of the neural network \\
\cline{1-2}
\multirow[t]{2}{*}{Conditional} & model\_kwargs.parameters\_fn\_kwargs.conditional \\
 & Indicates if the parameter functions should get the covariates as input \\
\cline{1-2}
\multirow[t]{2}{*}{Conditional Event Shape} & model\_kwargs.parameters\_fn\_kwargs.conditional\_event\_shape \\
 & Shape of the covariate vector \\
\cline{1-2}
\multirow[t]{2}{*}{JIT Compile} & compile\_kwargs.jit\_compile \\
 & Defines whether to compile the model using Just-In-Time compilation for performance \\
\cline{1-2}
\multirow[t]{2}{*}{Epochs} & fit\_kwargs.epochs \\
 & Maximum number of epochs \\
\cline{1-2}
\multirow[t]{2}{*}{Batch Size} & fit\_kwargs.batch\_size \\
 & Batch size used for training \\
\cline{1-2}
\multirow[t]{2}{*}{Learning Rate Decay Steps} & fit\_kwargs.learning\_rate.scheduler\_kwargs.decay\_steps \\
 & Steps after which to decay the learning rate with the scheduler \\
\cline{1-2}
\multirow[t]{2}{*}{Initial Learning Rate} & fit\_kwargs.learning\_rate.scheduler\_kwargs.initial\_learning\_rate \\
 & Initial learning rate for the learning rate scheduler \\
\cline{1-2}
\multirow[t]{2}{*}{Learning Rate Scheduler} & fit\_kwargs.learning\_rate.scheduler\_name \\
 & Name of the learning rate scheduler \\
\cline{1-2}
\multirow[t]{2}{*}{Early Stopping} & fit\_kwargs.early\_stopping \\
 & Enable early stopping \\
\cline{1-2}
\multirow[t]{2}{*}{Validation Split} & fit\_kwargs.validation\_split \\
 & Fraction of training data to be used as validation data \\
\cline{1-2}
\multirow[t]{2}{*}{Output Distribution} & model\_kwargs.distribution \\
 & Specifies the type of output distribution used in the model \\
\cline{1-2}
\multirow[t]{2}{*}{Number of Flows} & model\_kwargs.num\_layers \\
 & Total number of layers in the model \\
\cline{1-2}
\multirow[t]{2}{*}{Bijector} & model\_kwargs.bijector \\
 & Specifies the bijector type to be used in the model \\
\cline{1-2}
\multirow[t]{2}{*}{Number of Parameters} & model\_kwargs.num\_parameters \\
 & Total number of parameters in the model \\
\cline{1-2}
\multirow[t]{2}{*}{Activation Function} & model\_kwargs.parameters\_fn\_kwargs.activation \\
 & Activation function of the used neural network \\
\cline{1-2}
\multirow[t]{2}{*}{Batch Normalization} & model\_kwargs.parameters\_fn\_kwargs.batch\_norm \\
 & Specifies if batch normalization should be enabled in the used neural network \\
\cline{1-2}
\multirow[t]{2}{*}{Dropout Rate} & model\_kwargs.parameters\_fn\_kwargs.dropout \\
 & Dropout rate for the neural network \\
\cline{1-2}
\multirow[t]{2}{*}{First Hidden Units} & model\_kwargs.parameters\_fn\_kwargs.hidden\_units.0 \\
 & Number of units in the first hidden layer of the neural network \\
\cline{1-2}
\multirow[t]{2}{*}{Second Hidden Units} & model\_kwargs.parameters\_fn\_kwargs.hidden\_units.1 \\
 & Number of units in the second hidden layer of the neural network \\
\cline{1-2}
\multirow[t]{2}{*}{Third Hidden Units} & model\_kwargs.parameters\_fn\_kwargs.hidden\_units.2 \\
 & Number of units in the third hidden layer of the neural network \\
\cline{1-2}
\multirow[t]{2}{*}{Spline Bijector Domain Min} & model\_kwargs.bijector\_kwargs.range\_min \\
 & Lower bound of the quadratic spline bijector domain \\
\cline{1-2}
\multirow[t]{2}{*}{Interval Width for Nested Spline Bijector} & model\_kwargs.parameters\_constraint\_fn\_kwargs.interval\_width \\
 & Specifies the interval width for constraint parameters \\
\cline{1-2}
\multirow[t]{2}{*}{Minimum Slope for Nested Spline Bijector} & model\_kwargs.parameters\_constraint\_fn\_kwargs.min\_slope \\
 & Defines the minimum slope constraint for parameters \\
\cline{1-2}
\multirow[t]{2}{*}{Minimum Bin Width for Nested Spline Bijector} & model\_kwargs.parameters\_constraint\_fn\_kwargs.min\_bin\_width \\
 & Defines the minimum bin width for the nested Spline bijector \\
\cline{1-2}
\multirow[t]{2}{*}{Number of Bins for Nested Spline Bijector} & model\_kwargs.parameters\_constraint\_fn\_kwargs.nbins \\
 & Defines the number of bins used by the nested bijector \\
\cline{1-2}
\multirow[t]{2}{*}{Conditional} & model\_kwargs.parameters\_fn\_kwargs.conditional \\
 & Indicates if the parameter functions should get the covariates as input \\
\cline{1-2}
\multirow[t]{2}{*}{Conditional Event Shape} & model\_kwargs.parameters\_fn\_kwargs.conditional\_event\_shape \\
 & Shape of the covariate vector \\
\cline{1-2}
\multirow[t]{2}{*}{JIT Compile} & compile\_kwargs.jit\_compile \\
 & Defines whether to compile the model using Just-In-Time compilation for performance \\
\cline{1-2}
\multirow[t]{2}{*}{Epochs} & fit\_kwargs.epochs \\
 & Maximum number of epochs \\
\cline{1-2}
\multirow[t]{2}{*}{Batch Size} & fit\_kwargs.batch\_size \\
 & Batch size used for training \\
\cline{1-2}
\multirow[t]{2}{*}{Learning Rate Decay Steps} & fit\_kwargs.learning\_rate.scheduler\_kwargs.decay\_steps \\
 & Steps after which to decay the learning rate with the scheduler \\
\cline{1-2}
\multirow[t]{2}{*}{Initial Learning Rate} & fit\_kwargs.learning\_rate.scheduler\_kwargs.initial\_learning\_rate \\
 & Initial learning rate for the learning rate scheduler \\
\cline{1-2}
\multirow[t]{2}{*}{Learning Rate Scheduler} & fit\_kwargs.learning\_rate.scheduler\_name \\
 & Name of the learning rate scheduler \\
\cline{1-2}
\multirow[t]{2}{*}{Early Stopping} & fit\_kwargs.early\_stopping \\
 & Enable early stopping \\
\cline{1-2}
\multirow[t]{2}{*}{Validation Split} & fit\_kwargs.validation\_split \\
 & Fraction of training data to be used as validation data \\
\cline{1-2}
\multirow[t]{2}{*}{Output Distribution} & model\_kwargs.distribution \\
 & Specifies the type of output distribution used in the model \\
\cline{1-2}
\multirow[t]{2}{*}{Number of Flows} & model\_kwargs.num\_layers \\
 & Total number of layers in the model \\
\cline{1-2}
\multirow[t]{2}{*}{Bijector} & model\_kwargs.bijector \\
 & Specifies the bijector type to be used in the model \\
\cline{1-2}
\multirow[t]{2}{*}{Bernstein Bijector Extrapolation} & model\_kwargs.bijector\_kwargs.extrapolation \\
 & Defines whether to allow extrapolation in the bijector \\
\cline{1-2}
\multirow[t]{2}{*}{Bernstein Bijector Domain Min} & model\_kwargs.bijector\_kwargs.domain.0 \\
 & Lower bound of the Bernstein polynomial bijector domain \\
\cline{1-2}
\multirow[t]{2}{*}{Bernstein Bijector Domain Max} & model\_kwargs.bijector\_kwargs.domain.1 \\
 & Upper bound of the Bernstein polynomial bijector domain \\
\cline{1-2}
\multirow[t]{2}{*}{Invert Bijector} & model\_kwargs.invert \\
 & Indicates if the bijector should be inverted \\
\cline{1-2}
\multirow[t]{2}{*}{Allow Flexible Bounds for Parameters} & model\_kwargs.parameters\_constraint\_fn\_kwargs.allow\_flexible\_bounds \\
 & Defines if flexible bounds are allowed for the parameters \\
\cline{1-2}
\multirow[t]{2}{*}{Bounds constraints for Parameters} & model\_kwargs.parameters\_constraint\_fn\_kwargs.bounds \\
 & Defining the type of constraints on the bounds for the parameters \\
\cline{1-2}
\multirow[t]{2}{*}{Nested Bernstein Bijector Codomain Min} & model\_kwargs.parameters\_constraint\_fn\_kwargs.low \\
 & Defines the lower bound for the corresponding parameters \\
\cline{1-2}
\multirow[t]{2}{*}{Nested Bernstein Bijector Codomain Max} & model\_kwargs.parameters\_constraint\_fn\_kwargs.high \\
 & Defines the upper bound for the corresponding parameters \\
\cline{1-2}
\multirow[t]{2}{*}{Number of Parameters} & model\_kwargs.num\_parameters \\
 & Total number of parameters in the model \\
\cline{1-2}
\multirow[t]{2}{*}{Activation Function} & model\_kwargs.parameters\_fn\_kwargs.activation \\
 & Activation function of the used neural network \\
\cline{1-2}
\multirow[t]{2}{*}{First Hidden Units} & model\_kwargs.parameters\_fn\_kwargs.hidden\_units.0 \\
 & Number of units in the first hidden layer of the neural network \\
\cline{1-2}
\multirow[t]{2}{*}{Second Hidden Units} & model\_kwargs.parameters\_fn\_kwargs.hidden\_units.1 \\
 & Number of units in the second hidden layer of the neural network \\
\cline{1-2}
\multirow[t]{2}{*}{Third Hidden Units} & model\_kwargs.parameters\_fn\_kwargs.hidden\_units.2 \\
 & Number of units in the third hidden layer of the neural network \\
\cline{1-2}
\multirow[t]{2}{*}{Layer Overwrite 0 Domain Min} & model\_kwargs.layer\_overwrites.0.bijector\_kwargs.domain.0 \\
 & Lower boundary for the domain in layer overwrite 0 \\
\cline{1-2}
\multirow[t]{2}{*}{Layer Overwrite 0 Domain Max} & model\_kwargs.layer\_overwrites.0.bijector\_kwargs.domain.1 \\
 & Upper boundary for the domain in layer overwrite 0 \\
\cline{1-2}
\multirow[t]{2}{*}{Conditional} & model\_kwargs.parameters\_fn\_kwargs.conditional \\
 & Indicates if the parameter functions should get the covariates as input \\
\cline{1-2}
\multirow[t]{2}{*}{Conditional Event Shape} & model\_kwargs.parameters\_fn\_kwargs.conditional\_event\_shape \\
 & Shape of the covariate vector \\
\cline{1-2}
\multirow[t]{2}{*}{JIT Compile} & compile\_kwargs.jit\_compile \\
 & Defines whether to compile the model using Just-In-Time compilation for performance \\
\cline{1-2}
\multirow[t]{2}{*}{Epochs} & fit\_kwargs.epochs \\
 & Maximum number of epochs \\
\cline{1-2}
\multirow[t]{2}{*}{Batch Size} & fit\_kwargs.batch\_size \\
 & Batch size used for training \\
\cline{1-2}
\multirow[t]{2}{*}{Learning Rate Decay Steps} & fit\_kwargs.learning\_rate.scheduler\_kwargs.decay\_steps \\
 & Steps after which to decay the learning rate with the scheduler \\
\cline{1-2}
\multirow[t]{2}{*}{Initial Learning Rate} & fit\_kwargs.learning\_rate.scheduler\_kwargs.initial\_learning\_rate \\
 & Initial learning rate for the learning rate scheduler \\
\cline{1-2}
\multirow[t]{2}{*}{Learning Rate Scheduler} & fit\_kwargs.learning\_rate.scheduler\_name \\
 & Name of the learning rate scheduler \\
\cline{1-2}
\multirow[t]{2}{*}{Early Stopping} & fit\_kwargs.early\_stopping \\
 & Enable early stopping \\
\cline{1-2}
\multirow[t]{2}{*}{Validation Split} & fit\_kwargs.validation\_split \\
 & Fraction of training data to be used as validation data \\
\cline{1-2}
\multirow[t]{2}{*}{Output Distribution} & model\_kwargs.distribution \\
 & Specifies the type of output distribution used in the model \\
\cline{1-2}
\multirow[t]{2}{*}{Number of Flows} & model\_kwargs.num\_layers \\
 & Total number of layers in the model \\
\cline{1-2}
\multirow[t]{2}{*}{Bijector} & model\_kwargs.bijector \\
 & Specifies the bijector type to be used in the model \\
\cline{1-2}
\multirow[t]{2}{*}{Number of Parameters} & model\_kwargs.num\_parameters \\
 & Total number of parameters in the model \\
\cline{1-2}
\multirow[t]{2}{*}{Activation Function} & model\_kwargs.parameters\_fn\_kwargs.activation \\
 & Activation function of the used neural network \\
\cline{1-2}
\multirow[t]{2}{*}{First Hidden Units} & model\_kwargs.parameters\_fn\_kwargs.hidden\_units.0 \\
 & Number of units in the first hidden layer of the neural network \\
\cline{1-2}
\multirow[t]{2}{*}{Second Hidden Units} & model\_kwargs.parameters\_fn\_kwargs.hidden\_units.1 \\
 & Number of units in the second hidden layer of the neural network \\
\cline{1-2}
\multirow[t]{2}{*}{Third Hidden Units} & model\_kwargs.parameters\_fn\_kwargs.hidden\_units.2 \\
 & Number of units in the third hidden layer of the neural network \\
\cline{1-2}
\multirow[t]{2}{*}{Spline Bijector Domain Min} & model\_kwargs.bijector\_kwargs.range\_min \\
 & Lower bound of the quadratic spline bijector domain \\
\cline{1-2}
\multirow[t]{2}{*}{Interval Width for Nested Spline Bijector} & model\_kwargs.parameters\_constraint\_fn\_kwargs.interval\_width \\
 & Specifies the interval width for constraint parameters \\
\cline{1-2}
\multirow[t]{2}{*}{Minimum Slope for Nested Spline Bijector} & model\_kwargs.parameters\_constraint\_fn\_kwargs.min\_slope \\
 & Defines the minimum slope constraint for parameters \\
\cline{1-2}
\multirow[t]{2}{*}{Minimum Bin Width for Nested Spline Bijector} & model\_kwargs.parameters\_constraint\_fn\_kwargs.min\_bin\_width \\
 & Defines the minimum bin width for the nested Spline bijector \\
\cline{1-2}
\multirow[t]{2}{*}{Number of Bins for Nested Spline Bijector} & model\_kwargs.parameters\_constraint\_fn\_kwargs.nbins \\
 & Defines the number of bins used by the nested bijector \\
\cline{1-2}
\multirow[t]{2}{*}{Conditional} & model\_kwargs.parameters\_fn\_kwargs.conditional \\
 & Indicates if the parameter functions should get the covariates as input \\
\cline{1-2}
\multirow[t]{2}{*}{Conditional Event Shape} & model\_kwargs.parameters\_fn\_kwargs.conditional\_event\_shape \\
 & Shape of the covariate vector \\
\cline{1-2}
\multirow[t]{2}{*}{JIT Compile} & compile\_kwargs.jit\_compile \\
 & Defines whether to compile the model using Just-In-Time compilation for performance \\
\cline{1-2}
\multirow[t]{2}{*}{Epochs} & fit\_kwargs.epochs \\
 & Maximum number of epochs \\
\cline{1-2}
\multirow[t]{2}{*}{Batch Size} & fit\_kwargs.batch\_size \\
 & Batch size used for training \\
\cline{1-2}
\multirow[t]{2}{*}{Learning Rate Decay Steps} & fit\_kwargs.learning\_rate.scheduler\_kwargs.decay\_steps \\
 & Steps after which to decay the learning rate with the scheduler \\
\cline{1-2}
\multirow[t]{2}{*}{Initial Learning Rate} & fit\_kwargs.learning\_rate.scheduler\_kwargs.initial\_learning\_rate \\
 & Initial learning rate for the learning rate scheduler \\
\cline{1-2}
\multirow[t]{2}{*}{Learning Rate Scheduler} & fit\_kwargs.learning\_rate.scheduler\_name \\
 & Name of the learning rate scheduler \\
\cline{1-2}
\multirow[t]{2}{*}{Early Stopping} & fit\_kwargs.early\_stopping \\
 & Enable early stopping \\
\cline{1-2}
\multirow[t]{2}{*}{Validation Split} & fit\_kwargs.validation\_split \\
 & Fraction of training data to be used as validation data \\
\cline{1-2}
\multirow[t]{2}{*}{Output Distribution} & model\_kwargs.distribution \\
 & Specifies the type of output distribution used in the model \\
\cline{1-2}
\multirow[t]{2}{*}{Activation Function} & model\_kwargs.parameters\_fn\_kwargs.activation \\
 & Activation function of the used neural network \\
\cline{1-2}
\multirow[t]{2}{*}{Batch Normalization} & model\_kwargs.parameters\_fn\_kwargs.batch\_norm \\
 & Specifies if batch normalization should be enabled in the used neural network \\
\cline{1-2}
\multirow[t]{2}{*}{Dropout Rate} & model\_kwargs.parameters\_fn\_kwargs.dropout \\
 & Dropout rate for the neural network \\
\cline{1-2}
\multirow[t]{2}{*}{First Hidden Units} & model\_kwargs.parameters\_fn\_kwargs.hidden\_units.0 \\
 & Number of units in the first hidden layer of the neural network \\
\cline{1-2}
\multirow[t]{2}{*}{Second Hidden Units} & model\_kwargs.parameters\_fn\_kwargs.hidden\_units.1 \\
 & Number of units in the second hidden layer of the neural network \\
\cline{1-2}
\multirow[t]{2}{*}{Conditional} & model\_kwargs.parameters\_fn\_kwargs.conditional \\
 & Indicates if the parameter functions should get the covariates as input \\
\cline{1-2}
\multirow[t]{2}{*}{Conditional Event Shape} & model\_kwargs.parameters\_fn\_kwargs.conditional\_event\_shape \\
 & Shape of the covariate vector \\
\cline{1-2}
\multirow[t]{2}{*}{Epochs} & fit\_kwargs.epochs \\
 & Maximum number of epochs \\
\cline{1-2}
\multirow[t]{2}{*}{Batch Size} & fit\_kwargs.batch\_size \\
 & Batch size used for training \\
\cline{1-2}
\multirow[t]{2}{*}{Learning Rate Decay Steps} & fit\_kwargs.learning\_rate.scheduler\_kwargs.decay\_steps \\
 & Steps after which to decay the learning rate with the scheduler \\
\cline{1-2}
\multirow[t]{2}{*}{Initial Learning Rate} & fit\_kwargs.learning\_rate.scheduler\_kwargs.initial\_learning\_rate \\
 & Initial learning rate for the learning rate scheduler \\
\cline{1-2}
\multirow[t]{2}{*}{Learning Rate Scheduler} & fit\_kwargs.learning\_rate.scheduler\_name \\
 & Name of the learning rate scheduler \\
\cline{1-2}
\multirow[t]{2}{*}{Validation Split} & fit\_kwargs.validation\_split \\
 & Fraction of training data to be used as validation data \\
\cline{1-2}
\multirow[t]{2}{*}{Two-Stage Training} & two\_stage\_training \\
 & Indicates if the model should be trained in two stages \\
\cline{1-2}
\multirow[t]{2}{*}{First Marginal Bijector} & model\_kwargs.marginal\_bijectors.0.bijector \\
 & Specifies the type of the first marginal bijector \\
\cline{1-2}
\multirow[t]{2}{*}{Invert First Marginal Bijector} & model\_kwargs.marginal\_bijectors.0.invert \\
 & Indicates if the first marginal bijector should allow inversion \\
\cline{1-2}
\multirow[t]{2}{*}{Marginal Bijector Domain Min} & model\_kwargs.marginal\_bijectors.0.bijector\_kwargs.domain.0 \\
 & Lower boundary for the domain of the first marginal bijector \\
\cline{1-2}
\multirow[t]{2}{*}{Marginal Bijector Domain Max} & model\_kwargs.marginal\_bijectors.0.bijector\_kwargs.domain.1 \\
 & Upper boundary for the domain of the first marginal bijector \\
\cline{1-2}
\multirow[t]{2}{*}{Marginal Bijector Extrapolation} & model\_kwargs.marginal\_bijectors.0.bijector\_kwargs.extrapolation \\
 & Defines whether to allow extrapolation in the first marginal bijector \\
\cline{1-2}
\multirow[t]{2}{*}{First Marginal Bijector Parameter Constraint Function} & model\_kwargs.marginal\_bijectors.0.parameters\_constraint\_fn \\
 & Constraint function for the parameters of the first marginal bijector \\
\cline{1-2}
\multirow[t]{2}{*}{Allow Flexible Bounds for First Marginal Bijector} & model\_kwargs.marginal\_bijectors.0.parameters\_constraint\_fn\_kwargs.allow\_flexible\_bounds \\
 & Defines if flexible bounds are allowed for the first marginal bijector \\
\cline{1-2}
\multirow[t]{2}{*}{Bounds constraints for First Marginal Bijector} & model\_kwargs.marginal\_bijectors.0.parameters\_constraint\_fn\_kwargs.bounds \\
 & Defining the constraints on the bounds for the first marginal bijector \\
\cline{1-2}
\multirow[t]{2}{*}{First Marginal Bijector Codomain Max} & model\_kwargs.marginal\_bijectors.0.parameters\_constraint\_fn\_kwargs.high \\
 & Defines the upper bound for parameters in the first marginal bijector \\
\cline{1-2}
\multirow[t]{2}{*}{First Marginal Bijector Codomain Min} & model\_kwargs.marginal\_bijectors.0.parameters\_constraint\_fn\_kwargs.low \\
 & Defines the lower bound for parameters in the first marginal bijector \\
\cline{1-2}
\multirow[t]{2}{*}{Parameters Function for First Marginal Bijector} & model\_kwargs.marginal\_bijectors.0.parameters\_fn \\
 & Function for calculating parameters of the first marginal bijector \\
\cline{1-2}
\multirow[t]{2}{*}{First Marginal Bijector Parameter Shape 0} & model\_kwargs.marginal\_bijectors.0.parameters\_fn\_kwargs.parameter\_shape.0 \\
 & Defines the parameter shape dimension 0 for the first marginal bijector \\
\cline{1-2}
\multirow[t]{2}{*}{First Marginal Bijector Parameter Shape 1} & model\_kwargs.marginal\_bijectors.0.parameters\_fn\_kwargs.parameter\_shape.1 \\
 & Defines the parameter shape dimension 1 for the first marginal bijector \\
\cline{1-2}
\multirow[t]{2}{*}{First Joint Bijector} & model\_kwargs.joint\_bijectors.0.bijector \\
 & Specifies the first bijector type for the joint distribution \\
\cline{1-2}
\multirow[t]{2}{*}{Invert First Joint Bijector} & model\_kwargs.joint\_bijectors.0.invert \\
 & Indicates if the first joint bijector should allow inversion \\
\cline{1-2}
\multirow[t]{2}{*}{Parameter Shape 0 for Joint Bijector} & model\_kwargs.joint\_bijectors.0.parameters\_fn\_kwargs.parameter\_shape.0 \\
 & Defines the parameter shape dimension 0 for the joint bijector \\
\cline{1-2}
\multirow[t]{2}{*}{Early Stopping} & fit\_kwargs.early\_stopping \\
 & Enable early stopping \\
\cline{1-2}
\multirow[t]{2}{*}{Second Marginal Bijector} & model\_kwargs.marginal\_bijectors.1.bijector \\
 & Specifies the type of second marginal bijector \\
\cline{1-2}
\multirow[t]{2}{*}{Invert Second Marginal Bijector} & model\_kwargs.marginal\_bijectors.1.invert \\
 & Indicates if the second marginal bijector should allow inversion \\
\cline{1-2}
\multirow[t]{2}{*}{Parameters Function for Second Marginal Bijector} & model\_kwargs.marginal\_bijectors.1.parameters\_fn \\
 & Function for calculating parameters of the second marginal bijector \\
\cline{1-2}
\multirow[t]{2}{*}{Second Marginal Bijector Conditional Event Shape} & model\_kwargs.marginal\_bijectors.1.parameters\_fn\_kwargs.conditional\_event\_shape \\
 & Shape of the conditional events for the second marginal bijector \\
\cline{1-2}
\multirow[t]{2}{*}{Extrapolation for Second Marginal Bijector} & model\_kwargs.marginal\_bijectors.1.parameters\_fn\_kwargs.extrapolation \\
 & Defines whether to allow extrapolation in the second marginal bijector \\
\cline{1-2}
\multirow[t]{2}{*}{Second Marginal Bijector Parameter Shape 0} & model\_kwargs.marginal\_bijectors.1.parameters\_fn\_kwargs.parameter\_shape.0 \\
 & Defines the parameter shape dimension 0 for the second marginal bijector \\
\cline{1-2}
\multirow[t]{2}{*}{Polynomial Order for Second Marginal Bijector} & model\_kwargs.marginal\_bijectors.1.parameters\_fn\_kwargs.polynomial\_order \\
 & Defines the order of the polynomial for computations in the second marginal bijector \\
\cline{1-2}
\multirow[t]{2}{*}{Joint Bijector Conditional Event Shape} & model\_kwargs.joint\_bijectors.0.parameters\_fn\_kwargs.conditional\_event\_shape \\
 & Shape of the conditional events for the joint bijector \\
\cline{1-2}
\multirow[t]{2}{*}{Extrapolation for Joint Bijector} & model\_kwargs.joint\_bijectors.0.parameters\_fn\_kwargs.extrapolation \\
 & Defines whether to allow extrapolation in the joint bijector \\
\cline{1-2}
\multirow[t]{2}{*}{Polynomial Order for Joint Bijector} & model\_kwargs.joint\_bijectors.0.parameters\_fn\_kwargs.polynomial\_order \\
 & Defines the order of the polynomial for computations in the joint bijector \\
\cline{1-2}
\multirow[t]{2}{*}{Parameter Shape 1 for Joint Bijector} & model\_kwargs.joint\_bijectors.0.parameters\_fn\_kwargs.parameter\_shape.1 \\
 & Defines the parameter shape dimension 1 for the joint bijector \\
\cline{1-2}
\multirow[t]{2}{*}{Two-Stage Training} & two\_stage\_training \\
 & Indicates if the model should be trained in two stages \\
\cline{1-2}
\multirow[t]{2}{*}{First Marginal Bijector} & model\_kwargs.marginal\_bijectors.0.bijector \\
 & Specifies the type of the first marginal bijector \\
\cline{1-2}
\multirow[t]{2}{*}{Invert First Marginal Bijector} & model\_kwargs.marginal\_bijectors.0.invert \\
 & Indicates if the first marginal bijector should allow inversion \\
\cline{1-2}
\multirow[t]{2}{*}{Marginal Bijector Domain Min} & model\_kwargs.marginal\_bijectors.0.bijector\_kwargs.domain.0 \\
 & Lower boundary for the domain of the first marginal bijector \\
\cline{1-2}
\multirow[t]{2}{*}{Marginal Bijector Domain Max} & model\_kwargs.marginal\_bijectors.0.bijector\_kwargs.domain.1 \\
 & Upper boundary for the domain of the first marginal bijector \\
\cline{1-2}
\multirow[t]{2}{*}{Marginal Bijector Extrapolation} & model\_kwargs.marginal\_bijectors.0.bijector\_kwargs.extrapolation \\
 & Defines whether to allow extrapolation in the first marginal bijector \\
\cline{1-2}
\multirow[t]{2}{*}{First Marginal Bijector Parameter Constraint Function} & model\_kwargs.marginal\_bijectors.0.parameters\_constraint\_fn \\
 & Constraint function for the parameters of the first marginal bijector \\
\cline{1-2}
\multirow[t]{2}{*}{Allow Flexible Bounds for First Marginal Bijector} & model\_kwargs.marginal\_bijectors.0.parameters\_constraint\_fn\_kwargs.allow\_flexible\_bounds \\
 & Defines if flexible bounds are allowed for the first marginal bijector \\
\cline{1-2}
\multirow[t]{2}{*}{Bounds constraints for First Marginal Bijector} & model\_kwargs.marginal\_bijectors.0.parameters\_constraint\_fn\_kwargs.bounds \\
 & Defining the constraints on the bounds for the first marginal bijector \\
\cline{1-2}
\multirow[t]{2}{*}{First Marginal Bijector Codomain Max} & model\_kwargs.marginal\_bijectors.0.parameters\_constraint\_fn\_kwargs.high \\
 & Defines the upper bound for parameters in the first marginal bijector \\
\cline{1-2}
\multirow[t]{2}{*}{First Marginal Bijector Codomain Min} & model\_kwargs.marginal\_bijectors.0.parameters\_constraint\_fn\_kwargs.low \\
 & Defines the lower bound for parameters in the first marginal bijector \\
\cline{1-2}
\multirow[t]{2}{*}{First Marginal Bijector Parameter Shape 0} & model\_kwargs.marginal\_bijectors.0.parameters\_fn\_kwargs.parameter\_shape.0 \\
 & Defines the parameter shape dimension 0 for the first marginal bijector \\
\cline{1-2}
\multirow[t]{2}{*}{First Marginal Bijector Parameter Shape 1} & model\_kwargs.marginal\_bijectors.0.parameters\_fn\_kwargs.parameter\_shape.1 \\
 & Defines the parameter shape dimension 1 for the first marginal bijector \\
\cline{1-2}
\multirow[t]{2}{*}{First Joint Bijector} & model\_kwargs.joint\_bijectors.0.bijector \\
 & Specifies the first bijector type for the joint distribution \\
\cline{1-2}
\multirow[t]{2}{*}{Parameter Shape 0 for Joint Bijector} & model\_kwargs.joint\_bijectors.0.parameters\_fn\_kwargs.parameter\_shape.0 \\
 & Defines the parameter shape dimension 0 for the joint bijector \\
\cline{1-2}
\multirow[t]{2}{*}{Marginal Epochs} & fit\_kwargs.0.epochs \\
 & Number of epochs for training the marginal distribution model \\
\cline{1-2}
\multirow[t]{2}{*}{Marginal Batch Size} & fit\_kwargs.0.batch\_size \\
 & Batch size for training the marginal distribution model \\
\cline{1-2}
\multirow[t]{2}{*}{Marginal Learning Rate} & fit\_kwargs.0.learning\_rate \\
 & Learning rate used to optimize the marginal distribution \\
\cline{1-2}
\multirow[t]{2}{*}{Marginal Reduce LR on Plateau} & fit\_kwargs.0.reduce\_lr\_on\_plateau \\
 & Whether to reduce learning rate on a plateau during training \\
\cline{1-2}
\multirow[t]{2}{*}{Marginal Early Stopping} & fit\_kwargs.0.early\_stopping \\
 & Enables early stopping for the marginal model training \\
\cline{1-2}
\multirow[t]{2}{*}{Marginal Monitor} & fit\_kwargs.0.monitor \\
 & Metric to monitor during training for early stopping \\
\cline{1-2}
\multirow[t]{2}{*}{Marginal Verbosity} & fit\_kwargs.0.verbose \\
 & Verbosity mode for the marginal model training \\
\cline{1-2}
\multirow[t]{2}{*}{Marginal Validation Split} & fit\_kwargs.0.validation\_split \\
 & Fraction of training data to be used as validation data for the marginal model \\
\cline{1-2}
\multirow[t]{2}{*}{Joint Epochs} & fit\_kwargs.1.epochs \\
 & Maximum number of epochs for training the joint distribution model \\
\cline{1-2}
\multirow[t]{2}{*}{Joint Batch Size} & fit\_kwargs.1.batch\_size \\
 & Batch size for training the joint distribution model \\
\cline{1-2}
\multirow[t]{2}{*}{Joint Learning Rate Decay Steps} & fit\_kwargs.1.learning\_rate.scheduler\_kwargs.decay\_steps \\
 & Total number of decay steps for the learning rate scheduler of the joint model \\
\cline{1-2}
\multirow[t]{2}{*}{Joint Initial Learning Rate} & fit\_kwargs.1.learning\_rate.scheduler\_kwargs.initial\_learning\_rate \\
 & Initial learning rate for the learning rate scheduler for the joint model \\
\cline{1-2}
\multirow[t]{2}{*}{Joint Learning Rate Scheduler} & fit\_kwargs.1.learning\_rate.scheduler\_name \\
 & Name of the learning rate scheduler for the joint model \\
\cline{1-2}
\multirow[t]{2}{*}{Joint Early Stopping} & fit\_kwargs.1.early\_stopping \\
 & Enables early stopping for the joint model training \\
\cline{1-2}
\multirow[t]{2}{*}{Joint Validation Split} & fit\_kwargs.1.validation\_split \\
 & Fraction of training data to be used as validation data for the joint model \\
\cline{1-2}
\multirow[t]{2}{*}{Masked Dimensions} & model\_kwargs.joint\_bijectors.0.bijector\_kwargs.num\_masked \\
 & Number of deimensions mask in the Coupling Flow \\
\cline{1-2}
\multirow[t]{2}{*}{First Nested Bijector} & model\_kwargs.joint\_bijectors.0.nested\_bijector.bijector \\
 & Specifies the type of nested bijector used in the first joint bijector \\
\cline{1-2}
\multirow[t]{2}{*}{Invert Nested Bijector} & model\_kwargs.joint\_bijectors.0.nested\_bijector.invert \\
 & Specifies if the nested bijector should be invertible \\
\cline{1-2}
\multirow[t]{2}{*}{Nested Bernstein Bijector Domain Min} & model\_kwargs.joint\_bijectors.0.nested\_bijector.bijector\_kwargs.domain.0 \\
 & Lower boundary for the nested Bernstein bijector domain \\
\cline{1-2}
\multirow[t]{2}{*}{Nested Bernstein Bijector Domain Max} & model\_kwargs.joint\_bijectors.0.nested\_bijector.bijector\_kwargs.domain.1 \\
 & Upper boundary for the nested Bernstein bijector domain \\
\cline{1-2}
\multirow[t]{2}{*}{Nested Bernstein Bijector Extrapolation} & model\_kwargs.joint\_bijectors.0.nested\_bijector.bijector\_kwargs.extrapolation \\
 & Defines whether to allow extrapolation in the nested Bernstein bijector \\
\cline{1-2}
\multirow[t]{2}{*}{Nested Bijector Parameter Constraint Function} & model\_kwargs.joint\_bijectors.0.nested\_bijector.parameters\_constraint\_fn \\
 & Constraint function applied to the parameters of the nested bijector \\
\cline{1-2}
\multirow[t]{2}{*}{Bounds constraints of Nested Bernstein Bijector} & model\_kwargs.joint\_bijectors.0.nested\_bijector.parameters\_constraint\_fn\_kwargs.bounds \\
 & Defining the type of constraints on the bounds of the nested Bernstein bijector \\
\cline{1-2}
\multirow[t]{2}{*}{Nested Bernstein Bijector Codomain Max} & model\_kwargs.joint\_bijectors.0.nested\_bijector.parameters\_constraint\_fn\_kwargs.high \\
 & Defines the upper bound for the corresponding parameters \\
\cline{1-2}
\multirow[t]{2}{*}{Nested Bernstein Bijector Codomain Min} & model\_kwargs.joint\_bijectors.0.nested\_bijector.parameters\_constraint\_fn\_kwargs.low \\
 & Defines the lower bound for the corresponding parameters \\
\cline{1-2}
\multirow[t]{2}{*}{Activation Function for Joint Bijector} & model\_kwargs.joint\_bijectors.0.parameters\_fn\_kwargs.activation \\
 & Activation function used in the neural network used to estimate the joint bijector parameters \\
\cline{1-2}
\multirow[t]{2}{*}{Batch Normalization for Joint Bijector} & model\_kwargs.joint\_bijectors.0.parameters\_fn\_kwargs.batch\_norm \\
 & Specifies if batch normalization should be applied in the joint bijector \\
\cline{1-2}
\multirow[t]{2}{*}{Dropout Rate for Joint Bijector} & model\_kwargs.joint\_bijectors.0.parameters\_fn\_kwargs.dropout \\
 & Dropout rate to apply in the joint bijector neural network \\
\cline{1-2}
\multirow[t]{2}{*}{First Hidden Units for Joint Bijector} & model\_kwargs.joint\_bijectors.0.parameters\_fn\_kwargs.hidden\_units.0 \\
 & Number of units in the first hidden layer of the joint bijector \\
\cline{1-2}
\multirow[t]{2}{*}{Input Shape 0 for Joint Bijector} & model\_kwargs.joint\_bijectors.0.parameters\_fn\_kwargs.input\_shape.0 \\
 & Defines the input shape dimension 0 for the joint bijector \\
\cline{1-2}
\multirow[t]{2}{*}{JIT Compile} & compile\_kwargs.jit\_compile \\
 & Defines whether to compile the model using Just-In-Time compilation for performance \\
\cline{1-2}
\multirow[t]{2}{*}{Parameters Function for First Marginal Bijector} & model\_kwargs.marginal\_bijectors.0.parameters\_fn \\
 & Function for calculating parameters of the first marginal bijector \\
\cline{1-2}
\multirow[t]{2}{*}{Joint Bijector Conditional Event Shape} & model\_kwargs.joint\_bijectors.0.parameters\_fn\_kwargs.conditional\_event\_shape \\
 & Shape of the conditional events for the joint bijector \\
\cline{1-2}
\multirow[t]{2}{*}{Extrapolation for First Marginal Bijector} & model\_kwargs.marginal\_bijectors.0.parameters\_fn\_kwargs.extrapolation \\
 & Defines whether to allow extrapolation in the first marginal bijector \\
\cline{1-2}
\multirow[t]{2}{*}{First Marginal Bijector Conditional Event Shape} & model\_kwargs.marginal\_bijectors.0.parameters\_fn\_kwargs.conditional\_event\_shape \\
 & Shape of conditional events for the first marginal bijector \\
\cline{1-2}
\multirow[t]{2}{*}{Polynomial Order for First Marginal Bijector} & model\_kwargs.marginal\_bijectors.0.parameters\_fn\_kwargs.polynomial\_order \\
 & Defines the order of the polynomial for computations in the first marginal bijector \\
\cline{1-2}
\multirow[t]{2}{*}{Second Hidden Units for Joint Bijector} & model\_kwargs.joint\_bijectors.0.parameters\_fn\_kwargs.hidden\_units.1 \\
 & Number of units in the second hidden layer of the joint bijector \\
\cline{1-2}
\multirow[t]{2}{*}{Third Hidden Units for Joint Bijector} & model\_kwargs.joint\_bijectors.0.parameters\_fn\_kwargs.hidden\_units.2 \\
 & Number of units in the third hidden layer of the joint bijector \\
\cline{1-2}
\multirow[t]{2}{*}{Parameter Shape 1 for Joint Bijector} & model\_kwargs.joint\_bijectors.0.parameters\_fn\_kwargs.parameter\_shape.1 \\
 & Defines the parameter shape dimension 1 for the joint bijector \\
\cline{1-2}
\multirow[t]{2}{*}{Is Conditional for Joint Bijector} & model\_kwargs.joint\_bijectors.0.parameters\_fn\_kwargs.conditional \\
 & Indicates if the joint bijector is conditional \\
\cline{1-2}
\multirow[t]{2}{*}{Two-Stage Training} & two\_stage\_training \\
 & Indicates if the model should be trained in two stages \\
\cline{1-2}
\multirow[t]{2}{*}{First Marginal Bijector} & model\_kwargs.marginal\_bijectors.0.bijector \\
 & Specifies the type of the first marginal bijector \\
\cline{1-2}
\multirow[t]{2}{*}{Invert First Marginal Bijector} & model\_kwargs.marginal\_bijectors.0.invert \\
 & Indicates if the first marginal bijector should allow inversion \\
\cline{1-2}
\multirow[t]{2}{*}{Marginal Bijector Domain Min} & model\_kwargs.marginal\_bijectors.0.bijector\_kwargs.domain.0 \\
 & Lower boundary for the domain of the first marginal bijector \\
\cline{1-2}
\multirow[t]{2}{*}{Marginal Bijector Domain Max} & model\_kwargs.marginal\_bijectors.0.bijector\_kwargs.domain.1 \\
 & Upper boundary for the domain of the first marginal bijector \\
\cline{1-2}
\multirow[t]{2}{*}{Marginal Bijector Extrapolation} & model\_kwargs.marginal\_bijectors.0.bijector\_kwargs.extrapolation \\
 & Defines whether to allow extrapolation in the first marginal bijector \\
\cline{1-2}
\multirow[t]{2}{*}{First Marginal Bijector Parameter Constraint Function} & model\_kwargs.marginal\_bijectors.0.parameters\_constraint\_fn \\
 & Constraint function for the parameters of the first marginal bijector \\
\cline{1-2}
\multirow[t]{2}{*}{Allow Flexible Bounds for First Marginal Bijector} & model\_kwargs.marginal\_bijectors.0.parameters\_constraint\_fn\_kwargs.allow\_flexible\_bounds \\
 & Defines if flexible bounds are allowed for the first marginal bijector \\
\cline{1-2}
\multirow[t]{2}{*}{Bounds constraints for First Marginal Bijector} & model\_kwargs.marginal\_bijectors.0.parameters\_constraint\_fn\_kwargs.bounds \\
 & Defining the constraints on the bounds for the first marginal bijector \\
\cline{1-2}
\multirow[t]{2}{*}{First Marginal Bijector Codomain Max} & model\_kwargs.marginal\_bijectors.0.parameters\_constraint\_fn\_kwargs.high \\
 & Defines the upper bound for parameters in the first marginal bijector \\
\cline{1-2}
\multirow[t]{2}{*}{First Marginal Bijector Codomain Min} & model\_kwargs.marginal\_bijectors.0.parameters\_constraint\_fn\_kwargs.low \\
 & Defines the lower bound for parameters in the first marginal bijector \\
\cline{1-2}
\multirow[t]{2}{*}{First Marginal Bijector Parameter Shape 0} & model\_kwargs.marginal\_bijectors.0.parameters\_fn\_kwargs.parameter\_shape.0 \\
 & Defines the parameter shape dimension 0 for the first marginal bijector \\
\cline{1-2}
\multirow[t]{2}{*}{First Marginal Bijector Parameter Shape 1} & model\_kwargs.marginal\_bijectors.0.parameters\_fn\_kwargs.parameter\_shape.1 \\
 & Defines the parameter shape dimension 1 for the first marginal bijector \\
\cline{1-2}
\multirow[t]{2}{*}{First Joint Bijector} & model\_kwargs.joint\_bijectors.0.bijector \\
 & Specifies the first bijector type for the joint distribution \\
\cline{1-2}
\multirow[t]{2}{*}{Parameter Shape 0 for Joint Bijector} & model\_kwargs.joint\_bijectors.0.parameters\_fn\_kwargs.parameter\_shape.0 \\
 & Defines the parameter shape dimension 0 for the joint bijector \\
\cline{1-2}
\multirow[t]{2}{*}{Marginal Epochs} & fit\_kwargs.0.epochs \\
 & Number of epochs for training the marginal distribution model \\
\cline{1-2}
\multirow[t]{2}{*}{Marginal Batch Size} & fit\_kwargs.0.batch\_size \\
 & Batch size for training the marginal distribution model \\
\cline{1-2}
\multirow[t]{2}{*}{Marginal Learning Rate} & fit\_kwargs.0.learning\_rate \\
 & Learning rate used to optimize the marginal distribution \\
\cline{1-2}
\multirow[t]{2}{*}{Marginal Reduce LR on Plateau} & fit\_kwargs.0.reduce\_lr\_on\_plateau \\
 & Whether to reduce learning rate on a plateau during training \\
\cline{1-2}
\multirow[t]{2}{*}{Marginal Early Stopping} & fit\_kwargs.0.early\_stopping \\
 & Enables early stopping for the marginal model training \\
\cline{1-2}
\multirow[t]{2}{*}{Marginal Monitor} & fit\_kwargs.0.monitor \\
 & Metric to monitor during training for early stopping \\
\cline{1-2}
\multirow[t]{2}{*}{Marginal Verbosity} & fit\_kwargs.0.verbose \\
 & Verbosity mode for the marginal model training \\
\cline{1-2}
\multirow[t]{2}{*}{Marginal Validation Split} & fit\_kwargs.0.validation\_split \\
 & Fraction of training data to be used as validation data for the marginal model \\
\cline{1-2}
\multirow[t]{2}{*}{Joint Epochs} & fit\_kwargs.1.epochs \\
 & Maximum number of epochs for training the joint distribution model \\
\cline{1-2}
\multirow[t]{2}{*}{Joint Batch Size} & fit\_kwargs.1.batch\_size \\
 & Batch size for training the joint distribution model \\
\cline{1-2}
\multirow[t]{2}{*}{Joint Learning Rate Decay Steps} & fit\_kwargs.1.learning\_rate.scheduler\_kwargs.decay\_steps \\
 & Total number of decay steps for the learning rate scheduler of the joint model \\
\cline{1-2}
\multirow[t]{2}{*}{Joint Initial Learning Rate} & fit\_kwargs.1.learning\_rate.scheduler\_kwargs.initial\_learning\_rate \\
 & Initial learning rate for the learning rate scheduler for the joint model \\
\cline{1-2}
\multirow[t]{2}{*}{Joint Learning Rate Scheduler} & fit\_kwargs.1.learning\_rate.scheduler\_name \\
 & Name of the learning rate scheduler for the joint model \\
\cline{1-2}
\multirow[t]{2}{*}{Joint Early Stopping} & fit\_kwargs.1.early\_stopping \\
 & Enables early stopping for the joint model training \\
\cline{1-2}
\multirow[t]{2}{*}{Joint Validation Split} & fit\_kwargs.1.validation\_split \\
 & Fraction of training data to be used as validation data for the joint model \\
\cline{1-2}
\multirow[t]{2}{*}{Masked Dimensions} & model\_kwargs.joint\_bijectors.0.bijector\_kwargs.num\_masked \\
 & Number of deimensions mask in the Coupling Flow \\
\cline{1-2}
\multirow[t]{2}{*}{First Nested Bijector} & model\_kwargs.joint\_bijectors.0.nested\_bijector.bijector \\
 & Specifies the type of nested bijector used in the first joint bijector \\
\cline{1-2}
\multirow[t]{2}{*}{Activation Function for Joint Bijector} & model\_kwargs.joint\_bijectors.0.parameters\_fn\_kwargs.activation \\
 & Activation function used in the neural network used to estimate the joint bijector parameters \\
\cline{1-2}
\multirow[t]{2}{*}{Batch Normalization for Joint Bijector} & model\_kwargs.joint\_bijectors.0.parameters\_fn\_kwargs.batch\_norm \\
 & Specifies if batch normalization should be applied in the joint bijector \\
\cline{1-2}
\multirow[t]{2}{*}{Dropout Rate for Joint Bijector} & model\_kwargs.joint\_bijectors.0.parameters\_fn\_kwargs.dropout \\
 & Dropout rate to apply in the joint bijector neural network \\
\cline{1-2}
\multirow[t]{2}{*}{First Hidden Units for Joint Bijector} & model\_kwargs.joint\_bijectors.0.parameters\_fn\_kwargs.hidden\_units.0 \\
 & Number of units in the first hidden layer of the joint bijector \\
\cline{1-2}
\multirow[t]{2}{*}{Second Hidden Units for Joint Bijector} & model\_kwargs.joint\_bijectors.0.parameters\_fn\_kwargs.hidden\_units.1 \\
 & Number of units in the second hidden layer of the joint bijector \\
\cline{1-2}
\multirow[t]{2}{*}{Third Hidden Units for Joint Bijector} & model\_kwargs.joint\_bijectors.0.parameters\_fn\_kwargs.hidden\_units.2 \\
 & Number of units in the third hidden layer of the joint bijector \\
\cline{1-2}
\multirow[t]{2}{*}{Input Shape 0 for Joint Bijector} & model\_kwargs.joint\_bijectors.0.parameters\_fn\_kwargs.input\_shape.0 \\
 & Defines the input shape dimension 0 for the joint bijector \\
\cline{1-2}
\multirow[t]{2}{*}{Parameter Shape 1 for Joint Bijector} & model\_kwargs.joint\_bijectors.0.parameters\_fn\_kwargs.parameter\_shape.1 \\
 & Defines the parameter shape dimension 1 for the joint bijector \\
\cline{1-2}
\multirow[t]{2}{*}{Nested Spline Bijector Range Min} & model\_kwargs.joint\_bijectors.0.nested\_bijector.bijector\_kwargs.range\_min \\
 & Minimum range value for the nested bijector transformation \\
\cline{1-2}
\multirow[t]{2}{*}{Interval Width for Nested Spline Bijector} & model\_kwargs.joint\_bijectors.0.nested\_bijector.parameters\_constraint\_fn\_kwargs.interval\_width \\
 & Specifies the interval width for constraint parameters \\
\cline{1-2}
\multirow[t]{2}{*}{Minimum Slope for Nested Spline Bijector} & model\_kwargs.joint\_bijectors.0.nested\_bijector.parameters\_constraint\_fn\_kwargs.min\_slope \\
 & Defines the minimum slope constraint for parameters \\
\cline{1-2}
\multirow[t]{2}{*}{Minimum Bin Width for Nested Spline Bijector} & model\_kwargs.joint\_bijectors.0.nested\_bijector.parameters\_constraint\_fn\_kwargs.min\_bin\_width \\
 & Defines the minimum bin width for the nested Spline bijector \\
\cline{1-2}
\multirow[t]{2}{*}{Number of Bins for Nested Spline Bijector} & model\_kwargs.joint\_bijectors.0.nested\_bijector.parameters\_constraint\_fn\_kwargs.nbins \\
 & Defines the number of bins used by the nested bijector \\
\cline{1-2}
\multirow[t]{2}{*}{JIT Compile} & compile\_kwargs.jit\_compile \\
 & Defines whether to compile the model using Just-In-Time compilation for performance \\
\cline{1-2}
\multirow[t]{2}{*}{Parameters Function for First Marginal Bijector} & model\_kwargs.marginal\_bijectors.0.parameters\_fn \\
 & Function for calculating parameters of the first marginal bijector \\
\cline{1-2}
\multirow[t]{2}{*}{Joint Bijector Conditional Event Shape} & model\_kwargs.joint\_bijectors.0.parameters\_fn\_kwargs.conditional\_event\_shape \\
 & Shape of the conditional events for the joint bijector \\
\cline{1-2}
\multirow[t]{2}{*}{Extrapolation for First Marginal Bijector} & model\_kwargs.marginal\_bijectors.0.parameters\_fn\_kwargs.extrapolation \\
 & Defines whether to allow extrapolation in the first marginal bijector \\
\cline{1-2}
\multirow[t]{2}{*}{First Marginal Bijector Conditional Event Shape} & model\_kwargs.marginal\_bijectors.0.parameters\_fn\_kwargs.conditional\_event\_shape \\
 & Shape of conditional events for the first marginal bijector \\
\cline{1-2}
\multirow[t]{2}{*}{Polynomial Order for First Marginal Bijector} & model\_kwargs.marginal\_bijectors.0.parameters\_fn\_kwargs.polynomial\_order \\
 & Defines the order of the polynomial for computations in the first marginal bijector \\
\cline{1-2}
\multirow[t]{2}{*}{Is Conditional for Joint Bijector} & model\_kwargs.joint\_bijectors.0.parameters\_fn\_kwargs.conditional \\
 & Indicates if the joint bijector is conditional \\
\cline{1-2}
\multirow[t]{2}{*}{Epochs} & fit\_kwargs.epochs \\
 & Maximum number of epochs \\
\cline{1-2}
\multirow[t]{2}{*}{Learning Rate Decay Steps} & fit\_kwargs.learning\_rate.scheduler\_kwargs.decay\_steps \\
 & Steps after which to decay the learning rate with the scheduler \\
\cline{1-2}
\multirow[t]{2}{*}{Learning Rate Scheduler} & fit\_kwargs.learning\_rate.scheduler\_name \\
 & Name of the learning rate scheduler \\
\cline{1-2}
\multirow[t]{2}{*}{Early Stopping} & fit\_kwargs.early\_stopping \\
 & Enable early stopping \\
\cline{1-2}
\multirow[t]{2}{*}{Output Distribution} & model\_kwargs.distribution \\
 & Specifies the type of output distribution used in the model \\
\cline{1-2}
\multirow[t]{2}{*}{Number of Flows} & model\_kwargs.num\_layers \\
 & Total number of layers in the model \\
\cline{1-2}
\multirow[t]{2}{*}{Bijector} & model\_kwargs.bijector \\
 & Specifies the bijector type to be used in the model \\
\cline{1-2}
\multirow[t]{2}{*}{Activation Function} & model\_kwargs.parameters\_fn\_kwargs.activation \\
 & Activation function of the used neural network \\
\cline{1-2}
\multirow[t]{2}{*}{Spline Bijector Domain Min} & model\_kwargs.bijector\_kwargs.range\_min \\
 & Lower bound of the quadratic spline bijector domain \\
\cline{1-2}
\multirow[t]{2}{*}{Interval Width for Nested Spline Bijector} & model\_kwargs.parameters\_constraint\_fn\_kwargs.interval\_width \\
 & Specifies the interval width for constraint parameters \\
\cline{1-2}
\multirow[t]{2}{*}{Minimum Slope for Nested Spline Bijector} & model\_kwargs.parameters\_constraint\_fn\_kwargs.min\_slope \\
 & Defines the minimum slope constraint for parameters \\
\cline{1-2}
\multirow[t]{2}{*}{Minimum Bin Width for Nested Spline Bijector} & model\_kwargs.parameters\_constraint\_fn\_kwargs.min\_bin\_width \\
 & Defines the minimum bin width for the nested Spline bijector \\
\cline{1-2}
\multirow[t]{2}{*}{Use Invertible Linear Transformations} & model\_kwargs.use\_invertible\_linear\_transformations \\
 & Specifies if invertible linear transformations should be employed \\
\cline{1-2}
\multirow[t]{2}{*}{Batch Size} & fit\_kwargs.batch\_size \\
 & Batch size used for training \\
\cline{1-2}
\multirow[t]{2}{*}{Initial Learning Rate} & fit\_kwargs.learning\_rate.scheduler\_kwargs.initial\_learning\_rate \\
 & Initial learning rate for the learning rate scheduler \\
\cline{1-2}
\multirow[t]{2}{*}{Number of Parameters} & model\_kwargs.num\_parameters \\
 & Total number of parameters in the model \\
\cline{1-2}
\multirow[t]{2}{*}{First Hidden Units} & model\_kwargs.parameters\_fn\_kwargs.hidden\_units.0 \\
 & Number of units in the first hidden layer of the neural network \\
\cline{1-2}
\multirow[t]{2}{*}{Second Hidden Units} & model\_kwargs.parameters\_fn\_kwargs.hidden\_units.1 \\
 & Number of units in the second hidden layer of the neural network \\
\cline{1-2}
\multirow[t]{2}{*}{Number of Bins for Nested Spline Bijector} & model\_kwargs.parameters\_constraint\_fn\_kwargs.nbins \\
 & Defines the number of bins used by the nested bijector \\
\cline{1-2}
\multirow[t]{2}{*}{Two-Stage Training} & two\_stage\_training \\
 & Indicates if the model should be trained in two stages \\
\cline{1-2}
\multirow[t]{2}{*}{First Marginal Bijector} & model\_kwargs.marginal\_bijectors.0.bijector \\
 & Specifies the type of the first marginal bijector \\
\cline{1-2}
\multirow[t]{2}{*}{Invert First Marginal Bijector} & model\_kwargs.marginal\_bijectors.0.invert \\
 & Indicates if the first marginal bijector should allow inversion \\
\cline{1-2}
\multirow[t]{2}{*}{Marginal Bijector Extrapolation} & model\_kwargs.marginal\_bijectors.0.bijector\_kwargs.extrapolation \\
 & Defines whether to allow extrapolation in the first marginal bijector \\
\cline{1-2}
\multirow[t]{2}{*}{First Marginal Bijector Parameter Constraint Function} & model\_kwargs.marginal\_bijectors.0.parameters\_constraint\_fn \\
 & Constraint function for the parameters of the first marginal bijector \\
\cline{1-2}
\multirow[t]{2}{*}{Allow Flexible Bounds for First Marginal Bijector} & model\_kwargs.marginal\_bijectors.0.parameters\_constraint\_fn\_kwargs.allow\_flexible\_bounds \\
 & Defines if flexible bounds are allowed for the first marginal bijector \\
\cline{1-2}
\multirow[t]{2}{*}{Bounds constraints for First Marginal Bijector} & model\_kwargs.marginal\_bijectors.0.parameters\_constraint\_fn\_kwargs.bounds \\
 & Defining the constraints on the bounds for the first marginal bijector \\
\cline{1-2}
\multirow[t]{2}{*}{First Marginal Bijector Codomain Max} & model\_kwargs.marginal\_bijectors.0.parameters\_constraint\_fn\_kwargs.high \\
 & Defines the upper bound for parameters in the first marginal bijector \\
\cline{1-2}
\multirow[t]{2}{*}{First Marginal Bijector Codomain Min} & model\_kwargs.marginal\_bijectors.0.parameters\_constraint\_fn\_kwargs.low \\
 & Defines the lower bound for parameters in the first marginal bijector \\
\cline{1-2}
\multirow[t]{2}{*}{Parameters Function for First Marginal Bijector} & model\_kwargs.marginal\_bijectors.0.parameters\_fn \\
 & Function for calculating parameters of the first marginal bijector \\
\cline{1-2}
\multirow[t]{2}{*}{First Marginal Bijector Parameter Shape 1} & model\_kwargs.marginal\_bijectors.0.parameters\_fn\_kwargs.parameter\_shape.1 \\
 & Defines the parameter shape dimension 1 for the first marginal bijector \\
\cline{1-2}
\multirow[t]{2}{*}{Marginal Reduce LR on Plateau} & fit\_kwargs.0.reduce\_lr\_on\_plateau \\
 & Whether to reduce learning rate on a plateau during training \\
\cline{1-2}
\multirow[t]{2}{*}{Marginal Early Stopping} & fit\_kwargs.0.early\_stopping \\
 & Enables early stopping for the marginal model training \\
\cline{1-2}
\multirow[t]{2}{*}{Marginal Monitor} & fit\_kwargs.0.monitor \\
 & Metric to monitor during training for early stopping \\
\cline{1-2}
\multirow[t]{2}{*}{Marginal Verbosity} & fit\_kwargs.0.verbose \\
 & Verbosity mode for the marginal model training \\
\cline{1-2}
\multirow[t]{2}{*}{Joint Epochs} & fit\_kwargs.1.epochs \\
 & Maximum number of epochs for training the joint distribution model \\
\cline{1-2}
\multirow[t]{2}{*}{Joint Learning Rate Decay Steps} & fit\_kwargs.1.learning\_rate.scheduler\_kwargs.decay\_steps \\
 & Total number of decay steps for the learning rate scheduler of the joint model \\
\cline{1-2}
\multirow[t]{2}{*}{Joint Learning Rate Scheduler} & fit\_kwargs.1.learning\_rate.scheduler\_name \\
 & Name of the learning rate scheduler for the joint model \\
\cline{1-2}
\multirow[t]{2}{*}{Joint Early Stopping} & fit\_kwargs.1.early\_stopping \\
 & Enables early stopping for the joint model training \\
\cline{1-2}
\multirow[t]{2}{*}{Type of Joint Flow} & model\_kwargs.joint\_flow\_type \\
 & Specifies the type of flow used in the model \\
\cline{1-2}
\multirow[t]{2}{*}{Number of Flows} & model\_kwargs.joint\_bijectors.num\_layers \\
 & Total number of layers in the joint bijectors \\
\cline{1-2}
\multirow[t]{2}{*}{Use Invertible Linear Transformations} & model\_kwargs.joint\_bijectors.use\_invertible\_linear\_transformations \\
 & Specifies if invertible linear transformations should be used \\
\cline{1-2}
\multirow[t]{2}{*}{Random Permutation Seed} & model\_kwargs.joint\_bijectors.random\_permutation\_seed \\
 & Seed value for random permutation in bijectors \\
\cline{1-2}
\multirow[t]{2}{*}{Joint Bijector} & model\_kwargs.joint\_bijectors.bijector \\
 & Specifies the bijector type to be used in the joint model \\
\cline{1-2}
\multirow[t]{2}{*}{Joint Bijector Range Min} & model\_kwargs.joint\_bijectors.bijector\_kwargs.range\_min \\
 & Minimum range value for the joint bijector transformation \\
\cline{1-2}
\multirow[t]{2}{*}{Interval Width for Joint Bijector} & model\_kwargs.joint\_bijectors.parameters\_constraint\_fn\_kwargs.interval\_width \\
 & Specifies the interval width for constraint parameters \\
\cline{1-2}
\multirow[t]{2}{*}{Minimum Slope for Joint Bijector} & model\_kwargs.joint\_bijectors.parameters\_constraint\_fn\_kwargs.min\_slope \\
 & Defines the minimum slope constraint for parameters \\
\cline{1-2}
\multirow[t]{2}{*}{Minimum Bin Width for Joint Bijector} & model\_kwargs.joint\_bijectors.parameters\_constraint\_fn\_kwargs.min\_bin\_width \\
 & Defines the minimum bin width for the joint bijector \\
\cline{1-2}
\multirow[t]{2}{*}{Number of Bins for Joint Bijector} & model\_kwargs.joint\_bijectors.parameters\_constraint\_fn\_kwargs.nbins \\
 & Defines the number of bins used by the joint bijector \\
\cline{1-2}
\multirow[t]{2}{*}{Number of Parameters for Joint Bijectors} & model\_kwargs.joint\_bijectors.num\_parameters \\
 & Total number of parameters in the joint bijectors \\
\cline{1-2}
\multirow[t]{2}{*}{Activation Function for MAF} & model\_kwargs.joint\_bijectors.maf\_parameters\_fn\_kwargs.activation \\
 & Activation function used in the neural network for MAF parameters \\
\cline{1-2}
\multirow[t]{2}{*}{Activation Function for X0 Parameters} & model\_kwargs.joint\_bijectors.x0\_parameters\_fn\_kwargs.activation \\
 & Activation function for x0 parameters in the joint bijector \\
\cline{1-2}
\multirow[t]{2}{*}{Batch Normalization for X0 Parameters} & model\_kwargs.joint\_bijectors.x0\_parameters\_fn\_kwargs.batch\_norm \\
 & Specifies if batch normalization should be applied to x0 parameters \\
\cline{1-2}
\multirow[t]{2}{*}{Dropout Rate for X0 Parameters} & model\_kwargs.joint\_bijectors.x0\_parameters\_fn\_kwargs.dropout \\
 & Dropout rate for the x0 parameters neural network \\
\cline{1-2}
\multirow[t]{2}{*}{First Marginal Bijector Parameter Shape 0} & model\_kwargs.marginal\_bijectors.0.parameters\_fn\_kwargs.parameter\_shape.0 \\
 & Defines the parameter shape dimension 0 for the first marginal bijector \\
\cline{1-2}
\multirow[t]{2}{*}{Marginal Epochs} & fit\_kwargs.0.epochs \\
 & Number of epochs for training the marginal distribution model \\
\cline{1-2}
\multirow[t]{2}{*}{Marginal Batch Size} & fit\_kwargs.0.batch\_size \\
 & Batch size for training the marginal distribution model \\
\cline{1-2}
\multirow[t]{2}{*}{Marginal Learning Rate} & fit\_kwargs.0.learning\_rate \\
 & Learning rate used to optimize the marginal distribution \\
\cline{1-2}
\multirow[t]{2}{*}{Joint Batch Size} & fit\_kwargs.1.batch\_size \\
 & Batch size for training the joint distribution model \\
\cline{1-2}
\multirow[t]{2}{*}{Joint Initial Learning Rate} & fit\_kwargs.1.learning\_rate.scheduler\_kwargs.initial\_learning\_rate \\
 & Initial learning rate for the learning rate scheduler for the joint model \\
\cline{1-2}
\multirow[t]{2}{*}{First Hidden Units for MAF} & model\_kwargs.joint\_bijectors.maf\_parameters\_fn\_kwargs.hidden\_units.0 \\
 & Number of units in the first hidden layer of the MAF \\
\cline{1-2}
\multirow[t]{2}{*}{Second Hidden Units for MAF} & model\_kwargs.joint\_bijectors.maf\_parameters\_fn\_kwargs.hidden\_units.1 \\
 & Number of units in the second hidden layer of the MAF \\
\cline{1-2}
\multirow[t]{2}{*}{First Hidden Units for X0 Network} & model\_kwargs.joint\_bijectors.x0\_parameters\_fn\_kwargs.hidden\_units.0 \\
 & Number of units in the first hidden layer for x0 parameters \\
\cline{1-2}
\multirow[t]{2}{*}{Second Hidden Units for X0 Network} & model\_kwargs.joint\_bijectors.x0\_parameters\_fn\_kwargs.hidden\_units.1 \\
 & Number of units in the second hidden layer for x0 parameters \\
\cline{1-2}
\multirow[t]{2}{*}{Two-Stage Training} & two\_stage\_training \\
 & Indicates if the model should be trained in two stages \\
\cline{1-2}
\multirow[t]{2}{*}{First Marginal Bijector} & model\_kwargs.marginal\_bijectors.0.bijector \\
 & Specifies the type of the first marginal bijector \\
\cline{1-2}
\multirow[t]{2}{*}{Invert First Marginal Bijector} & model\_kwargs.marginal\_bijectors.0.invert \\
 & Indicates if the first marginal bijector should allow inversion \\
\cline{1-2}
\multirow[t]{2}{*}{Marginal Bijector Domain Min} & model\_kwargs.marginal\_bijectors.0.bijector\_kwargs.domain.0 \\
 & Lower boundary for the domain of the first marginal bijector \\
\cline{1-2}
\multirow[t]{2}{*}{Marginal Bijector Domain Max} & model\_kwargs.marginal\_bijectors.0.bijector\_kwargs.domain.1 \\
 & Upper boundary for the domain of the first marginal bijector \\
\cline{1-2}
\multirow[t]{2}{*}{Marginal Bijector Extrapolation} & model\_kwargs.marginal\_bijectors.0.bijector\_kwargs.extrapolation \\
 & Defines whether to allow extrapolation in the first marginal bijector \\
\cline{1-2}
\multirow[t]{2}{*}{First Marginal Bijector Parameter Constraint Function} & model\_kwargs.marginal\_bijectors.0.parameters\_constraint\_fn \\
 & Constraint function for the parameters of the first marginal bijector \\
\cline{1-2}
\multirow[t]{2}{*}{Allow Flexible Bounds for First Marginal Bijector} & model\_kwargs.marginal\_bijectors.0.parameters\_constraint\_fn\_kwargs.allow\_flexible\_bounds \\
 & Defines if flexible bounds are allowed for the first marginal bijector \\
\cline{1-2}
\multirow[t]{2}{*}{Bounds constraints for First Marginal Bijector} & model\_kwargs.marginal\_bijectors.0.parameters\_constraint\_fn\_kwargs.bounds \\
 & Defining the constraints on the bounds for the first marginal bijector \\
\cline{1-2}
\multirow[t]{2}{*}{First Marginal Bijector Codomain Max} & model\_kwargs.marginal\_bijectors.0.parameters\_constraint\_fn\_kwargs.high \\
 & Defines the upper bound for parameters in the first marginal bijector \\
\cline{1-2}
\multirow[t]{2}{*}{First Marginal Bijector Codomain Min} & model\_kwargs.marginal\_bijectors.0.parameters\_constraint\_fn\_kwargs.low \\
 & Defines the lower bound for parameters in the first marginal bijector \\
\cline{1-2}
\multirow[t]{2}{*}{Parameters Function for First Marginal Bijector} & model\_kwargs.marginal\_bijectors.0.parameters\_fn \\
 & Function for calculating parameters of the first marginal bijector \\
\cline{1-2}
\multirow[t]{2}{*}{First Marginal Bijector Parameter Shape 0} & model\_kwargs.marginal\_bijectors.0.parameters\_fn\_kwargs.parameter\_shape.0 \\
 & Defines the parameter shape dimension 0 for the first marginal bijector \\
\cline{1-2}
\multirow[t]{2}{*}{First Marginal Bijector Parameter Shape 1} & model\_kwargs.marginal\_bijectors.0.parameters\_fn\_kwargs.parameter\_shape.1 \\
 & Defines the parameter shape dimension 1 for the first marginal bijector \\
\cline{1-2}
\multirow[t]{2}{*}{Second Marginal Bijector} & model\_kwargs.marginal\_bijectors.1.bijector \\
 & Specifies the type of second marginal bijector \\
\cline{1-2}
\multirow[t]{2}{*}{Invert Second Marginal Bijector} & model\_kwargs.marginal\_bijectors.1.invert \\
 & Indicates if the second marginal bijector should allow inversion \\
\cline{1-2}
\multirow[t]{2}{*}{Parameters Function for Second Marginal Bijector} & model\_kwargs.marginal\_bijectors.1.parameters\_fn \\
 & Function for calculating parameters of the second marginal bijector \\
\cline{1-2}
\multirow[t]{2}{*}{Second Marginal Bijector Conditional Event Shape} & model\_kwargs.marginal\_bijectors.1.parameters\_fn\_kwargs.conditional\_event\_shape \\
 & Shape of the conditional events for the second marginal bijector \\
\cline{1-2}
\multirow[t]{2}{*}{Extrapolation for Second Marginal Bijector} & model\_kwargs.marginal\_bijectors.1.parameters\_fn\_kwargs.extrapolation \\
 & Defines whether to allow extrapolation in the second marginal bijector \\
\cline{1-2}
\multirow[t]{2}{*}{Second Marginal Bijector Parameter Shape 0} & model\_kwargs.marginal\_bijectors.1.parameters\_fn\_kwargs.parameter\_shape.0 \\
 & Defines the parameter shape dimension 0 for the second marginal bijector \\
\cline{1-2}
\multirow[t]{2}{*}{Polynomial Order for Second Marginal Bijector} & model\_kwargs.marginal\_bijectors.1.parameters\_fn\_kwargs.polynomial\_order \\
 & Defines the order of the polynomial for computations in the second marginal bijector \\
\cline{1-2}
\multirow[t]{2}{*}{Marginal Epochs} & fit\_kwargs.0.epochs \\
 & Number of epochs for training the marginal distribution model \\
\cline{1-2}
\multirow[t]{2}{*}{Marginal Batch Size} & fit\_kwargs.0.batch\_size \\
 & Batch size for training the marginal distribution model \\
\cline{1-2}
\multirow[t]{2}{*}{Marginal Learning Rate} & fit\_kwargs.0.learning\_rate \\
 & Learning rate used to optimize the marginal distribution \\
\cline{1-2}
\multirow[t]{2}{*}{Marginal Reduce LR on Plateau} & fit\_kwargs.0.reduce\_lr\_on\_plateau \\
 & Whether to reduce learning rate on a plateau during training \\
\cline{1-2}
\multirow[t]{2}{*}{Marginal Early Stopping} & fit\_kwargs.0.early\_stopping \\
 & Enables early stopping for the marginal model training \\
\cline{1-2}
\multirow[t]{2}{*}{Marginal Monitor} & fit\_kwargs.0.monitor \\
 & Metric to monitor during training for early stopping \\
\cline{1-2}
\multirow[t]{2}{*}{Marginal Verbosity} & fit\_kwargs.0.verbose \\
 & Verbosity mode for the marginal model training \\
\cline{1-2}
\multirow[t]{2}{*}{Joint Epochs} & fit\_kwargs.1.epochs \\
 & Maximum number of epochs for training the joint distribution model \\
\cline{1-2}
\multirow[t]{2}{*}{Joint Batch Size} & fit\_kwargs.1.batch\_size \\
 & Batch size for training the joint distribution model \\
\cline{1-2}
\multirow[t]{2}{*}{Joint Learning Rate Decay Steps} & fit\_kwargs.1.learning\_rate.scheduler\_kwargs.decay\_steps \\
 & Total number of decay steps for the learning rate scheduler of the joint model \\
\cline{1-2}
\multirow[t]{2}{*}{Joint Learning Rate Scheduler} & fit\_kwargs.1.learning\_rate.scheduler\_name \\
 & Name of the learning rate scheduler for the joint model \\
\cline{1-2}
\multirow[t]{2}{*}{Joint Early Stopping} & fit\_kwargs.1.early\_stopping \\
 & Enables early stopping for the joint model training \\
\cline{1-2}
\multirow[t]{2}{*}{Second Marginal Bijector Domain Min} & model\_kwargs.marginal\_bijectors.1.parameters\_fn\_kwargs.domain.0 \\
 & Lower boundary for the domain of the second marginal bijector \\
\cline{1-2}
\multirow[t]{2}{*}{Second Marginal Bijector Domain Max} & model\_kwargs.marginal\_bijectors.1.parameters\_fn\_kwargs.domain.1 \\
 & Upper boundary for the domain of the second marginal bijector \\
\cline{1-2}
\multirow[t]{2}{*}{JIT Compile} & compile\_kwargs.jit\_compile \\
 & Defines whether to compile the model using Just-In-Time compilation for performance \\
\cline{1-2}
\multirow[t]{2}{*}{First Joint Bijector} & model\_kwargs.joint\_bijectors.0.bijector \\
 & Specifies the first bijector type for the joint distribution \\
\cline{1-2}
\multirow[t]{2}{*}{Invert First Joint Bijector} & model\_kwargs.joint\_bijectors.0.invert \\
 & Indicates if the first joint bijector should allow inversion \\
\cline{1-2}
\multirow[t]{2}{*}{Joint Bijector Conditional Event Shape} & model\_kwargs.joint\_bijectors.0.parameters\_fn\_kwargs.conditional\_event\_shape \\
 & Shape of the conditional events for the joint bijector \\
\cline{1-2}
\multirow[t]{2}{*}{Extrapolation for Joint Bijector} & model\_kwargs.joint\_bijectors.0.parameters\_fn\_kwargs.extrapolation \\
 & Defines whether to allow extrapolation in the joint bijector \\
\cline{1-2}
\multirow[t]{2}{*}{Parameter Shape 0 for Joint Bijector} & model\_kwargs.joint\_bijectors.0.parameters\_fn\_kwargs.parameter\_shape.0 \\
 & Defines the parameter shape dimension 0 for the joint bijector \\
\cline{1-2}
\multirow[t]{2}{*}{Polynomial Order for Joint Bijector} & model\_kwargs.joint\_bijectors.0.parameters\_fn\_kwargs.polynomial\_order \\
 & Defines the order of the polynomial for computations in the joint bijector \\
\cline{1-2}
\multirow[t]{2}{*}{Joint Initial Learning Rate} & fit\_kwargs.1.learning\_rate.scheduler\_kwargs.initial\_learning\_rate \\
 & Initial learning rate for the learning rate scheduler for the joint model \\
\cline{1-2}
\multirow[t]{2}{*}{Type of Joint Flow} & model\_kwargs.joint\_flow\_type \\
 & Specifies the type of flow used in the model \\
\cline{1-2}
\multirow[t]{2}{*}{Number of Flows} & model\_kwargs.joint\_bijectors.num\_layers \\
 & Total number of layers in the joint bijectors \\
\cline{1-2}
\multirow[t]{2}{*}{Joint Bijector} & model\_kwargs.joint\_bijectors.bijector \\
 & Specifies the bijector type to be used in the joint model \\
\cline{1-2}
\multirow[t]{2}{*}{Joint Bijector Range Min} & model\_kwargs.joint\_bijectors.bijector\_kwargs.range\_min \\
 & Minimum range value for the joint bijector transformation \\
\cline{1-2}
\multirow[t]{2}{*}{Interval Width for Joint Bijector} & model\_kwargs.joint\_bijectors.parameters\_constraint\_fn\_kwargs.interval\_width \\
 & Specifies the interval width for constraint parameters \\
\cline{1-2}
\multirow[t]{2}{*}{Minimum Slope for Joint Bijector} & model\_kwargs.joint\_bijectors.parameters\_constraint\_fn\_kwargs.min\_slope \\
 & Defines the minimum slope constraint for parameters \\
\cline{1-2}
\multirow[t]{2}{*}{Minimum Bin Width for Joint Bijector} & model\_kwargs.joint\_bijectors.parameters\_constraint\_fn\_kwargs.min\_bin\_width \\
 & Defines the minimum bin width for the joint bijector \\
\cline{1-2}
\multirow[t]{2}{*}{Number of Bins for Joint Bijector} & model\_kwargs.joint\_bijectors.parameters\_constraint\_fn\_kwargs.nbins \\
 & Defines the number of bins used by the joint bijector \\
\cline{1-2}
\multirow[t]{2}{*}{Number of Parameters for Joint Bijectors} & model\_kwargs.joint\_bijectors.num\_parameters \\
 & Total number of parameters in the joint bijectors \\
\cline{1-2}
\multirow[t]{2}{*}{Activation Function for MAF} & model\_kwargs.joint\_bijectors.maf\_parameters\_fn\_kwargs.activation \\
 & Activation function used in the neural network for MAF parameters \\
\cline{1-2}
\multirow[t]{2}{*}{First Hidden Units for MAF} & model\_kwargs.joint\_bijectors.maf\_parameters\_fn\_kwargs.hidden\_units.0 \\
 & Number of units in the first hidden layer of the MAF \\
\cline{1-2}
\multirow[t]{2}{*}{Second Hidden Units for MAF} & model\_kwargs.joint\_bijectors.maf\_parameters\_fn\_kwargs.hidden\_units.1 \\
 & Number of units in the second hidden layer of the MAF \\
\cline{1-2}
\multirow[t]{2}{*}{Activation Function for X0 Parameters} & model\_kwargs.joint\_bijectors.x0\_parameters\_fn\_kwargs.activation \\
 & Activation function for x0 parameters in the joint bijector \\
\cline{1-2}
\multirow[t]{2}{*}{Batch Normalization for X0 Parameters} & model\_kwargs.joint\_bijectors.x0\_parameters\_fn\_kwargs.batch\_norm \\
 & Specifies if batch normalization should be applied to x0 parameters \\
\cline{1-2}
\multirow[t]{2}{*}{Dropout Rate for X0 Parameters} & model\_kwargs.joint\_bijectors.x0\_parameters\_fn\_kwargs.dropout \\
 & Dropout rate for the x0 parameters neural network \\
\cline{1-2}
\multirow[t]{2}{*}{First Hidden Units for X0 Network} & model\_kwargs.joint\_bijectors.x0\_parameters\_fn\_kwargs.hidden\_units.0 \\
 & Number of units in the first hidden layer for x0 parameters \\
\cline{1-2}
\multirow[t]{2}{*}{Second Hidden Units for X0 Network} & model\_kwargs.joint\_bijectors.x0\_parameters\_fn\_kwargs.hidden\_units.1 \\
 & Number of units in the second hidden layer for x0 parameters \\
\cline{1-2}
\multirow[t]{2}{*}{Nested Bernstein Bijector Domain Min} & model\_kwargs.joint\_bijectors.0.parameters\_fn\_kwargs.domain.0 \\
 & Lower boundary for the nested Bernstein bijector domain \\
\cline{1-2}
\multirow[t]{2}{*}{Nested Bernstein Bijector Domain Max} & model\_kwargs.joint\_bijectors.0.parameters\_fn\_kwargs.domain.1 \\
 & Upper boundary for the nested Bernstein bijector domain \\
\cline{1-2}
\multirow[t]{2}{*}{Joint Bijector Extrapolation} & model\_kwargs.joint\_bijectors.bijector\_kwargs.extrapolation \\
 & Defines whether to allow extrapolation in the joint bijector \\
\cline{1-2}
\multirow[t]{2}{*}{Nested Bernstein Bijector Domain Min} & model\_kwargs.joint\_bijectors.bijector\_kwargs.domain.0 \\
 & Lower boundary for the nested Bernstein bijector domain \\
\cline{1-2}
\multirow[t]{2}{*}{Nested Bernstein Bijector Domain Max} & model\_kwargs.joint\_bijectors.bijector\_kwargs.domain.1 \\
 & Upper boundary for the nested Bernstein bijector domain \\
\cline{1-2}
\multirow[t]{2}{*}{Invert Joint Bijector} & model\_kwargs.joint\_bijectors.invert \\
 & Indicates if the joint bijector should be invertible \\
\cline{1-2}
\multirow[t]{2}{*}{Joint Bijector Codomain Min} & model\_kwargs.joint\_bijectors.parameters\_constraint\_fn\_kwargs.low \\
 & Defines the lower bound for the corresponding parameters \\
\cline{1-2}
\multirow[t]{2}{*}{Joint Bijector Codomain Max} & model\_kwargs.joint\_bijectors.parameters\_constraint\_fn\_kwargs.high \\
 & Defines the upper bound for the corresponding parameters \\
\cline{1-2}
\multirow[t]{2}{*}{Third Hidden Units for MAF} & model\_kwargs.joint\_bijectors.maf\_parameters\_fn\_kwargs.hidden\_units.2 \\
 & Number of units in the third hidden layer of the MAF \\
\cline{1-2}
\multirow[t]{2}{*}{Is Conditional for MAF} & model\_kwargs.joint\_bijectors.maf\_parameters\_fn\_kwargs.conditional \\
 & Indicates if the MAF parameters are conditional \\
\cline{1-2}
\multirow[t]{2}{*}{MAF Conditional Event Shape} & model\_kwargs.joint\_bijectors.maf\_parameters\_fn\_kwargs.conditional\_event\_shape \\
 & Shape of the conditional events for the MAF \\
\cline{1-2}
\multirow[t]{2}{*}{Third Hidden Units for X0 Network} & model\_kwargs.joint\_bijectors.x0\_parameters\_fn\_kwargs.hidden\_units.2 \\
 & Number of units in the third hidden layer for x0 parameters \\
\cline{1-2}
\multirow[t]{2}{*}{Is Conditional for X0 Parameters} & model\_kwargs.joint\_bijectors.x0\_parameters\_fn\_kwargs.conditional \\
 & Indicates if the x0 parameters are conditional \\
\cline{1-2}
\multirow[t]{2}{*}{X0 Parameters Conditional Event Shape} & model\_kwargs.joint\_bijectors.x0\_parameters\_fn\_kwargs.conditional\_event\_shape \\
 & Shape of the conditional events for x0 parameters \\
\cline{1-2}
\end{longtable}

\end{landscape}

\printbibliography